\documentclass[11pt]{article}

\usepackage{acl}

\usepackage{times}
\usepackage{latexsym}
\usepackage[T1]{fontenc}
\usepackage[utf8]{inputenc}
\usepackage{microtype}
\usepackage{inconsolata}
\usepackage{graphicx}
\usepackage{tikz}
\usetikzlibrary{positioning,backgrounds,fit}
\usepackage{booktabs}
\usepackage{multirow}
\usepackage{array}
\usepackage{amsmath}
\usepackage{placeins}

\title{Synthetic Longitudinal Case Notes for Human Services: \\
       A Free-Text Corpus and Utility-Centered Evaluation}

\author{Hender Lin\textsuperscript{1}, Whitney Nelson\textsuperscript{2}, Junyi Jessy Li\textsuperscript{3}, and Min Kyung Lee\textsuperscript{2,4} \\
  \textsuperscript{1}Department of Computer Science, The University of Texas at Austin \\
  \textsuperscript{2}School of Information, The University of Texas at Austin \\
  \textsuperscript{3}Department of Linguistics, The University of Texas at Austin \\
  \textsuperscript{4}School of Information, University of Michigan Ann Arbor \\
  \texttt{hlin01@utexas.edu}, \texttt{whitney.nelson@utexas.edu} \\
  \texttt{jessy@austin.utexas.edu}, \texttt{mkleehai@umich.edu}}

\hypersetup{
  pdftitle={Synthetic Longitudinal Case Notes for Human Services: A Free-Text Corpus and Utility-Centered Evaluation},
  pdfauthor={Hender Lin; Whitney Nelson; Junyi Jessy Li; Min Kyung Lee}
}

\begin{document}
\maketitle

\begin{abstract}
Longitudinal human service case notes---progress records written after each client contact---accumulate needs, decisions, and trajectory over months or years, supporting trajectory reasoning, primary-need identification, and handoff summarization. Even after removing Personally Identifiable Information (PII), the longitudinal combination of events, services, and timing creates privacy risks that make such data hard to share. No free-text longitudinal human service case notes are public; closest precedents release tabular client records or single-encounter clinical notes. We release, to our knowledge, the first free-text synthetic longitudinal case note corpus for human services---$3{,}137$ notes across $87$ synthetic clients---from a four-stage structured-first pipeline that learns distributions over a private case-management corpus and decouples sampling from generation behind an explicit, audited privacy boundary. A six-part utility evaluation (Parts~A--F), a privacy sketch (Part~G), and a caseworker pilot ($N{=}8$) support bounded utility for training, review, and benchmarking. A primary-need classifier trained on the synthetic corpus reaches $83\%$ of the upper-bound macro-F1 on held-out real notes, vs.\ $51\%$ for a direct-prompt baseline.
\end{abstract}

\section{Introduction}

Longitudinal case notes are the central documentary record in human service settings---agencies providing housing assistance, mental health support, child welfare, and related social services \citep{gondimalla2024aligning,nelson2024homelessness}. A single note captures one encounter; a sequence of notes reveals whether a housing crisis stabilizes, whether a referral leads to action, whether a case escalates or re-opens after disengagement. The value is cumulative: reasoning about client trajectories requires understanding how needs, barriers, and decisions accumulate and change across encounters, supporting tasks such as predicting a client's primary service need, summarizing a case for handoff, and reasoning about how circumstances change over time---tasks that require records of multiple encounters. These same properties make case notes difficult to share: even after names and contact information are removed, the longitudinal combination of events, services, and timing remains identifying in a way that a single de-identified record does not.

Existing corpora in adjacent domains leave the free-text, longitudinal case note setting unaddressed. Clinical resources such as MIMIC-III discharge summaries \citep{johnson2016mimic} and the 2010 i2b2 shared task \citep{uzuner2011i2b2} include encounter-level medical documentation oriented toward medical decision support, not the multi-encounter, service-coordination narratives social workers write to track a client's housing, benefits, employment, and follow-up over months or years; the cross-domain genre gap is large enough that off-the-shelf clinical models are unlikely to transfer (Appendix~\ref{sec:appendix-related-work}). The closest precedent releases tabular human service records only \citep{pickens2023synthhumanservices}. This paper contributes a corpus and evaluation framework that treat \emph{utility} rather than indistinguishability as the target: a synthetic corpus can remain distinguishable from real data while still enabling training, benchmarking, and methodological research.

We make four contributions toward this goal. (i)~\textbf{The corpus}---$3{,}137$ free-text synthetic notes across $87$ clients, each paired with a schema-validated JSON of structured synthetic facts (presenting issues, services, barriers, resources, outcomes); to our knowledge the first publicly released free-text synthetic \emph{longitudinal} case note corpus for human service settings, generated from learned distributions over a private real case-management note corpus. (ii)~\textbf{The pipeline}---a structured-first generation pipeline that separates trajectory construction from prose realization behind an explicit privacy boundary: unredacted source notes, source identifier records, and real extracted fact trajectories stay local, and the only real-derived text the external generator receives is a small rotated pool of PII-redacted style exemplars whose reuse the release audit screens (\S\ref{sec:partd}). (iii)~\textbf{The evaluation framework}---a six-Part utility-centered evaluation (Parts A--F) covering client-scenario coverage, documentation structure, embedding fidelity, no-copy auditing, fact-to-text alignment, and downstream-task transfer, complemented by a privacy-side nearest-neighbor membership-inference sketch (Part G) and a practitioner pilot ($N{=}8$).

(iv)~\textbf{Quantified bounded-utility evidence, anchored on downstream-task transfer.} The headline result is the train-on-synth / eval-on-real reading (Part~F, \S\ref{sec:partf}): a primary-need classifier---which predicts the client's most urgent service need---trained on the synthetic corpus reaches \textbf{$83\%$} of the in-distribution upper-bound macro-F1 on $6{,}254$ held-out real notes, vs.\ $51\%$ for the same generator's direct-prompt baseline. At matched training size of $162$ notes, the synthetic condition retains a $+0.050$ transfer-ratio advantage across ten holdout splits. Structured-first generation also outperforms the direct-prompt baseline on two size-insensitive distribution metrics (\S\ref{sec:baseline-summary}): $4.1\times$ closer on discourse-distribution distance (Part~B) and $7.1\times$ closer on embedding-space distance (Part~C). The corpus shows no detectable membership signal on either extracted-fact or prose features (Part~G, Appendix~\ref{sec:appendix-mia}), while remaining distinguishable from real data---supporting use for research, training, and benchmarking rather than as a substitute for real records.

\section{Related Work}
\label{sec:related-work}

\paragraph{Privacy-sensitive NLP data and synthetic alternatives.}
Human service case notes sit at an underserved intersection: as sensitive as clinical records yet structurally distinct in genre, purpose, and discourse, with limited representation in both sensitive-NLP corpora and synthetic-data research. Clinical resources such as MIMIC-III \citep{johnson2016mimic} and de-identification shared tasks \citep{stubbs2015i2b2} demonstrate both the value and the infrastructure cost of releasing sensitive narrative data, but they target a different genre \citep{gondimalla2024aligning,nelson2024homelessness}. Synthetic data is often proposed as the privacy escape hatch but is neither automatically private nor automatically useful: empirical similarity is not formal privacy \citep{stadler2022syntheticnotsilver,royalsociety2022synth}, and even differentially-private synthetic text underperforms on critical subgroups despite formal DP guarantees \citep{ramesh2024dpeval}. The closest precedent for synthetic human service data, \citet{pickens2023synthhumanservices}, releases tabular client records only (demographics and service-utilization codes), leaving prose realization and discourse structure unaddressed.

\paragraph{LLM-based synthetic clinical / sensitive-domain text.}
An early clinical-text example of the train-on-synth / eval-on-real utility benchmark for synthetic corpora is \citet{ive2020artificialmh}, who train a Transformer LM on de-identified mental-health records to evaluate text-classification transfer---a precedent no prior work has extended to longitudinal human service case notes. Recent LLM-based work spans three patterns: \emph{synthetic corpus construction or augmentation} (\citealp{mitra2025synthsbdh}, \citealp{kim2024generalizing}, \citealp{tang2023synthclinical}); \emph{synthetic supervision rather than synthetic text} (\citealp{mishra2024synfacedit}); and \emph{direction-of-flow ablations} (\citealp{li2023twodirections}). Cautionary work shows classification using generic LLM-generated text degrades on subjective tasks \citep{li2023synthetic} and that textbook-style pure synthetic pretraining can exhibit model-collapse-like patterns \citep{demystify2025}; see also \citep{long2024llmsynthsurvey}. \citet{kruse2025longitudinal} report that even long-context LLMs struggle with temporal progression in clinical text, motivating our explicit longitudinal sampler. Privacy-aware pipelines employ explicit threat models: entity-aware control codes \citep{zhao2025controlled}, attack-based generator audits \citep{wang2022promptehr}, and client--server and DP variants \citep{knowledgesg2024,ochs2025diffusion}. We sit on the label-to-data end with a four-stage decoupled pipeline (\S\ref{sec:baseline-summary} validates this empirically).

\paragraph{Synthetic data evaluation and our position.}
Synthetic-data releases are commonly evaluated along fidelity, utility, and privacy axes \citep{ramesh2024dpeval,ramesh2025synthtexteval}. Standard fidelity instruments are classifier two-sample tests \citep{lopezpaz2017classifier}, MMD \citep{gretton2012mmd}, Fr\'echet-style distances \citep{heusel2017gans}, and Kynk\"a\"anniemi-style precision/recall \citep{kynkaanniemi2019improved}. Domain-specific extensions add protocol adherence in synthetic therapy dialogues \citep{petherapy2025}, trajectory-level emotional-arc comparisons \citep{cbtarcs2025}, and discourse-similarity measures for LLM text \citep{namuduri2025qudsim}. Process-supervised reward models score clinical-note steps \citep{wang2025processrm}, while checklist methods evaluate note or summary quality \citep{zhou2025feedback,dou2026gavel}; these are orthogonal to the corpus-level axes we evaluate.

Data Statements \citep{bender2018data} and Datasheets \citep{gebru2021datasheets} provide dataset-documentation frameworks; Model Cards serve released models \citep{mitchell2019modelcards}. We adopt both dataset frameworks (Appendices~\ref{sec:datastatement} and~\ref{sec:datasheet}) and evaluate fidelity with the instruments above (Parts~A--C; \S\ref{sec:eval-framework}). What no prior release offers, to our knowledge, is an evaluation pairing all six utility Parts (\S\ref{sec:eval-framework}) with practitioner review on case notes ($N{=}8$, \S\ref{sec:practitioner}). Three further axes distinguish this work: a structural privacy posture (an explicit local/API boundary plus empirical no-copy and membership-inference screens; \S\ref{sec:partd}, Appendix~\ref{sec:appendix-mia}) rather than formal DP; a label-to-data pipeline rather than rich-prompt generation; and a downstream-task transfer reading on the fixed corpus itself (Part~F, \S\ref{sec:partf}). Appendix~\ref{sec:appendix-related-work} gives the per-precedent discussion.

\section{Dataset and Source Constraints}
\label{sec:source-constraints}

The private source corpus consists of $63{,}142$ longitudinal case notes from $1{,}682$ real clients in a US-based agency that provides housing-focused case management. After filtering, $62{,}901$ notes from $1{,}681$ clients remain usable for quantitative evaluation. One client has no usable notes after sentinel (placeholder) rows are removed.

The synthetic corpus contains $3{,}137$ notes across $87$ synthetic clients ($54$ multi-note and $33$ single-note clients); the trajectory tail is intentionally heavy (median $13.5$ notes per multi-note client, mean $57.5$, max $363$) to cover both single-touch cases and extended case-management arcs (see Appendix~\ref{sec:appendix-dataset} for an example synthetic note paired with its structured-fact JSON). A separate ablation (Appendix~\ref{sec:appendix-transfer}) adds $30$ category-conditioned synthetic clients to improve transfer on two sparse primary-need labels; those clients are not part of the fixed corpus. We treat case notes as having three structures---\emph{content} (needs, services, barriers, resources, outcomes), \emph{longitudinal} (how facts change within a trajectory), and \emph{documentation} (caseworker phrasing, abbreviation, formatting)---and the pipeline is designed to preserve all three while excluding real records from release.

Two privacy boundaries are the central design constraints. The \emph{release} boundary: only the $3{,}137$ synthetic notes, their structured synthetic facts, evaluation code, and documentation leave; source notes at both redaction stages, extracted facts, learned distributions, identifier records, few-shot examples, logs, and the full history stay private. The \emph{API} boundary: unredacted source notes, source identifier records, and real extracted fact trajectories stay local; for the fixed corpus, the only real-derived text sent to the external synthesis model is a rotated pool of $146$ PII-redacted exemplar notes used for style conditioning (\S\ref{sec:pipeline}), subject to residual-redaction risk, and the fixed corpus is audited against exactly those exemplars for copying (\S\ref{sec:partd}). The corpus is intended for research, education, prototyping, and benchmarking in privacy-sensitive NLP (case-note summarization, handoff, primary-need classification, extraction, trajectory summarization, retrieval); it is \emph{not} intended for operational decisions about real clients, real-population estimation, or real-caseworker evaluation. The count chain, release matrix, and intended/out-of-scope uses are in Appendix~\ref{sec:appendix-dataset}.

\paragraph{Availability.} The fixed v1.0 corpus and companion software are available through Zenodo (\url{https://doi.org/10.5281/zenodo.24269796}). Under the approved research-use terms, each researcher must obtain approval for a noncommercial project.

\section{Generation Pipeline}
\label{sec:pipeline}

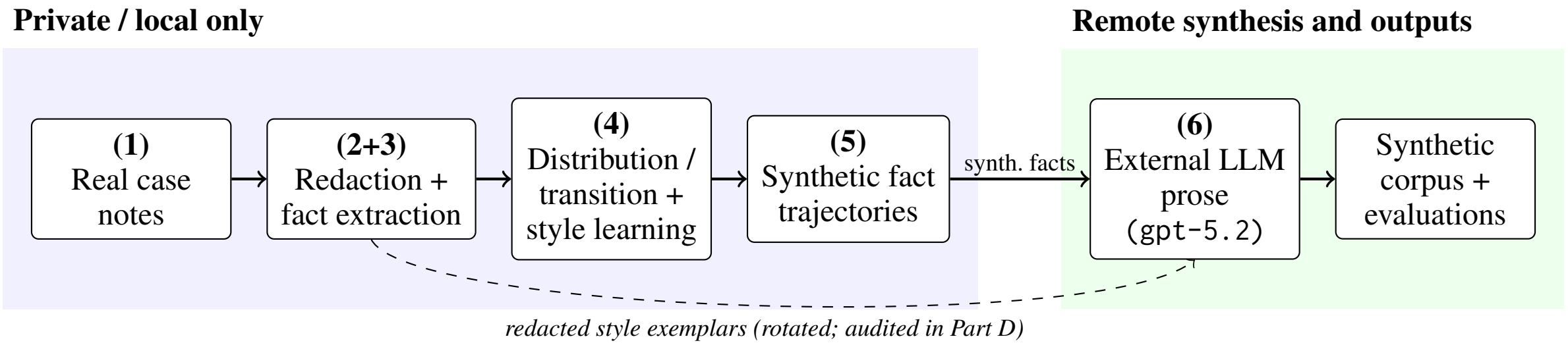
\begin{figure*}[t!]
\centering
\begin{tikzpicture}[
  font=\footnotesize,
  stage/.style={draw, rounded corners=2pt, inner sep=4pt, align=center,
                minimum width=2.0cm, minimum height=1.05cm, fill=white},
  bandlbl/.style={font=\footnotesize\bfseries, anchor=south west,
                  inner sep=3pt},
  node distance=0.35cm,
]
\node[stage] (rn) {\textbf{(1)}\\ Real case\\ notes};
\node[stage, right=of rn] (redex) {\textbf{(2+3)}\\ Redaction +\\ fact extraction};
\node[stage, right=of redex] (learn) {\textbf{(4)}\\ Distribution /\\ transition +\\ style learning};
\node[stage, right=of learn] (sf) {\textbf{(5)}\\ Synthetic fact\\ trajectories};
\node[stage, right=1.4cm of sf] (api) {\textbf{(6)}\\ External LLM\\ prose\\ \texttt{(gpt-5.2)}};
\node[stage, right=of api] (out) {Synthetic\\ corpus +\\ evaluations};

\draw[->, thick] (rn) -- (redex);
\draw[->, thick] (redex) -- (learn);
\draw[->, thick] (learn) -- (sf);
\draw[->, thick] (sf) -- node[midway, above, font=\scriptsize, inner sep=0.5pt] {synth.\ facts} (api);
\draw[->, dashed] (redex.south) .. controls +(0.4,-0.75) and +(-0.9,-0.75) .. node[below, font=\scriptsize\itshape] {redacted style exemplars (rotated; audited in Part D)} (api.south);
\draw[->, thick] (api) -- (out);

\begin{scope}[on background layer]
  \node[fit=(rn)(redex)(learn)(sf), draw=none, fill=blue!6,
        inner xsep=8pt, inner ysep=14pt] (privbg) {};
  \node[fit=(api)(out), draw=none, fill=green!7,
        inner xsep=8pt, inner ysep=14pt] (relbg) {};
\end{scope}
\node[bandlbl] at (privbg.north west) {Private / local only};
\node[bandlbl] at (relbg.north west) {Remote synthesis and outputs};
\end{tikzpicture}
\caption{Privacy boundaries of the structured-first pipeline. Local stages process source notes and derived artifacts. The external generator receives synthetic facts (solid arrow) and rotated PII-redacted style exemplars (dashed arrow); unredacted source notes, source identifier records, and real extracted fact trajectories stay local. Part~D audits v1.0 notes against their conditioning exemplars (\S\ref{sec:partd}).}
\label{fig:pipeline}
\end{figure*}

The pipeline is designed around a structured-first principle: synthetic notes should not be produced through unconstrained prompting alone but grounded in structured facts, corpus-level distributions, longitudinal transitions, and documentation-style controls. The method offers a five-step recipe that could be adapted to other privacy-constrained narrative-record domains: (1)~\emph{extract} structured facts from redacted records with a local model; (2)~\emph{estimate} field marginals, longitudinal transitions, and style statistics; (3)~\emph{sample} synthetic fact trajectories from those estimates; (4)~\emph{realize} each fact bundle as prose with a generator; (5)~\emph{audit} the output against the source corpus before release. Figure~\ref{fig:pipeline} shows the instantiation here (figure stages 2--5 implement recipe steps 1--3 with redaction split out). Everything up to fact sampling uses \emph{local} models only (spaCy NER, then a local Ollama daemon running \texttt{qwen3:32b-q8\_0}); prose realization (figure stage 6) is the generation pipeline's only external API call (OpenAI \texttt{gpt-5.2}) and receives synthetic structured facts plus the rotated PII-redacted style exemplars---the API boundary defined in \S\ref{sec:source-constraints}.

\paragraph{Stages.} Figure stages~(2+3)--6 implement recipe steps~(1)--(4); recipe step~(5) is the release audit. Stages~1--2 (split-by-client, PII redaction) are introduced in \S\ref{sec:source-constraints}. (3)~\emph{Extraction}: a local Ollama \texttt{qwen3:32b-q8\_0} (Appendix~\ref{sec:reproducibility}) converts each post-redaction note into a schema-validated JSON capturing presenting issues, services, barriers, crises, resources, collateral, housing/legal status, and narrative context. The schema covers $40$ fields across $12$ groups (full reference in Appendix~\ref{sec:appendix-schema}). (4)~\emph{Distribution analysis} estimates field marginals, first-note distributions, same-note conditionals, list co-occurrence, short/long-gap transitions, per-issue persistence, and text-style features. (5)~\emph{Synthetic fact sampling} draws complete latent trajectories then applies observed missingness (\emph{sample-then-mask}) plus consistency rules.

(6)~\emph{Style-conditioned prose synthesis}: \texttt{gpt-5.2} is conditioned on the synthetic fact JSON, prior-note context, a sampled \texttt{text\_style} block, and a deterministically rotated set of few-shot style exemplars drawn from the private pool of $146$ PII-redacted real notes (the exemplars are never released; the ones used for each note are logged and audited in Part~D); per-note post-processing applies placeholder validation, length checks, prohibited-phrase filtering, and grammar adjustment. Per-stage hyperparameters are in Appendix~\ref{sec:reproducibility}. The separate software-v2 sampling extension and its experimental evidence are described in Appendix~\ref{sec:appendix-v2}; it did not produce the fixed corpus evaluated here.

\paragraph{Direct-prompt baseline.}
\label{sec:baseline-summary}
A controlled ablation tests whether the four-stage decoupling buys anything over a single direct-prompt call to the same generator. The baseline asks \texttt{gpt-5.2} to generate notes from a one-line scenario seed (drawn from the cross-product of schema visit-type, issue, and service values) under the same documentation-style system instruction---no extracted facts, no learned distributions, no sampler, no trajectory context, and no few-shot exemplars. The comparison is therefore an ablation of the full structured-first stack (fact conditioning, trajectory context, and exemplar-based style conditioning together), not a claim against the strongest possible prompting system. The $162$-note baseline train pool is $\sim\!12\times$ smaller than the synthetic train pool used for Part~F ($162$ vs.\ $1{,}895$ labeled notes; Appendix~\ref{sec:appendix-transfer}); Part~B/\allowbreak Part~C fidelity metrics do not scale with downstream classifier training-set size. The baseline diverges \textbf{$4.1\times$} more on Part B discourse-move JSD ($0.074$ vs.\ $0.018$) and \textbf{$7.1\times$} more on Part C BGE-Large MMD$^2$ ($0.172$ vs.\ $0.024$) than the structured pipeline against the same $62{,}901$-note real corpus (Table~\ref{tab:baseline}; full protocol in Appendix~\ref{sec:appendix-baseline}).

\begin{table}[t]
\centering
\small
\setlength{\tabcolsep}{3pt}
\begin{tabular}{lrrr}
\toprule
\textbf{Metric} & \textbf{Pipeline} & \textbf{Direct} & \textbf{Ratio} \\
\midrule
Part B JSD                & 0.018 & 0.074 & $4.1\times$ \\
Part C MMD$^2$ (BGE)    & 0.024 & 0.172 & $7.1\times$ \\
\bottomrule
\end{tabular}
\caption{Direct-prompt baseline vs.\ structured pipeline, scored against the same $62{,}901$-note real corpus. Lower is closer to real. Full protocol in Appendix~\ref{sec:appendix-baseline}.}
\label{tab:baseline}
\end{table}

\section{Evaluation Framework}
\label{sec:eval-framework}

We evaluate the corpus by the functional properties that make case notes useful---\emph{do synthetic notes preserve enough content, longitudinal structure, and documentation style to support bounded research, training, and evaluation uses?}---via six Parts plus a privacy-side sketch (Appendix~\ref{sec:appendix-mia}). \emph{Part~A (client scenario coverage)} compares real and synthetic clients in a structured-fact-derived feature space (Table~\ref{tab:partavector}), split into Track~1 (multi-note, $33$-d trajectory vector) and Track~2 (single-note, $28$-d), reporting nearest-neighbor distances, coverage/precision, balanced 1-NN \citep{lopezpaz2017classifier}, and size-matched AUC. \emph{Part~B} labels sentences with one of $12$ documentation moves using a local embedding-prototype zero-shot classifier (\texttt{BAAI/bge-large-en-v1.5}, no external API), and reports move-marginal JSD, row-normalized transitions, stickiness, and template diversity. \emph{Part~C} embeds notes under two local encoders (BGE-Large primary, MPNet secondary) and reports MMD$^2$ \citep{gretton2012mmd}, Fr\'echet in PCA-$256$ \citep{heusel2017gans}, size-matched classifier AUC, and Kynk\"a\"anniemi-style precision/recall \citep{kynkaanniemi2019improved} plus a coverage variant.

\emph{Part~D} scores few-shot leakage and real-corpus phrase reuse against pre-specified token-run and Jaccard thresholds. \emph{Part~E} measures lexical fact recall; a synthetic-only LLM judge re-audits semantic realization and contradiction. \emph{Part~F} trains a primary-need classifier on (real, synthetic, direct-prompt-baseline) pools and evaluates on a by-client held-out real-note set, reporting macro-F1 and a project-defined \emph{transfer ratio}. \emph{Part~G} computes ROC-AUC of a nearest-neighbor membership predictor on (i)~$358$-d sparse multi-hot extracted-fact features and (ii)~$768$-d MPNet prose embeddings, with a project-defined $0.55$ follow-up threshold per channel. \emph{Practitioner evaluation} (\S\ref{sec:practitioner}) is an $N{=}8$ caseworker pilot (schema review, chronological-set review, pairwise comparison, three-note Turing task; hybrid deductive/inductive coding). Per-Part detail in Appendix~\ref{sec:appendix-evalframework}.

\section{Results}

\subsection{Overview}

The evaluation shows a mixed but useful pattern. Synthetic notes preserve broad documentation structure, avoid strong template collapse, and support practitioner-identified training and benchmarking uses, while remaining distinguishable from real data and covering only part of the real distribution ($4.3\%$ trajectory coverage versus $18.6\%$ for a same-size real control)---consistent with a bounded-use synthetic resource (Table~\ref{tab:headline}).

\begin{table*}[t]
\centering
\small
\setlength{\tabcolsep}{4pt}
\begin{tabular}{p{2.0cm}p{2.6cm}p{5.2cm}p{4.6cm}}
\toprule
\textbf{Component} & \textbf{Unit} & \textbf{Headline result} & \textbf{Interpretation} \\
\midrule
Part A & Client trajectories & Track~1 (multi): bal.\ 1-NN $0.764$, matched AUC $0.929$ [$0.878,\!0.970$]; Track~2 (single): bal.\ 1-NN $0.584$, AUC $0.573$ [$0.438,\!0.682$] & Distinguishable; coverage $23\%$/$83\%$ of same-size real reference (T1/T2) \\
Part B & Discourse moves & JSD $0.018$; mean stickiness $0.209{\rightarrow}0.136$; Heaps' $\beta_\text{real}{=}0.930$, $\beta_\text{synth}{=}0.920$ & Broad structure similar; synth less repetitive at move level \\
Part C & Note embeddings & BGE: MMD$^2{=}0.024$, AUC $0.966$; MPNet: MMD$^2{=}0.025$, AUC $0.938$ & Separable to classifiers; precision $86$--$98\%$ of real-real reference, recall $18$--$26\%$ \\
Part D & No-copy audit & $0$ above-threshold few-shot leaks; $13/210{,}039$ unique synth 10-grams overlap real ($6.2{\times}10^{-5}$); baseline $0/10{,}419$ & No above-threshold copying detected under audit \\
Part E & Fact-to-text & Lexical: $68.7\%$ concrete / $41.3\%$ soft; judge-labeled: $81.3\%$ grounded, $3.8\%$ contradicted & Concrete fields stronger; soft fields need caution \\
Part F & Downstream transfer & Macro-F1 ratio $0.83$ vs.\ $0.51$ baseline; $+0.050$ at matched $n{=}162$ & \textbf{Bounded utility quantified} \\
Part G & MIA sketch (privacy) & Fact AUC $0.48$, prose AUC $0.53$ (both $<0.55$) & No detectable membership signal under this nearest-neighbor attack \\
\bottomrule
\end{tabular}
\caption{Headline results across the six evaluation Parts plus the Part G privacy-side membership-inference sketch. Practitioner-pilot summary is in \S\ref{sec:practitioner}.}
\label{tab:headline}
\end{table*}

\subsection{Client scenario coverage}
\label{sec:partacoverage}

Part~A asks whether synthetic clients occupy plausible regions of the real trajectory space, and how much of that space they cover. For multi-note clients (Track~1; $n_\text{real}{=}1{,}331$ vs.\ $n_\text{synth}{=}54$) overlap is limited: $29.6\%$ of synthetic clients sit within the median real-to-real reference radius of a real client, but only $4.3\%$ of real clients are covered. A size-matched control quantifies how much of that gap is sample size rather than fidelity: a random $54$-client \emph{real} subsample covers $18.6\%$ [$13.6, 22.2$] of the remaining real clients under the same radius, so synthetic coverage is $23\%$ of that same-size real reference. Appendix~\ref{sec:appendix-v2} examines whether controlled sampling changes address multi-note undercoverage.

The corpora remain distinguishable (balanced 1-NN $0.764$; matched AUC $0.929$ [$0.878, 0.970$]). A cluster-level check is descriptive only (weak structure, silhouette $0.157$; ratios inside the project's $0.5\times$--$2.0\times$ band) and does not establish that no region is missing.

Track~2 (single-note; $n_\text{real}{=}350$ vs.\ $n_\text{synth}{=}33$) sits close to chance (matched AUC $0.573$ [$0.438, 0.682$]); its $26.3\%$ coverage is $83\%$ of the $31.7\%$ same-size real reference. Cluster ratios, the radius sweep, the control protocol, and the PCA projection are in Appendix~\ref{sec:appendix-coverage-discourse}.

\subsection{Documentation-structure similarity}
\label{sec:partbsummary}

Part~B asks whether synthetic notes are organized like real documentation at the discourse-move level. Across $369{,}679$ real and $16{,}151$ synthetic labeled sentences the move-marginal distributions are close (JSD $0.018$), though at these sample sizes most per-label differences remain statistically detectable ($11/12$ under Holm). The residual gaps are systematic in direction: synthetic notes over-use explicit single-purpose moves such as \texttt{resource\_referral} and under-use conversational ones such as \texttt{client\_response} (per-label deltas and transition summaries in Appendix~\ref{sec:appendix-coverage-discourse}). We adapt dialogue tactic stickiness \citep{zhan2026discoursediversity} to measure the probability that a move repeats in the next sentence. This probability \emph{decreases} from $0.209$ (real) to $0.136$ (synth): synthetic notes are less repetitive at the move level than real ones, the opposite of the pattern reported for dialogue LLMs. Template diversity does not collapse (top-$1$ template share $2.2\%$ vs.\ $2.8\%$ real; Heaps' $\beta$ $0.920$ vs.\ $0.930$). Full per-label and template tables are in Appendix~\ref{sec:appendix-coverage-discourse}.

\subsection{Embedding-space semantic fidelity}

Part~C tests whether $3{,}137$ synthetic and $62{,}901$ real notes are semantically close under sentence encoders---and whether they are nonetheless separable. Under BGE-Large, MMD$^2{=}0.024$ and precision/recall/coverage is $0.690/0.144/0.566$. Against a real-vs-real control ($0.806/0.800/0.971$; Appendix~\ref{sec:appendix-embeddings}), precision retains $86\%$ of the reference level while recall retains $18\%$, indicating stronger local-support retention than manifold coverage without implying indistinguishability. A size-matched classifier separates the corpora easily (AUC $0.966$), so distributional artifacts remain; practitioners in the Turing task (\S\ref{sec:practitioner}) scored $62.5\%$. MPNet replicates the pattern (MMD$^2{=}0.025$, AUC $0.938$; Appendix~\ref{sec:appendix-embeddings}).

\subsection{No-copy / overlap audit}
\label{sec:partd}

\paragraph{Part D---token-overlap audit.} Across the $3{,}137$ synthetic notes, \emph{no} synthetic note crossed the pre-specified few-shot leakage thresholds; the longest shared contiguous run between a v1.0 note and its conditioning exemplars is $8$ tokens, and only $13 / 210{,}039$ unique synthetic 10-grams overlap any 10-gram in the $63{,}142$-note real corpus ($6.2 \times 10^{-5}$), all confirmed domain boilerplate. The direct-prompt baseline corpus ($162$ notes, $10{,}419$ unique 10-grams) shows $0 / 10{,}419$ real-corpus 10-gram overlap on the same audit. Its 10-gram pool is $\sim\!20\times$ smaller, so the fraction is not directly comparable; neither pipeline shows above-threshold copying under this audit. Audit thresholds, the $13$-row overlap table, and per-note distributions are in Appendix~\ref{sec:appendix-nocopy}.

\paragraph{Part G---nearest-neighbor MIA sketch.} Prior attacks test membership in generator training data \citep{wang2022promptehr,kaneko2025sampling}. Our nearest-neighbor sketch instead asks whether the real notes used as few-shot exemplars sit systematically closer to the synthetic pool than other real notes. On a $358$-d structured-fact encoding, ROC-AUC is $0.48$ ($n_{\text{mem}}{=}126$ vs.\ $n_{\text{non-mem}}{=}61{,}645$); on MPNet prose embeddings it is $0.53$ ($n_{\text{mem}}{=}127$). Both remain near chance and below a project-defined $0.55$ follow-up threshold (Table~\ref{tab:mia}); a trajectory-uniqueness screen finds synthetic singling-out ($k{=}1$) rarer than the real-vs-real baseline (Appendix~\ref{sec:appendix-mia}). Parts~D and~G are leakage \emph{screens}, not a formal privacy guarantee; reconstruction and attribute inference remain future work (Appendix~\ref{sec:appendix-futurework}).

\begin{table}[t]
\centering
\small
\setlength{\tabcolsep}{3pt}
\begin{tabular}{lrr}
\toprule
\textbf{Metric} & \textbf{Fact} & \textbf{Prose} \\
\midrule
ROC-AUC & 0.475 & 0.535 \\
Threshold & 0.55 & 0.55 \\
Verdict & not detected & not detected \\
$n_{\text{mem}}$ & 126 & 127 \\
$n_{\text{non-mem}}$ & 61{,}645 & 62{,}774 \\
\bottomrule
\end{tabular}
\caption{Part G nearest-neighbor MIA on structured-fact and prose channels. Neither AUC exceeds the project-defined $0.55$ follow-up threshold; no membership signal is detected under this screen. Full protocol in Appendix~\ref{sec:appendix-mia}.}
\label{tab:mia}
\end{table}

\subsection{Fact-to-text alignment}
\label{sec:parte}

Part~E checks whether the prose actually surfaces the structured facts that conditioned it. For every populated field of a note's fact JSON we test whether any surface form of the value appears in the prose. Fields split into \emph{concrete} items (expected approximately verbatim) and \emph{soft / latent-shaper} items (which shape phrasing without appearing literally; listings in Appendix~\ref{sec:appendix-alignment}); the concrete bucket aligns with the practitioner-derived load-bearing fields (\S\ref{sec:practitioner}). The primary grounding number is concrete-field pooled recall \textbf{$68.7\%$} ($6{,}762/9{,}848$; per-field $55$--$82\%$); soft-field recall is $41.3\%$. Under its rubric, an automatic LLM judge labels $91.7\%$ of concrete pairs grounded and $3.8\%$ of populated pairs contradicted; without human-gold validation, these are diagnostic outputs, and prose claims outside the schema remain future work (limitations~M8--M9). Tables in Appendix~\ref{sec:appendix-alignment}.

\subsection{Downstream-task transfer (Part F)}
\label{sec:partf}

Part~F asks whether a classifier trained on the synthetic notes transfers to held-out real notes, the train-on-synth / eval-on-real reading established for synthetic clinical and biomedical text \citep{ive2020artificialmh,tang2023synthclinical}. The task is single-label primary-need classification; real labels are \emph{silver}, derived from the local extractor's \texttt{issues\_reported} field; synthetic labels come from conditioning facts or direct-prompt scenario seeds (implications flagged below). The head is logistic regression on cached BGE-Large embeddings; evaluation uses a $20\%$ by-client real holdout ($308$ clients, $6{,}254$ notes, $18$ labels). \textbf{Train-on-synth} reaches macro-F1 $0.438$, a transfer ratio of \textbf{$0.83$} against the real-on-real upper bound of $0.528$, vs.\ $0.51$ for the direct-prompt baseline and $0.05$ majority (Table~\ref{tab:transfer}). The low absolute upper bound reflects task hardness at $18$ imbalanced labels, so the ratio is the right summary; the direction replicates under an MLP head and under MPNet (ratio $0.82$).

Two qualifications bound this headline. First, the v1.0 training pool is ${\sim}12\times$ larger than the baseline pool, so a size-matched control retrains on $162$ stratified synth notes: the paired transfer-ratio difference over the same-$N$ baseline is \textbf{$+0.050$} (Wilcoxon $p{=}0.005$, $10/10$ outer seeds), a size-independent fidelity edge consistent with the size-insensitive Part~B/C gaps in \S\ref{sec:baseline-summary}. Second, the silver labels bound what the ratio means: real labels and the distributions used to sample synthetic conditioning labels share dependence on the local extractor, so the design may favor the synthetic condition rather than merely shifting the absolute number (limitation~M7); an author-coded gold subset ($n{=}250$, single coder) on the eval slice reads the gold-referenced ratio at $0.84$ (Appendix~\ref{sec:appendix-transfer}). Subgroup robustness, per-class breakdowns, and the sparse-label expansion ablation are in Appendix~\ref{sec:appendix-transfer}.

\begin{table}[t]
\centering
\small
\setlength{\tabcolsep}{2.5pt}
\begin{tabular}{lrrr}
\toprule
\textbf{Train pool} & \textbf{Macro-F1} & \textbf{Acc.} & \textbf{Ratio} \\
\midrule
Real ($n{=}30{,}216$)              & 0.528 & 0.674 & 1.00 \\
Synth (rel., $n{=}1{,}895$)        & \textbf{0.438} & 0.566 & \textbf{0.83} \\
\quad under MPNet$^{\ddagger}$     & 0.407 & 0.508 & 0.82 \\
Direct-prompt ($n{=}162$)          & 0.267 & 0.346 & 0.51 \\
Majority class                     & 0.026 & 0.312 & 0.05 \\
\midrule
\multicolumn{4}{l}{Matched size: ten real holdout splits} \\
Synth ($n{=}162$) & $0.295_{\pm.017}$ & 0.474 & $0.561_{\pm.032}$ \\
Direct ($n{=}162$) & $0.269_{\pm.010}$ & 0.359 & $0.511_{\pm.017}$ \\
\bottomrule
\end{tabular}
\caption{Part F primary-need transfer (BGE-Large logistic regression; $^{\ddagger}$MPNet). Upper rows use the fixed $308$-client, $6{,}254$-note holdout with $18$ labels. Lower rows use ten holdout splits: mean $\pm$ SD over $200$ synthetic fits or $10$ direct-prompt fits. Ratio = macro-F1 / real-on-real macro-F1 on the same split.}
\label{tab:transfer}
\end{table}

\subsection{Practitioner pilot}
\label{sec:practitioner}

Eight practicing social workers across eight service sectors each reviewed a fixed packet (schema review, two chronological note sets, two pairwise comparisons, a three-note Turing task; Appendix~\ref{sec:appendix-practitioner}). The protocol was fixed in advance, and we report the results descriptively: the pilot elicits judgments of schema adequacy and trajectory plausibility, intended uses, and failure modes from domain experts; it is not a population-level estimate and does not validate the automatic metrics (limitation~M9). Table~\ref{tab:pilot} summarizes the outcomes. Schema critiques targeted setting-specific over-specificity rather than missing fields. The seven-element ``useful-note'' structure participants converged on maps onto the Part~E concrete fields, and \emph{controlled imperfection} (abbreviations, dropped apostrophes) was repeatedly named as an authenticity cue: overly polished prose reads as synthetic. Use-case endorsements concentrate on training and evaluation uses rather than substitution for real records; none endorsed clinical use against real clients (attribution-tracked counts in Table~\ref{tab:usecases}, Appendix~\ref{sec:appendix-practitioner}).

\begin{table}[t]
\centering
\small
\setlength{\tabcolsep}{3pt}
\begin{tabular}{lr}
\toprule
\textbf{Pilot outcome ($N{=}8$)} & \textbf{Result} \\
\midrule
Schema judged adequate for own setting        & 8/8 \\
Preferred fact-rich Set~B over brief Set~A    & 7/8 \\
Endorsed training / onboarding use            & 8/8 \\
Endorsed clinical use against real clients    & \textbf{0/8} \\
Turing-task accuracy ($3$ notes $\times$ $8$) & 15/24 \\
\bottomrule
\end{tabular}
\caption{Practitioner-pilot headline outcomes (descriptive, $N{=}8$). Per-note Turing breakdown in Table~\ref{tab:turing}; use-case counts in Table~\ref{tab:usecases} (Appendix~\ref{sec:appendix-practitioner}).}
\label{tab:pilot}
\end{table}

\paragraph{Turing-task elicitation (descriptive).} The three-note Turing task is used descriptively to elicit which surface cues practitioners rely on to judge authenticity, not as a population-level indistinguishability claim. Aggregate accuracy was $15/24 = 62.5\%$ ($N=8$), inside the pre-specified $33$--$67\%$ indistinguishability window, but the value of the task lies in the per-note breakdown (Table~\ref{tab:turing}): the short synthetic note (73 words) was misclassified as real by $7/8$ participants, illustrating that the {``AI is polished''} authenticity heuristic fails on short clipped notes. Concrete defects and per-tier use-case attributions are in Appendix~\ref{sec:appendix-practitioner}.

\begin{table}[t]
\centering
\small
\begin{tabular}{lrrl}
\toprule
\textbf{Note} & \textbf{Words} & \textbf{Truth} & \textbf{Detected} \\
\midrule
1 & 152 & synth & 7/8 correct \\
2 & 134 & real  & 7/8 correct \\
3 &  73 & synth & 1/8 correct \\
\midrule
\textbf{Total} &---&---& \textbf{15/24 = 62.5\%} \\
\bottomrule
\end{tabular}
\caption{Per-note Turing breakdown ($N=8$). The short synthetic note (Note 3) was misclassified by $7/8$ practitioners; aggregate inside pre-specified $33$--$67\%$ indistinguishability window.}
\label{tab:turing}
\end{table}

\section{Discussion}
\label{sec:discussion}

The central result is not indistinguishability---trajectories are distinguishable and note embeddings are separable under both encoders---but \emph{downstream-task transfer} (Part~F, Table~\ref{tab:headline}), with a size-matched ratio edge over the same generator's direct-prompt baseline (Table~\ref{tab:transfer}). This is the most direct evidence that the fixed corpus is useful for the bounded research, training, and benchmarking uses we claim. Parts~A--C bound where that utility comes from: partial trajectory coverage and preserved documentation structure without collapse. Structured-first conditioning anchors each note to a corpus-aligned fact JSON; Parts~B and~C measure the resulting corpus-level structure and variation. \citet{knowledgesg2024} take a client--server approach to sensitive-domain transfer; ours enforces an analogous split structurally (local learning and sampling; external realization from synthetic facts plus audited redacted exemplars). The practitioner-utility reading, sparse-class behavior, and the concrete/soft field split are discussed further in Appendix~\ref{sec:appendix-discussion}.

A separate follow-up found improved structured coverage without consistent downstream gains (Appendix~\ref{sec:appendix-v2}), underscoring the need to evaluate these outcomes separately.

\paragraph{Why structured-first outperforms direct prompting.} Both pipelines call the same \texttt{gpt-5.2} model under the same documentation-style instruction, though the structured pipeline additionally receives redacted style exemplars (the ablation-scope caveat in \S\ref{sec:baseline-summary}). The comparison therefore shows that the \emph{complete} structured-first stack outperforms this sparse direct-prompt baseline; it does not isolate any single component as causal.

\section{Conclusion}
\label{sec:conclusion}

Longitudinal case notes are difficult to release. We introduced a $3{,}137$-note free-text synthetic corpus, a structured-first pipeline with an audited local/API boundary (\S\ref{sec:source-constraints}), a six-Part utility framework, and a Part~G privacy-side MIA sketch. The pipeline outperforms a same-model direct-prompt baseline on size-insensitive fidelity metrics, achieves higher transfer performance on held-out real notes (Table~\ref{tab:headline}), and does not trigger the Part~G follow-up threshold on either channel (Table~\ref{tab:mia}). The practitioner pilot (\S\ref{sec:practitioner}) supports a bounded-utility reading: appropriate for training, prototyping, and benchmarking, not a substitute for real records.

\section*{Acknowledgments}

This research was supported in part by National Science Foundation grant DGE-2125858 and by Good Systems, a UT Austin Grand Challenge for developing responsible AI technologies.\footnote{\url{https://goodsystems.utexas.edu}} The Texas Advanced Computing Center (TACC) at The University of Texas at Austin provided computational resources that contributed to the results reported in this paper.\footnote{\url{http://www.tacc.utexas.edu}}

We thank Amy Pavel for early-stage advising, a US-based agency for access to the source corpus under a data-use agreement, and the eight practitioner participants for their time and expertise. This work forms part of Hender Lin's graduate research at The University of Texas at Austin.

The authors used AI assistants, including OpenAI Codex, Cursor, and GitHub Copilot, to support code development, literature search, documentation, and manuscript drafting and editing. All assistance was guided by the authors, who reviewed and verified the resulting content, citations, code, and reported results and retain full responsibility for the final work. Models used for data generation and evaluation are documented separately in the methods and reproducibility appendix.

\label{sec:content-end}

\section*{Limitations}

We catalog limitations by category; per-item enumeration (M1--M9, C1--C4, P1--P5, Pr1--Pr2 including Pr1b, G1--G2 including G1b, H1, and Cov1) with mitigations is in Appendix~\ref{sec:appendix-limitations}.

\paragraph{Methodological (M1--M9).} Part~C and Part~F metrics are encoder-specific, mitigated by reporting both BGE-Large and MPNet (the dual-encoder Part~F robustness numbers are in Appendix~\ref{sec:appendix-transfer}); Part~B uses an embedding-prototype zero-shot labeler (mean cosine margin $0.027$). Extraction uses a single LLM (\texttt{qwen3:32b-q8\_0}); the Part~A cluster structure is weak ($k{=}4$ silhouette $0.157$); and fields like \texttt{issue\_severity} record a documented judgment rather than observer-independent ground truth (M6). Part~F real labels come from the extractor, while synthetic labels come from facts sampled using distributions learned from its outputs; this shared dependence bounds the headline ratio (M7); an author-coded $n{=}250$ gold subset measures silver--gold agreement at $0.69$ and a gold-referenced transfer ratio of $0.84$ (Appendix~\ref{sec:appendix-transfer}). Part~E's judge checks populated fields but not unsupported claims outside the schema (M8), and neither Part~B nor Part~E is validated against a human-coded gold subset on this corpus (M9): the practitioner pilot (\S\ref{sec:practitioner}) evaluates utility, not metric validity.

\paragraph{Corpus (C1--C4).} Distributional claims are scoped to one US-based agency (English, 2016--2022). The synthetic prose corpus is $\sim 5\%$ the size of the real corpus ($3{,}137/62{,}901$ notes; $87/1{,}681$ clients). Per-issue persistence treats an issue missing from the next note as non-persistence, so under-extracted issues read as ``quickly resolving.''

\paragraph{Pipeline (P1--P5).} Caseworker identity is not modeled; the email-thread sub-genre ($\sim 9\%$ of real) is not generated; intra-note lumpiness, typos, numeric detail, and within-client style drift on long trajectories are not addressable by a single calibration knob.

\paragraph{Privacy (Pr1--Pr2).} The NER+regex redactor is high-precision-low-recall (Pr1), and the few-shot style exemplars sent to the external synthesis API are PII-redacted \emph{real} notes (Pr2)---so a residual identifier in that exemplar pool can reach the API (API-exposure risk), while Parts~D and~G audit release-side copying and membership rather than residual identifiers in the prompts themselves. A manual residual-identifier review of all $146$ exemplars used for the fixed corpus found $0$ residual PII. Defense-in-depth is the small rotated private pool, the Part~D no-copy audit of every v1.0 note against exactly the exemplars that conditioned it (\S\ref{sec:partd}; Table~\ref{tab:overlap-boilerplate}), and the Part~G MIA sketch below the project-defined follow-up threshold on both channels (Table~\ref{tab:mia}; Appendix~\ref{sec:appendix-mia}). Unredacted source notes, identifier records, and extracted real trajectories remain local; the exemplars retain the residual risk above. The redactor targets named-entity surface forms only; legal/immigration/health/identity sensitivity remains after PII removal (Pr1b) and is not subject to per-attribute controls.

\paragraph{Generalization (G1--G2).} The corpus, schema, and prompts are English and US-centric; all eight practitioners endorsed the fact schema as comprehensive but named distinct setting-specific extensions, so the schema is a broad superset that requires customization for a new setting. The pipeline has not been run end-to-end on a second corpus.

\paragraph{Practitioner evaluation (H1) and what is not in this paper.} The practitioner pilot (\S\ref{sec:practitioner}) is descriptive and recruitment skews toward source-organization adjacency. Demographic-fairness disaggregation (race / gender / age) is not reported because the source \texttt{client\_info.csv} does not record these attributes; we instead report a project-observable subgroup-robustness audit by trajectory length, severity, and primary-issue category (Appendix~\ref{sec:appendix-transfer}). \citet{ducel2025gender} document that LMs over-generate male patient cases under unspecified demographic prompts, so absence of demographic conditioning is not the same as absence of demographic bias; a downstream-corpus demographic audit is future work. We also do \emph{not} report full attribute-inference attacks against the generator \citep{wang2022promptehr} or higher-order time-axis prose dynamics \citep{cbtarcs2025,petherapy2025}.

\section*{Ethical Considerations}

The release excludes raw notes, post-redaction source notes, real extracted facts, real client identifiers, real-redacted few-shot examples, private logs, and internal git history; extraction and evaluation involving real text run locally, and the external synthesis model (\texttt{gpt-5.2}, also used for the direct-prompt baseline) receives synthetic structured facts plus a rotated pool of PII-redacted style exemplars that never appear in the release and are screened for copying (\S\ref{sec:partd}). The Part~E LLM-judge audit sends only synthetic fact--note pairs---data that already crossed the API boundary at generation time---to a second external model (\texttt{gpt-5.6-terra}); no real-derived text is involved. The separate common-prose diagnostic in Appendix~\ref{sec:v2-downstream} used an additional OpenAI model and an audited training-only redacted exemplar pool under the same local-data boundary.

Following \citet{stadler2022syntheticnotsilver}, our defense-in-depth (locality boundary + Part~D no-copy audit + Part~G MIA sketch in Appendix~\ref{sec:appendix-mia}) mitigates risk but provides no differential-privacy guarantee. Part~D (Table~\ref{tab:overlap-boilerplate}) finds no above-threshold few-shot leaks and only boilerplate $n$-gram overlap on either the pipeline or the direct-prompt baseline; Part~G AUCs (Table~\ref{tab:mia}) sit below the project-defined follow-up threshold on both channels. Full attribute-inference attacks \citep{wang2022promptehr} remain future work, and a residual-risk catalog is in Appendix~\ref{sec:riskcatalog}. The corpus is intended for research, education, and benchmarking in privacy-sensitive NLP; notes do not describe real clients, should not be used for decisions about real people or as representative evidence about the source population, and require validation on real-world data before any operational use.

\bibliography{custom}

\begin{thebibliography}{46}
\providecommand{\natexlab}[1]{#1}

\bibitem[{Belkadi et~al.(2025)Belkadi, Ren, Micheletti, Han, and
  Nenadic}]{belkadi2025synthfreetext}
Samuel Belkadi, Libo Ren, Nicolo Micheletti, Lifeng Han, and Goran Nenadic.
  2025.
\newblock \href {https://doi.org/10.18653/v1/2025.naacl-srw.20} {Generating
  synthetic free-text medical records with low re-identification risk using
  masked language modeling}.
\newblock In \emph{Proceedings of the 2025 Conference of the Nations of the
  Americas Chapter of the Association for Computational Linguistics: Human
  Language Technologies (Volume 4: Student Research Workshop)}, pages 200--206,
  Albuquerque, USA. Association for Computational Linguistics.

\bibitem[{Bender and Friedman(2018)}]{bender2018data}
Emily~M. Bender and Batya Friedman. 2018.
\newblock \href {https://doi.org/10.1162/tacl_a_00041} {Data statements for
  natural language processing: {Toward} mitigating system bias and enabling
  better science}.
\newblock \emph{Transactions of the Association for Computational Linguistics},
  6:587--604.

\bibitem[{Bn et~al.(2025)Bn, Mattioli, Sherrill, Arriaga, Wiese, and
  Abdullah}]{petherapy2025}
Suhas Bn, Dominik~O. Mattioli, Andrew~M. Sherrill, Rosa~I. Arriaga, Christopher
  Wiese, and Saeed Abdullah. 2025.
\newblock \href {https://doi.org/10.18653/v1/2025.findings-emnlp.1144} {How
  real are synthetic therapy conversations? evaluating fidelity in prolonged
  exposure dialogues}.
\newblock In \emph{Findings of the Association for Computational Linguistics:
  EMNLP 2025}, pages 20986--20995, Suzhou, China. Association for Computational
  Linguistics.

\bibitem[{Chen et~al.(2025)Chen, Han, and Miyao}]{chen2025revisitmia}
Bowen Chen, Namgi Han, and Yusuke Miyao. 2025.
\newblock \href {https://doi.org/10.18653/v1/2025.acl-long.1114} {A statistical
  and multi-perspective revisiting of the membership inference attack in large
  language models}.
\newblock In \emph{Proceedings of the 63rd Annual Meeting of the Association
  for Computational Linguistics (Volume 1: Long Papers)}, pages 22854--22874,
  Vienna, Austria. Association for Computational Linguistics.

\bibitem[{Dou et~al.(2026)Dou, Mamut, and Xu}]{dou2026gavel}
Yao Dou, Benjamin Mamut, and Wei Xu. 2026.
\newblock \href {https://doi.org/10.18653/2026.emnlp-main.676} {{Gavel}: Agent
  meets checklist for evaluating {LLMs} on long-context legal summarization}.
\newblock In \emph{Proceedings of the 2026 Conference on Empirical Methods in
  Natural Language Processing}, pages 15553--15603, Budapest, Hungary.
  Association for Computational Linguistics.

\bibitem[{Ducel et~al.(2025)Ducel, Hiebel, Ferret, Fort, and
  N{\'e}v{\'e}ol}]{ducel2025gender}
Fanny Ducel, Nicolas Hiebel, Olivier Ferret, Kar{\"e}n Fort, and Aur{\'e}lie
  N{\'e}v{\'e}ol. 2025.
\newblock \href {https://doi.org/10.18653/v1/2025.findings-naacl.398}
  {``{W}omen do not have heart attacks!'' {G}ender biases in automatically
  generated clinical cases in {F}rench}.
\newblock In \emph{Findings of the Association for Computational Linguistics:
  {NAACL} 2025}, pages 7160--7174, Albuquerque, New Mexico. Association for
  Computational Linguistics.

\bibitem[{Gebru et~al.(2021)Gebru, Morgenstern, Vecchione, Vaughan, Wallach,
  Daum{\'e}~III, and Crawford}]{gebru2021datasheets}
Timnit Gebru, Jamie Morgenstern, Briana Vecchione, Jennifer~Wortman Vaughan,
  Hanna Wallach, Hal Daum{\'e}~III, and Kate Crawford. 2021.
\newblock \href {https://doi.org/10.1145/3458723} {Datasheets for datasets}.
\newblock \emph{Communications of the ACM}, 64(12):86--92.

\bibitem[{Gondimalla et~al.(2024)Gondimalla, Sreekanth, Joshi, Nelson, Choi,
  Slota, Greenberg, Fleischmann, and Lee}]{gondimalla2024aligning}
Apoorva Gondimalla, Varshinee Sreekanth, Govind Joshi, Whitney Nelson, Eunsol
  Choi, Stephen~C. Slota, Sherri~R. Greenberg, Kenneth~R. Fleischmann, and
  Min~Kyung Lee. 2024.
\newblock \href {https://doi.org/10.1145/3613904.3642014} {Aligning data with
  the goals of an organization and its workers: Designing data labeling for
  social service case notes}.
\newblock In \emph{Proceedings of the 2024 {CHI} Conference on Human Factors in
  Computing Systems}, CHI '24. Association for Computing Machinery.

\bibitem[{Gretton et~al.(2012)Gretton, Borgwardt, Rasch, Sch{\"o}lkopf, and
  Smola}]{gretton2012mmd}
Arthur Gretton, Karsten~M. Borgwardt, Malte~J. Rasch, Bernhard Sch{\"o}lkopf,
  and Alexander Smola. 2012.
\newblock \href {https://jmlr.org/papers/v13/gretton12a.html} {A kernel
  two-sample test}.
\newblock \emph{Journal of Machine Learning Research}, 13(25):723--773.

\bibitem[{Heusel et~al.(2017)Heusel, Ramsauer, Unterthiner, Nessler, and
  Hochreiter}]{heusel2017gans}
Martin Heusel, Hubert Ramsauer, Thomas Unterthiner, Bernhard Nessler, and Sepp
  Hochreiter. 2017.
\newblock \href
  {https://proceedings.neurips.cc/paper/2017/hash/8a1d694707eb0fefe65871369074926d-Abstract.html}
  {{GANs} trained by a two time-scale update rule converge to a local {Nash}
  equilibrium}.
\newblock In \emph{Advances in Neural Information Processing Systems},
  volume~30.

\bibitem[{Ive et~al.(2020)Ive, Viani, Kam, Yin, Verma, Puntis, Cardinal,
  Roberts, Stewart, and Velupillai}]{ive2020artificialmh}
Julia Ive, Natalia Viani, Joyce Kam, Lucia Yin, Somain Verma, Stephen Puntis,
  Rudolf~N. Cardinal, Angus Roberts, Robert Stewart, and Sumithra Velupillai.
  2020.
\newblock \href {https://doi.org/10.1038/s41746-020-0267-x} {Generation and
  evaluation of artificial mental health records for natural language
  processing}.
\newblock \emph{npj Digital Medicine}, 3(1):69.

\bibitem[{Johnson et~al.(2016)Johnson, Pollard, Shen, Lehman, Feng, Ghassemi,
  Moody, Szolovits, Celi, and Mark}]{johnson2016mimic}
Alistair E.~W. Johnson, Tom~J. Pollard, Lu~Shen, Li-wei~H. Lehman, Mengling
  Feng, Mohammad Ghassemi, Benjamin Moody, Peter Szolovits, Leo~Anthony Celi,
  and Roger~G. Mark. 2016.
\newblock \href {https://doi.org/10.1038/sdata.2016.35} {{MIMIC-III}, a freely
  accessible critical care database}.
\newblock \emph{Scientific Data}, 3(1):160035.

\bibitem[{Jordon et~al.(2022)Jordon, Szpruch, Houssiau, Bottarelli, Cherubin,
  Maple, Cohen, and Weller}]{royalsociety2022synth}
James Jordon, Lukasz Szpruch, Florimond Houssiau, Mirko Bottarelli, Giovanni
  Cherubin, Carsten Maple, Samuel~N. Cohen, and Adrian Weller. 2022.
\newblock \href
  {https://royalsociety.org/-/media/policy/projects/privacy-enhancing-technologies/Synthetic_Data_Survey-24.pdf}
  {Synthetic data -- what, why and how?}
\newblock {The Royal Society} explainer report.

\bibitem[{Kaneko et~al.(2025)Kaneko, Ma, Wata, and
  Okazaki}]{kaneko2025sampling}
Masahiro Kaneko, Youmi Ma, Yuki Wata, and Naoaki Okazaki. 2025.
\newblock \href {https://doi.org/10.18653/v1/2025.findings-acl.465}
  {Sampling-based pseudo-likelihood for membership inference attacks}.
\newblock In \emph{Findings of the Association for Computational Linguistics:
  ACL 2025}, pages 8894--8907, Vienna, Austria. Association for Computational
  Linguistics.

\bibitem[{Kang et~al.(2025)Kang, Ardalani, Kuchnik, Emad, Elhoushi, Sengupta,
  Li, Raghavendra, Jia, and Wu}]{demystify2025}
Feiyang Kang, Newsha Ardalani, Michael Kuchnik, Youssef Emad, Mostafa Elhoushi,
  Shubhabrata Sengupta, Shang-Wen Li, Ramya Raghavendra, Ruoxi Jia, and
  Carole-Jean Wu. 2025.
\newblock \href {https://doi.org/10.18653/v1/2025.emnlp-main.544} {Demystifying
  synthetic data in {LLM} pre-training: A systematic study of scaling laws,
  benefits, and pitfalls}.
\newblock In \emph{Proceedings of the 2025 Conference on Empirical Methods in
  Natural Language Processing}, pages 10739--10758, Suzhou, China. Association
  for Computational Linguistics.

\bibitem[{Kim et~al.(2024)Kim, Hahm, and Lee}]{kim2024generalizing}
Woojin Kim, Sungeun Hahm, and Jaejin Lee. 2024.
\newblock \href {https://doi.org/10.18653/v1/2024.emnlp-main.1181}
  {Generalizing clinical de-identification models by privacy-safe data
  augmentation using {GPT}-4}.
\newblock In \emph{Proceedings of the 2024 Conference on Empirical Methods in
  Natural Language Processing}, pages 21204--21218, Miami, Florida, USA.
  Association for Computational Linguistics.

\bibitem[{Kruse et~al.(2025)Kruse, Hu, Derby, Wu, Stonbraker, Yao, Wang,
  Goldberg, and Gao}]{kruse2025longitudinal}
Maya Kruse, Shiyue Hu, Nicholas Derby, Yifu Wu, Samantha Stonbraker, Bingsheng
  Yao, Dakuo Wang, Elizabeth~M. Goldberg, and Yanjun Gao. 2025.
\newblock \href {https://doi.org/10.18653/v1/2025.findings-emnlp.1128} {Large
  language models with temporal reasoning for longitudinal clinical
  summarization and prediction}.
\newblock In \emph{Findings of the Association for Computational Linguistics:
  EMNLP 2025}, pages 20715--20735, Suzhou, China. Association for Computational
  Linguistics.

\bibitem[{Kynk{\"a}{\"a}nniemi et~al.(2019)Kynk{\"a}{\"a}nniemi, Karras, Laine,
  Lehtinen, and Aila}]{kynkaanniemi2019improved}
Tuomas Kynk{\"a}{\"a}nniemi, Tero Karras, Samuli Laine, Jaakko Lehtinen, and
  Timo Aila. 2019.
\newblock \href
  {https://proceedings.neurips.cc/paper/2019/hash/0234c510bc6d908b28c70ff313743079-Abstract.html}
  {Improved precision and recall metric for assessing generative models}.
\newblock In \emph{Advances in Neural Information Processing Systems},
  volume~32.

\bibitem[{Li et~al.(2023{\natexlab{a}})Li, Wang, and Yu}]{li2023twodirections}
Rumeng Li, Xun Wang, and Hong Yu. 2023{\natexlab{a}}.
\newblock \href {https://doi.org/10.18653/v1/2023.findings-emnlp.474} {Two
  directions for clinical data generation with large language models:
  Data-to-label and label-to-data}.
\newblock In \emph{Findings of the Association for Computational Linguistics:
  EMNLP 2023}, pages 7129--7143, Singapore. Association for Computational
  Linguistics.

\bibitem[{Li et~al.(2023{\natexlab{b}})Li, Zhu, Lu, and Yin}]{li2023synthetic}
Zhuoyan Li, Hangxiao Zhu, Zhuoran Lu, and Ming Yin. 2023{\natexlab{b}}.
\newblock \href {https://doi.org/10.18653/v1/2023.emnlp-main.647} {Synthetic
  data generation with large language models for text classification: Potential
  and limitations}.
\newblock In \emph{Proceedings of the 2023 Conference on Empirical Methods in
  Natural Language Processing}, pages 10443--10461, Singapore. Association for
  Computational Linguistics.

\bibitem[{Long et~al.(2024)Long, Wang, Xiao, Zhao, Ding, Chen, and
  Wang}]{long2024llmsynthsurvey}
Lin Long, Rui Wang, Ruixuan Xiao, Junbo Zhao, Xiao Ding, Gang Chen, and Haobo
  Wang. 2024.
\newblock \href {https://doi.org/10.18653/v1/2024.findings-acl.658} {On
  {LLM}s-driven synthetic data generation, curation, and evaluation: A survey}.
\newblock In \emph{Findings of the Association for Computational Linguistics:
  ACL 2024}, pages 11065--11082, Bangkok, Thailand. Association for
  Computational Linguistics.

\bibitem[{Loni et~al.(2025)Loni, Poursalim, Asadi, and
  Gharehbaghi}]{loni2025synthreview}
Mohammad Loni, Fatemeh Poursalim, Mehdi Asadi, and Arash Gharehbaghi. 2025.
\newblock \href {https://doi.org/10.1038/s41746-024-01409-w} {A review on
  generative {AI} models for synthetic medical text, time series, and
  longitudinal data}.
\newblock \emph{npj Digital Medicine}, 8:281.

\bibitem[{Lopez-Paz and Oquab(2017)}]{lopezpaz2017classifier}
David Lopez-Paz and Maxime Oquab. 2017.
\newblock \href {https://openreview.net/forum?id=SJkXfE5xx} {Revisiting
  classifier two-sample tests}.
\newblock In \emph{International Conference on Learning Representations}.

\bibitem[{Meeus et~al.(2025)Meeus, Wutschitz, Zanella-B{\'e}guelin, Tople, and
  Shokri}]{meeus2025canary}
Matthieu Meeus, Lukas Wutschitz, Santiago Zanella-B{\'e}guelin, Shruti Tople,
  and Reza Shokri. 2025.
\newblock \href {https://proceedings.mlr.press/v267/meeus25a.html} {The
  canary's echo: Auditing privacy risks of {LLM}-generated synthetic text}.
\newblock In \emph{Proceedings of the 42nd International Conference on Machine
  Learning}, volume 267 of \emph{Proceedings of Machine Learning Research},
  pages 43557--43580. PMLR.

\bibitem[{Meoni et~al.(2025)Meoni, Ryffel, and
  De~La~Clergerie}]{meoni2025synthdocs}
Simon Meoni, Th{\'e}o Ryffel, and {\'E}ric De~La~Clergerie. 2025.
\newblock \href {https://doi.org/10.18653/v1/2025.cl4health-1.2} {Synthetic
  documents for medical tasks: Bridging privacy with knowledge injection and
  reward mechanism}.
\newblock In \emph{Proceedings of the Second Workshop on Patient-Oriented
  Language Processing ({CL4Health})}, pages 12--25, Albuquerque, New Mexico.
  Association for Computational Linguistics.

\bibitem[{Mishra et~al.(2024)Mishra, Yao, Vashisht, Ouyang, Wang, Mody, and
  Yu}]{mishra2024synfacedit}
Prakamya Mishra, Zonghai Yao, Parth Vashisht, Feiyun Ouyang, Beining Wang,
  Vidhi~Dhaval Mody, and Hong Yu. 2024.
\newblock \href {https://doi.org/10.18653/v1/2024.emnlp-main.1120}
  {{SYNFAC}-{EDIT}: Synthetic imitation edit feedback for factual alignment in
  clinical summarization}.
\newblock In \emph{Proceedings of the 2024 Conference on Empirical Methods in
  Natural Language Processing}, pages 20061--20083, Miami, Florida, USA.
  Association for Computational Linguistics.

\bibitem[{Mitchell et~al.(2019)Mitchell, Wu, Zaldivar, Barnes, Vasserman,
  Hutchinson, Spitzer, Raji, and Gebru}]{mitchell2019modelcards}
Margaret Mitchell, Simone Wu, Andrew Zaldivar, Parker Barnes, Lucy Vasserman,
  Ben Hutchinson, Elena Spitzer, Inioluwa~Deborah Raji, and Timnit Gebru. 2019.
\newblock \href {https://doi.org/10.1145/3287560.3287596} {Model cards for
  model reporting}.
\newblock In \emph{Proceedings of the Conference on Fairness, Accountability,
  and Transparency}, pages 220--229.

\bibitem[{Mitra et~al.(2025)Mitra, Yang, Druhl, Goodwin, and
  Yu}]{mitra2025synthsbdh}
Avijit Mitra, Zhichao Yang, Emily Druhl, Raelene Goodwin, and Hong Yu. 2025.
\newblock \href {https://doi.org/10.18653/v1/2025.emnlp-main.1418}
  {{Synth-SBDH}: A synthetic dataset of social and behavioral determinants of
  health for clinical text}.
\newblock In \emph{Proceedings of the 2025 Conference on Empirical Methods in
  Natural Language Processing}, pages 27899--27935, Suzhou, China. Association
  for Computational Linguistics.

\bibitem[{Namuduri et~al.(2025)Namuduri, Wu, Zheng, Wadhwa, Durrett, and
  Li}]{namuduri2025qudsim}
Ramya Namuduri, Yating Wu, Anshun~Asher Zheng, Manya Wadhwa, Greg Durrett, and
  Junyi~Jessy Li. 2025.
\newblock \href {https://openreview.net/forum?id=zFz1BJu211} {{QUDsim}:
  Quantifying discourse similarities in {LLM}-generated text}.
\newblock In \emph{Proceedings of the Conference on Language Modeling
  ({COLM})}.

\bibitem[{Nelson et~al.(2024)Nelson, Wang, Choi, and
  Lee}]{nelson2024homelessness}
Whitney Nelson, Victor Wang, Eunsol Choi, and Min~Kyung Lee. 2024.
\newblock \href {https://openreview.net/forum?id=XyLgLwT5wJ} {Designing
  {LLM}-based support for homelessness caseworkers}.
\newblock In \emph{Proceedings of the {AAAI} 2024 Workshop on Public Sector
  {LLMs} ({PubLLM})}.

\bibitem[{Ochs and Habernal(2025)}]{ochs2025diffusion}
Sebastian Ochs and Ivan Habernal. 2025.
\newblock \href {https://doi.org/10.18653/v1/2025.naacl-long.532} {Private
  synthetic text generation with diffusion models}.
\newblock In \emph{Proceedings of the 2025 Conference of the Nations of the
  Americas Chapter of the Association for Computational Linguistics: Human
  Language Technologies (Volume 1: Long Papers)}, pages 10612--10626,
  Albuquerque, New Mexico. Association for Computational Linguistics.

\bibitem[{Pickens et~al.(2023)Pickens, Andre, and
  Morrison}]{pickens2023synthhumanservices}
Madeline Pickens, Jennifer Andre, and Gabriel Morrison. 2023.
\newblock \href
  {https://www.urban.org/research/publication/generating-fully-synthetic-human-services-dataset}
  {Generating a fully synthetic human services dataset}.
\newblock Technical report, Urban Institute.

\bibitem[{Puerto et~al.(2025)Puerto, Gubri, Yun, and Oh}]{puerto2025scaling}
Haritz Puerto, Martin Gubri, Sangdoo Yun, and Seong~Joon Oh. 2025.
\newblock \href {https://doi.org/10.18653/v1/2025.findings-naacl.234} {Scaling
  up membership inference: When and how attacks succeed on large language
  models}.
\newblock In \emph{Findings of the Association for Computational Linguistics:
  NAACL 2025}, pages 4165--4182, Albuquerque, New Mexico. Association for
  Computational Linguistics.

\bibitem[{Ramesh et~al.(2024)Ramesh, Gandhi, Madaan, Bauer, Peris, and
  Field}]{ramesh2024dpeval}
Krithika Ramesh, Nupoor Gandhi, Pulkit Madaan, Lisa Bauer, Charith Peris, and
  Anjalie Field. 2024.
\newblock \href {https://doi.org/10.18653/v1/2024.findings-emnlp.894}
  {Evaluating differentially private synthetic data generation in high-stakes
  domains}.
\newblock In \emph{Findings of the Association for Computational Linguistics:
  EMNLP 2024}, pages 15254--15269, Miami, Florida, USA. Association for
  Computational Linguistics.

\bibitem[{Ramesh et~al.(2025)Ramesh, Smolyak, Zhao, Gandhi, Agarwal,
  Bjarnad{\'o}ttir, and Field}]{ramesh2025synthtexteval}
Krithika Ramesh, Daniel Smolyak, Zihao Zhao, Nupoor Gandhi, Ritu Agarwal,
  Margr{\'e}t~V. Bjarnad{\'o}ttir, and Anjalie Field. 2025.
\newblock \href {https://doi.org/10.18653/v1/2025.emnlp-demos.35}
  {{SynthTextEval}: Synthetic text data generation and evaluation for
  high-stakes domains}.
\newblock In \emph{Proceedings of the 2025 Conference on Empirical Methods in
  Natural Language Processing: System Demonstrations}, pages 487--499, Suzhou,
  China. Association for Computational Linguistics.

\bibitem[{Stadler et~al.(2022)Stadler, Oprisanu, and
  Troncoso}]{stadler2022syntheticnotsilver}
Theresa Stadler, Bristena Oprisanu, and Carmela Troncoso. 2022.
\newblock \href
  {https://www.usenix.org/conference/usenixsecurity22/presentation/stadler}
  {Synthetic data -- anonymisation groundhog day}.
\newblock In \emph{Proceedings of the 31st {USENIX} Security Symposium}, pages
  1451--1468. USENIX Association.

\bibitem[{Stubbs and Uzuner(2015)}]{stubbs2015i2b2}
Amber Stubbs and {\"O}zlem Uzuner. 2015.
\newblock \href {https://doi.org/10.1016/j.jbi.2015.07.020} {Annotating
  longitudinal clinical narratives for de-identification: {The} 2014
  {i2b2/UTHealth} corpus}.
\newblock \emph{Journal of Biomedical Informatics}, 58:S20--S29.

\bibitem[{Tang et~al.(2023)Tang, Han, Jiang, and Hu}]{tang2023synthclinical}
Ruixiang Tang, Xiaotian Han, Xiaoqian Jiang, and Xia Hu. 2023.
\newblock \href {https://arxiv.org/abs/2303.04360} {Does synthetic data
  generation of {LLMs} help clinical text mining?}
\newblock \emph{Preprint}, arXiv:2303.04360.

\bibitem[{Uzuner et~al.(2011)Uzuner, South, Shen, and DuVall}]{uzuner2011i2b2}
{\"O}zlem Uzuner, Brett~R. South, Shuying Shen, and Scott~L. DuVall. 2011.
\newblock \href {https://doi.org/10.1136/amiajnl-2011-000203} {2010 {i2b2/VA}
  challenge on concepts, assertions, and relations in clinical text}.
\newblock \emph{Journal of the American Medical Informatics Association},
  18(5):552--556.

\bibitem[{Wang et~al.(2025{\natexlab{a}})Wang, Gao, Xu, Liu, Hussein,
  Korsapati, El~Labban, Iheasirim, Hassan, Anil, Bartlett, and
  Sun}]{wang2025processrm}
Hanyin Wang, Chufan Gao, Qiping Xu, Bolun Liu, Guleid Hussein, Hariprasad~Reddy
  Korsapati, Mohamad El~Labban, Kingsley Iheasirim, Mohamed Hassan, Gokhan
  Anil, Brian Bartlett, and Jimeng Sun. 2025{\natexlab{a}}.
\newblock \href {https://doi.org/10.18653/v1/2025.emnlp-main.967}
  {Process-supervised reward models for verifying clinical note generation: A
  scalable approach guided by domain expertise}.
\newblock In \emph{Proceedings of the 2025 Conference on Empirical Methods in
  Natural Language Processing}, pages 19127--19147, Suzhou, China. Association
  for Computational Linguistics.

\bibitem[{Wang et~al.(2024)Wang, Liang, Ye, Chai, Chen, and
  Wang}]{knowledgesg2024}
Wenhao Wang, Xiaoyu Liang, Rui Ye, Jingyi Chai, Siheng Chen, and Yanfeng Wang.
  2024.
\newblock \href {https://doi.org/10.18653/v1/2024.emnlp-main.438}
  {{KnowledgeSG}: Privacy-preserving synthetic text generation with knowledge
  distillation from server}.
\newblock In \emph{Proceedings of the 2024 Conference on Empirical Methods in
  Natural Language Processing}, pages 7677--7695, Miami, Florida, USA.
  Association for Computational Linguistics.

\bibitem[{Wang et~al.(2025{\natexlab{b}})Wang, Zhang, Zhang, and
  Guo}]{cbtarcs2025}
Xiaoyi Wang, Jiwei Zhang, Guangtao Zhang, and Honglei Guo. 2025{\natexlab{b}}.
\newblock \href {https://doi.org/10.18653/v1/2025.findings-emnlp.1089} {Feel
  the difference? a comparative analysis of emotional arcs in real and
  {LLM}-generated {CBT} sessions}.
\newblock In \emph{Findings of the Association for Computational Linguistics:
  EMNLP 2025}, pages 19999--20017, Suzhou, China. Association for Computational
  Linguistics.

\bibitem[{Wang and Sun(2022)}]{wang2022promptehr}
Zifeng Wang and Jimeng Sun. 2022.
\newblock \href {https://doi.org/10.18653/v1/2022.emnlp-main.185} {{PromptEHR}:
  Conditional electronic healthcare records generation with prompt learning}.
\newblock In \emph{Proceedings of the 2022 Conference on Empirical Methods in
  Natural Language Processing}, pages 2873--2885, Abu Dhabi, United Arab
  Emirates. Association for Computational Linguistics.

\bibitem[{Zhan et~al.(2026)Zhan, Gueorguieva, Hernandez, Suh, Ong, and
  Li}]{zhan2026discoursediversity}
Hongli Zhan, Emma~S. Gueorguieva, Javier Hernandez, Jina Suh, Desmond~C. Ong,
  and Junyi~Jessy Li. 2026.
\newblock \href {https://arxiv.org/abs/2604.11742} {Discourse diversity in
  multi-turn empathic dialogue}.
\newblock \emph{Preprint}, arXiv:2604.11742.

\bibitem[{Zhao and Field(2025)}]{zhao2025controlled}
Zihao Zhao and Anjalie Field. 2025.
\newblock \href {https://doi.org/10.18653/v1/2025.emnlp-main.1663} {Controlled
  generation for private synthetic text}.
\newblock In \emph{Proceedings of the 2025 Conference on Empirical Methods in
  Natural Language Processing}, pages 32720--32735, Suzhou, China. Association
  for Computational Linguistics.

\bibitem[{Zhou et~al.(2025)Zhou, Giorgi, Mani, Xu, Liang, and
  Tan}]{zhou2025feedback}
Karen Zhou, John~Michael Giorgi, Pranav Mani, Peng Xu, Davis Liang, and Chenhao
  Tan. 2025.
\newblock \href {https://doi.org/10.18653/v1/2025.emnlp-industry.104} {From
  feedback to checklists: Grounded evaluation of {AI}-generated clinical
  notes}.
\newblock In \emph{Proceedings of the 2025 Conference on Empirical Methods in
  Natural Language Processing: Industry Track}, pages 1485--1499, Suzhou,
  China. Association for Computational Linguistics.

\end{thebibliography}

\clearpage
\appendix
\section{Part B: Per-label Confidence Margins}
\label{sec:partb-margins}

The Part B sentence-level labeler assigns each sentence the highest-cosine label among the 12 taxonomy prototypes. Per-label \emph{top1-minus-top2 cosine margin} is reported as a confidence proxy and used to qualify per-label results (limitation~M2). Table~\ref{tab:margins} reports overall, real-only, and synthetic-only margins, sorted by overall margin descending.

\begin{table}[t]
\centering
\small
\setlength{\tabcolsep}{2pt}
\begin{tabular}{@{}lrrrl@{}}
\toprule
\textbf{Label} & \textbf{Overall} & \textbf{Real} & \textbf{Synth} & \textbf{Verdict} \\
\midrule
\texttt{resource\_referral}    & 0.046 & 0.045 & 0.059 & well-sep. \\
\texttt{follow\_up\_plan}      & 0.034 & 0.034 & 0.032 & well-sep. \\
\texttt{caseworker\_action}    & 0.027 & 0.027 & 0.028 & OK \\
\texttt{risk\_crisis}          & 0.027 & 0.026 & 0.035 & OK \\
\texttt{barrier}               & 0.026 & 0.026 & 0.033 & OK \\
\texttt{client\_response}      & 0.024 & 0.024 & 0.025 & OK \\
\texttt{collateral\_contact}   & 0.023 & 0.023 & 0.022 & OK \\
\texttt{client\_report}        & 0.023 & 0.023 & 0.020 & borderline \\
\texttt{presenting\_need}      & 0.020 & 0.020 & 0.024 & borderline \\
\texttt{administrative\_update}& 0.020 & 0.020 & 0.021 & borderline \\
\texttt{other}                 & 0.016 & 0.016 & 0.020 & fuzzy \\
\texttt{background\_context}   & 0.016 & 0.016 & 0.012 & fuzzy \\
\bottomrule
\end{tabular}
\caption{Part B per-label cosine margins for the zero-shot labeler, reported overall and by corpus. Higher margins indicate clearer prototype separation.}
\label{tab:margins}
\end{table}

\section{Related Work---Position and Detailed Discussion}
\label{sec:appendix-related-work}

This appendix expands the Related Work in \S\ref{sec:related-work} into four thematic subsections that go beyond the citation buckets already summarized in the main body: privacy-sensitive NLP data, the privacy / utility trade-offs that bound this work's posture, LLM-based synthetic clinical / sensitive-domain text, and the explicit position-of-this-work statement.

\subsection{Privacy-sensitive NLP data}

Many high-value NLP tasks involve language data that cannot be freely shared. Clinical notes, social-service case records, legal documents, educational records, workplace communications, and user-generated text may contain sensitive personal information. Carefully documented clinical resources such as MIMIC-III have enabled major progress in clinical NLP, but they also illustrate the infrastructure required to release sensitive data responsibly, including de-identification and controlled access \citep{johnson2016mimic}. De-identification shared tasks have similarly shown that identifying protected health information in narrative records is difficult and domain-specific \citep{stubbs2015i2b2}.

Human service case notes share these privacy challenges but differ from clinical notes along four axes that this paper repeatedly relies on. \emph{Encounter granularity:} a clinical discharge summary or progress note typically documents a single medical encounter, whereas a case manager produces a sequence of notes over months or years that document a client's housing, benefits, employment, legal, and follow-up trajectory. \emph{Register:} clinical notes are written for downstream medical decision support (handoff to other clinicians, billing, coding), whereas case notes are written for internal service-coordination, supervision, and reimbursement audiences. \emph{Documentation structure:} clinical narrative is organized around problem lists, diagnoses, and orders, whereas case note prose is organized around presenting needs, services provided, barriers, referrals, and next steps. \emph{Identifiability profile:} both contain identifying information, but case note re-identification risk is dominated by the longitudinal combination of events rather than any single encounter. Together these differences explain why off-the-shelf clinical-NLP resources (which we cite as exemplars throughout) are not drop-in for case note tasks. Case notes in this genre often document housing instability, benefits access, referrals, legal involvement, family circumstances, client engagement, resources provided, and caseworker follow-up. These details can be identifying even when names and contact information are removed. Longitudinal sequences further increase risk because combinations of events may identify a person more strongly than any isolated note. NLP and HCI work on this exact document genre is comparatively scarce: \citet{gondimalla2024aligning} study how to redesign data-labeling interfaces for social-service case notes in a homelessness-services agency, and \citet{nelson2024homelessness} elicit caseworker requirements for LLM-based case note summarization in the same setting; both motivate the need for data similar to real case notes that do not exist publicly.

This paper does not release real or redacted real notes. Instead, real notes are used privately to learn distributions, transitions, and style patterns. The dataset archive contains synthetic records.

\subsection{Synthetic data for privacy and utility}

Synthetic data is often proposed as a way to preserve useful statistical structure while reducing exposure of real individuals, but it is neither automatically private nor automatically useful: prior work shows that empirical similarity checks should not be mistaken for formal privacy guarantees \citep{stadler2022syntheticnotsilver,royalsociety2022synth}, and that even differentially-private synthetic text in high-stakes domains can underperform on critical subgroups despite formal DP guarantees \citep{ramesh2024dpeval}. Privacy-aware text-generation pipelines have responded with explicit threat models: \citet{zhao2025controlled} add entity-aware control codes and report entity-leakage and privacy-preserving predictive performance jointly, while \citet{wang2022promptehr} evaluate training-membership and attribute-inference risks using generated EHRs; \citet{knowledgesg2024} architect the entire pipeline around a client--server split so only models, not raw data, cross the boundary. A complementary recent thread audits the privacy posture of \emph{the released synthetic text itself}: \citet{meeus2025canary} show that traditional out-of-distribution canaries are ineffective for synthetic-text auditing and design canaries with in-distribution prefixes and high-perplexity suffixes that leave detectable traces. For LLM training membership, \citet{chen2025revisitmia} show across thousands of LLM-MIA experiments that effectiveness varies substantially by domain and that most methods do not significantly outperform simple baselines. Our Part~G nearest-neighbor MIA sketch sits firmly on the simple-baseline side of this spectrum---it does not run a shadow-generator or canary-insertion protocol---and the reported structured-fact AUC ($0.48$, below our project-defined $0.55$ follow-up threshold) should be read as a lightweight sanity check rather than as evidence of strong privacy. The closest precedent for synthetic human services data is \citet{pickens2023synthhumanservices}, who release a fully-synthetic version of the Allegheny County Integrated Services dataset---but their artifact is structured-record only, not free-text notes, so the prose-realization and discourse-structure questions we study are out of scope for their release. We adopt the same posture: the corpus is privacy-conscious and bounded-use, not formally private; what we add is a structural boundary (unredacted source notes, source identifier records, and real fact trajectories stay local; only PII-redacted style exemplars accompany the synthetic facts sent to the external generator) plus an empirical no-copy audit (\S\ref{sec:partd}).

\subsection{LLM-based synthetic clinical / sensitive-domain text}

Pre-LLM work in this space is anchored by \citet{ive2020artificialmh}, who train a Transformer language model on de-identified UK mental-health records and evaluate text-classification transfer from synthetic to real data---a foundational precedent for both the artificial-mental-health-records framing and the train-on-synth-evaluate-on-real utility benchmark. Recent work on LLM-generated text for sensitive domains spans four complementary patterns. \emph{Synthetic-corpus construction or augmentation}: Synth-SBDH \citep{mitra2025synthsbdh} uses GPT-4 to generate a fully synthetic SBDH-labeled corpus from expert-written seed examples, then fine-tunes extractors on synthetic data before fine-tuning on real target data; \citet{kim2024generalizing} use GPT-4 to generate augmentation text for clinical de-identification; \citet{tang2023synthclinical} show that ChatGPT-generated labeled synthetic data trains a local BioBERT NER model to F1 $64.0$, versus $23.4$ for zero-shot ChatGPT, without exposing real patient text to the API, directly supporting the bounded-utility argument we make in \S\ref{sec:practitioner}. \emph{Lower-resource generators}: \citet{belkadi2025synthfreetext} sidestep the API-call privacy concern by training a $120$M-parameter masked-LM that produces synthetic medical records, reporting $96\%$ HIPAA-compliant PHI recall and $3.5\%$ re-identification risk, an architectural alternative to the structured-first pipeline we use; \citet{meoni2025synthdocs} combine knowledge injection with a reward mechanism so a small open-weight LLM generates clinical documents that preserve downstream-task utility while bounding what crosses external API boundaries. \emph{Synthetic supervision rather than synthetic text}: SYNFAC-EDIT \citep{mishra2024synfacedit} generates synthetic edit feedback for clinical summarization rather than a synthetic source-note corpus. \emph{Direction-of-flow ablations}: \citet{li2023twodirections} explicitly compare data-to-label and label-to-data pipelines for clinical extraction. A separate strand cautions about generic LLM-generated text: \citet{li2023synthetic} show that synthetic-trained classification performance degrades on more subjective tasks, and \citet{demystify2025} show that textbook-style synthetic pretraining exhibits model-collapse-like patterns (rephrased synthetic data shows no such degradation), and that beneficial mixture ratios are model-size and data-budget dependent. We position ourselves on the label-to-data end of the spectrum---structured synthetic facts first, then prose---and use the generation LLM only for final prose realization, so each note can be compared against its intended synthetic facts (\S\ref{sec:parte}). The \citet{long2024llmsynthsurvey} and \citet{loni2025synthreview} surveys provide broader background on LLM-generated synthetic text, with the latter explicitly noting the lack of standardized re-identification-risk measures across the generative-medical-text literature.

\subsection{Position of this work (extended)}

The precedents above cover three complementary axes that bracket the design space of this work. (i)~\emph{Privacy posture.} \citet{ramesh2024dpeval}, \citet{wang2022promptehr}, and \citet{knowledgesg2024} take privacy-aware routes (DP, attack-based audits, client--server splits); we take the structural route (an explicit local/API boundary + empirical no-copy audit, \S\ref{sec:partd}) without claiming formal privacy. (ii)~\emph{Generation grounding.} \citet{mitra2025synthsbdh} and \citet{kim2024generalizing} use rich-prompt LLM generation; \citet{li2023twodirections} and \citet{wang2022promptehr} use structured-input conditioning; we sit on the label-to-data end with a four-stage decoupled pipeline (\S\ref{sec:baseline-summary} validates this empirically). (iii)~\emph{Fidelity surface.} Most prior synthetic-clinical-text papers evaluate on note-level features or downstream-task transfer; \citet{petherapy2025} and \citet{cbtarcs2025} introduce protocol- and arc-level evaluations; our Part B and Part E adapt this to case note discourse moves and fact grounding, while Part A introduces a per-client trajectory-level evaluation that we are not aware of being applied to case notes before. The synthetic corpus is the artifact; the six-Part evaluation framework plus the privacy-side Part G membership-inference sketch on structured facts is the methodological contribution; their explicit pairing is the position we stake out.

\section{Dataset and Source Constraints (Extended)}
\label{sec:appendix-dataset}

\paragraph{Source setting.} The private source corpus consists of longitudinal case notes written by case managers in a community-based human service setting. Notes are authored after or during client encounters and are intended for internal program record-keeping, service delivery, reimbursement, case continuity, supervision, and review. Each note is associated with a client identifier and timestamp information, allowing notes to be grouped into client-level trajectories.

\paragraph{Example synthetic note.} Figure~\ref{fig:examplenote} shows one note from the v1.0 corpus (synthetic client \texttt{7}, second encounter in that client's trajectory) paired with a trimmed view of the structured-fact JSON that conditioned its generation. The example illustrates the schema-to-prose relationship that Parts~E and~F evaluate at corpus scale: concrete-bucket fields such as \texttt{services\_provided} and \texttt{contacted\_parties} are surfaced approximately verbatim, while other conditioning fields such as \texttt{client\_affect} can shape phrasing latently. Bracketed tokens (\texttt{[NAME]}, \texttt{[ORG]}, \texttt{[PHONE]}) are synthetic placeholders that stand in for identifying surface forms; no real client information is referenced. Per the release boundary (\S\ref{sec:source-constraints}), this is the only kind of note example the paper can carry---a release-side synthetic note---and we do not display redacted real notes in any form.

\begin{figure}[t]
\small
\textit{Synthetic case note} (\texttt{7\_note\_1\_SYNTH0001.txt}, 84 words):
\begin{verbatim}
Case Manager met with client [NAME] in
office on 05/23/2025 to follow up after
last visit's application receipt. Client
arrived irritable and focused on worsening
anxiety and trouble sleeping, still
unemployed. Case Manager talked about
options and scheduled a mental health
intake with [ORG] for Tuesday 05/28 at
10am. Case Manager contacted emergency
services at [PHONE] to confirm local
crisis line steps and hours, then printed
guidance from an email and gave it to
[NAME]. NEXT VISIT SET FOR FRIDAY 06/06
AT 1PM.
~[NAME]
\end{verbatim}

\textit{Trimmed structured-fact JSON} (populated fields shown; null fields and \texttt{text\_style} block elided):
\begin{verbatim}
{
  "contact_details": {
    "visit_type": "intake",
    "visit_location": "office"},
  "client_states": {
    "employment_status": "unemployed"},
  "presenting_issues": {
    "issues_reported": ["mental_health"],
    "issue_severity": "moderate"},
  "services_and_actions": {
    "services_provided":
      ["appointment_scheduling",
       "follow_up_check_in"],
    "service_step": "in_progress",
    "next_steps": {
      "next_steps_present": "yes",
      "next_step_type":
        "scheduled_appointment"}},
  "collateral_contacts": {
    "contacted_parties":
      ["emergency_services"],
    "contact_purpose":
      "information_gathering"},
  "narrative_content": {
    "client_affect": "angry_irritable"},
  "encounter_date": "05/23/2025"
}
\end{verbatim}
\caption{One synthetic case note (top) paired with the structured-fact JSON that conditioned its generation (bottom, trimmed). This is the schema-to-prose relationship Part~E audits at corpus scale (\S\ref{sec:parte}). Drawn from \texttt{data/synthetic/\allowbreak\{notes,facts\}/\allowbreak 7/\allowbreak 7\_note\_1\_\allowbreak SYNTH0001.\allowbreak\{txt,json\}}.}
\label{fig:examplenote}
\end{figure}

\paragraph{Corpus scale and count chain.} Table~\ref{tab:countchain} reports the count chain from source to release; only the synthetic records are included in the dataset archive.

\begin{table}[t]
\centering
\small
\setlength{\tabcolsep}{3pt}
\begin{tabular}{p{2.6cm}r@{\hspace{0.5em}}p{2.6cm}}
\toprule
\textbf{Quantity} & \textbf{Value} & \textbf{Interpretation} \\
\midrule
Post-redaction source notes & 63{,}142 & Private source rows with non-empty text and ID \\
Extracted fact JSONs & 63{,}031 & Notes as structured fact files \\
Usable facts / notes & 62{,}901 & Eval denominator after sentinel filters \\
Real clients & 1{,}682 & Private source client count \\
Synthetic clients & 87 & Fixed v1.0 corpus \\
Synthetic notes & 3{,}137 & Fixed v1.0 corpus \\
Multi-note clients & 54 & Primary trajectory track \\
Single-note clients & 33 & Reduced-feature track \\
\bottomrule
\end{tabular}
\caption{Pipeline count chain. Private source figures are reported for context only; only the 87 / 3{,}137 synthetic records are packaged.}
\label{tab:countchain}
\end{table}

\paragraph{Release boundary.} Table~\ref{tab:release} scopes the data artifacts, their relationship to real text, and their inclusion in the package. The companion software includes generation and evaluation code, documentation, intended-use guidance, pinned dependencies, and citation metadata. The bundle excludes raw notes, post-redaction real notes, real extracted facts, real client identifiers, real-redacted few-shot examples, internal logs, and the full internal git history.

\begin{table*}[t]
\centering
\small
\setlength{\tabcolsep}{4pt}
\begin{tabular}{p{2.5cm}p{3.2cm}p{2.9cm}p{2.1cm}p{3.7cm}}
\toprule
\textbf{Artifact} & \textbf{Real text?} & \textbf{Real-derived?} & \textbf{Included?} & \textbf{Role} \\
\midrule
Raw source notes & Yes & Yes & No & Private source data \\
Post-redaction source notes & Yes, redacted & Yes & No & Private extraction / evaluation input \\
Extracted real facts & Structured real-derived facts & Yes & No & Private distribution-learning input \\
Learned distributions & Aggregate patterns & Yes & No & Sampling profile (local only) \\
Synthetic facts & No real client facts & Sampled from aggregates & Yes & Structured synthetic labels / reference \\
Synthetic notes & No real-individual records & Generated from synthetic facts & Yes & Main corpus \\
Internal git history & May contain private artifacts & Yes & No & Excluded as privacy safeguard \\
\bottomrule
\end{tabular}
\caption{Data-artifact release boundary; only synthetic facts and synthetic notes are included.}
\label{tab:release}
\end{table*}

\paragraph{Intended and out-of-scope uses.} The corpus is intended for research, education, prototyping, and benchmarking in privacy-sensitive NLP. Appropriate uses include case note summarization, handoff-summary generation, primary-need classification, referral, and action extraction, barrier identification, trajectory summarization, retrieval over synthetic case histories, documentation-quality exercises, and synthetic-data evaluation research. The corpus is \emph{not} intended for operational decision-making. It should not be used to make decisions about real clients, infer actual service outcomes, evaluate real caseworker performance, estimate real population statistics, or validate deployment-ready systems without real-world testing.

\section{Schema Documentation and Feature-Vector Construction}
\label{sec:appendix-schema}

This appendix consolidates the schema reference and Part~A feature-vector construction that most benefit from explicit documentation for readers. Three artifacts: (i)~the field-by-field schema reference (Table~\ref{tab:schemafields}, every field captured in the extracted-fact JSON, with type and which evaluation Parts use it); (ii)~the Part~A client-level feature vector (Table~\ref{tab:partavector}, the 33 / 28 features that go into the Track-1 / Track-2 vectors of \S\ref{sec:partacoverage}, together with the unit-of-analysis statement); (iii)~the real-vs-synth field-populated-rate table (Table~\ref{tab:fieldfreq}, where the structured-fact distribution diverges between corpora, separate from the prose-side gap captured by Parts~B/C). The metric-to-schema crosswalk that ties evaluation Parts back to the schema is in Table~\ref{tab:metric-schema-map}.

\paragraph{Schema --- field reference.} The schema has $40$ fields organized into $12$ groups; $7$ are list-valued and the remaining $33$ are scalar (\texttt{enum} or \texttt{yes/no}). All fields are \texttt{null}-when-absent in the extracted-fact JSON; the single source of truth for the enum vocabulary is \texttt{config/valid\_schema\_values.json}. Per-field one-line descriptions are documented in the practitioner reference at \texttt{docs/interview\_packet/schema\_summary.md} (an internal reference); Table~\ref{tab:schemafields} surfaces type, list-vs-scalar, and the evaluation Parts that reference each field group.

\begin{table*}[t]
\centering
\small
\setlength{\tabcolsep}{3pt}
\begin{tabular}{p{1.8cm}p{6.0cm}p{1.0cm}p{2.0cm}}
\toprule
\textbf{Group} & \textbf{Field(s)} & \textbf{Kind} & \textbf{Used in} \\
\midrule
Contact         & \texttt{visit\_type}, \texttt{contact\_method}, \texttt{visit\_location} & enum & A, E \\
Case phase      & \texttt{case\_phase} & enum & A \\
Client state    & \texttt{housing\_status}, \texttt{employment\_status}, \texttt{benefit\_status}, \texttt{substance\_use\_status}, \texttt{mental\_health\_status}, \texttt{physical\_health\_status}, \texttt{engagement\_level} & enum & A, E \\
Issue           & \texttt{issues\_reported} & list & A, E, F \\
                & \texttt{issue\_severity}, \texttt{issue\_nature} & enum & F \\
Service         & \texttt{services\_provided} & list & A, E \\
                & \texttt{service\_step}, \texttt{service\_acceptance} & enum & --- \\
                & \texttt{barriers\_encountered.barrier\_present} & y/n & A \\
                & \texttt{barriers\_encountered.barrier\_types} & list & A \\
                & \texttt{next\_steps.next\_steps\_present} & y/n & A \\
                & \texttt{next\_steps.next\_step\_type} & enum & A, E \\
Crisis          & \texttt{crisis\_type}, \texttt{crisis\_response} & enum & A, E \\
Resource        & \texttt{resource\_types} & list & A, E \\
                & \texttt{resource\_amount} & enum & --- \\
Collateral      & \texttt{contacted\_parties} & list & A, E \\
                & \texttt{contact\_purpose}, \texttt{contact\_outcome} & enum & --- \\
Doc.            & \texttt{document\_types} & list & A, E \\
Housing app.    & \texttt{application\_submitted} & y/n & --- \\
                & \texttt{application\_outcome}, \texttt{move\_in\_status} & enum & --- \\
Legal           & \texttt{warrant\_status}, \texttt{csr\_hours} & enum & --- \\
                & \texttt{court\_date\_scheduled} & y/n & --- \\
Narrative       & \texttt{source\_of\_information}, \texttt{interaction\_completeness}, \texttt{client\_cooperation}, \texttt{client\_affect} & enum & --- \\
                & \texttt{temporal\_types} & list & --- \\
\bottomrule
\end{tabular}
\caption{Schema fields and evaluation use. Kind: \texttt{enum}, \texttt{list}, or yes/no; A, E, and F indicate use of at least one listed field, including subgroup stratification; ``---'' means no use in these three Parts. Within Client state, E uses housing and employment status; within Crisis, E uses crisis type only.}
\label{tab:schemafields}
\end{table*}

\paragraph{Part A client-level feature vector --- construction.} Each row of the Part~A feature matrix is \emph{one client's full trajectory}. The $n_\text{notes}$ notes for a given client are aggregated into a single $33$-d feature vector for Track~1 (multi-note, $n_\text{notes} \geq 2$) and a $28$-d reduced vector for Track~2 (single-note, $n_\text{notes}=1$, no transition-derived features). Table~\ref{tab:partavector} groups the features into seven blocks. Two implementation notes are worth pinning here: (i)~\textbf{volatility is per-field, not aggregate} --- there are four named volatility scalars (\texttt{engagement\_volatility}, \texttt{housing\_volatility}, \texttt{employment\_volatility}, \texttt{case\_phase\_volatility}), each defined as the \emph{transition-change rate} (fraction of consecutive observed-pair value-changes) rather than a standard deviation, because the underlying values are categorical; mental-health volatility is intentionally omitted because the trajectory averages already encode the same signal as group shares (\texttt{mental\_health\_traj\_*}). (ii)~\textbf{persistence is one aggregate scalar} --- \texttt{unresolved\_need\_persistence} is the mean per-issue carry-over rate across consecutive notes, averaged across all issues that ever appear in the trajectory; per-issue persistence is computed and used in the sampler (\S\ref{sec:pipeline}) but not exposed in the Part~A vector. The implementation lives in \texttt{pipeline/\allowbreak evaluation/\allowbreak part\_a\_coverage/\allowbreak build\_client\_vectors.py}.

\begin{table*}[t]
\centering
\small
\setlength{\tabcolsep}{3pt}
\begin{tabular}{p{1.7cm}p{4.3cm}p{1.0cm}p{2.1cm}p{5.0cm}}
\toprule
\textbf{Group} & \textbf{Features} & \textbf{Track~1} & \textbf{Track~2} & \textbf{Captures} \\
\midrule
Volume / activity & \texttt{log\_n\_notes}, \texttt{log\_case\_duration\_days}, \texttt{notes\_per\_active\_month} & 3 & 3 (case duration $=0$) & Trajectory length, span, contact rate \\
Issue / service mix & \texttt{n\_unique\_issues}, \texttt{issue\_entropy}, \texttt{n\_unique\_services}, \texttt{service\_entropy} & 4 & 4 & Diversity / breadth of presenting issues and CM actions \\
Per-note rates & \texttt{crisis\_note\_rate}, \texttt{barrier\_note\_rate}, \texttt{intake\_share}, \texttt{resource\_provision\_rate}, \texttt{collateral\_rate}, \texttt{documentation\_rate}, \texttt{next\_steps\_rate} & 7 & 7 & Within-trajectory share of crisis / barrier / resource / collateral notes \\
Sparsity-as-feature & \texttt{engagement\_observed\_rate}, \texttt{housing\_observed\_rate}, \texttt{employment\_observed\_rate}, \texttt{mental\_health\_observed\_rate}, \texttt{case\_phase\_observed\_rate} & 5 & 5 & Whether status fields are populated, separately from their values \\
Volatility (per field) & \texttt{engagement\_volatility}, \texttt{housing\_volatility}, \texttt{employment\_volatility}, \texttt{case\_phase\_volatility} & 4 & --- & Transition-change rate per status field; undefined for $n_\text{notes}=1$ \\
Persistence & \texttt{unresolved\_need\_persistence} & 1 & --- & Mean per-issue carry-over across notes; undefined for $n_\text{notes}=1$ \\
Trajectory mix & \texttt{housing\_traj\_*} (4), \texttt{employment\_traj\_*} (2), \texttt{mental\_health\_traj\_*} (3) & 9 & 9 & Group-mean one-hot on \texttt{state\_groups} for housing, employment, mental health \\
\midrule
\textbf{Total} & --- & \textbf{33$^{\dagger}$} & \textbf{28} & \\
\bottomrule
\end{tabular}
\caption{Part~A client-level feature-vector construction. Track~1 (multi-note, $n_\text{notes} \geq 2$): full $33$-d vector. Track~2 (single-note): reduced $28$-d vector dropping volatility and persistence features. $^{\dagger}$NaN values (when $<2$ notes have the relevant field observed) are imputed to $0$; missingness is carried by \texttt{*\_observed\_rate}.}
\label{tab:partavector}
\end{table*}

\paragraph{Real-vs-synth field-populated rates.} Table~\ref{tab:fieldfreq} reports the share of notes in each corpus that populate each schema field, separating real ($n_\text{real} = 63{,}031$ extracted-fact JSONs) from synthetic ($n_\text{synth} = 3{,}137$ v1.0 notes). Most fields are within $\pm 3$~pp; the systematic gaps are concentrated in the longitudinal / collateral / temporal fields. Synthetic notes under-populate \texttt{temporal\_types} ($-59.3$~pp), the three \texttt{collateral\_*} fields (\texttt{contacted\_parties} $-35.1$~pp, \texttt{contact\_purpose} $-37.0$~pp, \texttt{contact\_outcome} $-30.1$~pp), the two \texttt{next\_steps\_*} fields ($-25.8$ / $-25.6$~pp), \texttt{resource\_amount} ($-17.4$~pp), and the barrier flags ($-8.7$~pp). Synthetic notes \emph{over}-populate only \texttt{visit\_type} ($+22.2$~pp). The collateral / next-step deficits occur in sampled facts and align with the corresponding Part~B move-frequency gaps (Figure~\ref{fig:partbmove}) and practitioner emphasis on explicit coordination and next steps (\S\ref{sec:practitioner}, P3/P4/P8). The full $40$-row table applies the population rules in \texttt{pipeline/\allowbreak stages/\allowbreak analysis/\allowbreak field\_frequency\_table.py} to the v1.0 corpus and is stored in \texttt{data/\allowbreak derived/\allowbreak analysis/\allowbreak field\_frequency\_table\_\allowbreak release.\{json,csv\}}; here we show the twelve largest absolute gaps and five near-aligned fields.

\begin{table}[t]
\centering
\small
\setlength{\tabcolsep}{3pt}
\begin{tabular}{lrrr}
\toprule
\textbf{Field} & \textbf{Real} & \textbf{Synth} & \textbf{$\Delta$} \\
\midrule
\texttt{temporal\_types} & 61.0\% & 1.7\% & $-59.3$ \\
\texttt{contact\_purpose} & 40.0\% & 3.0\% & $-37.0$ \\
\texttt{contacted\_parties} & 42.0\% & 6.9\% & $-35.1$ \\
\texttt{contact\_outcome} & 32.3\% & 2.2\% & $-30.1$ \\
\texttt{next\_steps\_present} & 42.2\% & 16.4\% & $-25.8$ \\
\texttt{next\_step\_type} & 42.0\% & 16.4\% & $-25.6$ \\
\texttt{visit\_type} & 75.5\% & 97.7\% & $\textbf{+22.2}$ \\
\texttt{resource\_amount} & 23.5\% & 6.1\% & $-17.4$ \\
\texttt{barrier\_present} & 17.9\% & 9.1\% & $-8.7$ \\
\texttt{barrier\_types} & 15.5\% & 9.1\% & $-6.4$ \\
\texttt{application\_submitted} & 7.4\% & 1.2\% & $-6.2$ \\
\texttt{application\_outcome} & 5.8\% & 0.2\% & $-5.6$ \\
\midrule
\multicolumn{4}{l}{\textit{Reference: near-aligned fields ($|\Delta| \leq 3$~pp)}} \\
\texttt{services\_provided} & 80.0\% & 82.8\% & $+2.8$ \\
\texttt{issues\_reported} & 57.9\% & 60.5\% & $+2.6$ \\
\texttt{contact\_method} & 77.0\% & 77.0\% & $0.0$ \\
\texttt{client\_affect} & 34.2\% & 33.9\% & $-0.2$ \\
\texttt{housing\_status} & 28.2\% & 27.2\% & $-1.0$ \\
\bottomrule
\end{tabular}
\caption{Real-vs-synth populated-rate gaps (largest $|\Delta|$ plus five near-aligned fields; $n_\text{real}=63{,}031$, $n_\text{synth}=3{,}137$). Populated means non-null and nonempty after removing schema null-equivalents; most other fields are within approximately $\pm3$~pp.}
\label{tab:fieldfreq}
\end{table}

\paragraph{Consolidated evaluation crosswalk.} A single Part-by-Part crosswalk pairing each evaluation Part with what it measures, the structured-fact fields it depends on, the bounded claim it supports, and the claim it does \emph{not} test lives with the per-Part protocol detail in Appendix~\ref{sec:appendix-evalframework} (Table~\ref{tab:metric-schema-map}).

\section{Direct-Prompt Baseline}
\label{sec:appendix-baseline}

To test whether the four-stage decoupling actually buys anything over a single direct-prompt call, we run a controlled ablation. The baseline asks the same \texttt{gpt-5.2} model to generate notes from a one-line scenario seed only --- no extracted facts, no learned distributions, no sampler, no per-client trajectory context, and no few-shot rotation. The $162$-note baseline corpus is scored on the same Part B discourse-marginal Jensen-Shannon divergence and Part C BGE-Large MMD$^2$ used for the structured pipeline (Table~\ref{tab:baseline} in \S\ref{sec:pipeline}).

The direct-prompt corpus diverges $4.1\times$ more on discourse-move JSD and $7.1\times$ more on BGE-Large MMD$^2$ than the structured pipeline. This supports the structured-first decoupling: it materially improves both discourse-structure and embedding-space fidelity over a competent direct-prompt baseline using the same generator. A size-matched sensitivity check draws $40$ random $162$-note synthetic subsamples (matching the baseline note count) and recomputes both metrics against the same real corpus: mean Part~B JSD $0.020$ [$0.015, 0.027$] vs.\ full-corpus $0.018$, and mean Part~C MMD$^2$ $0.024$ [$0.022, 0.028$] vs.\ full-corpus $0.024$ --- both remain far below the direct-prompt values ($0.074$ / $0.172$). Client-capped synth pools (max $5$/$10$/$20$ notes per client) likewise leave JSD and MMD$^2$ within a few thousandths of the full-corpus figures, so neither sample size nor long-trajectory dominance drives the gap. The same conclusion is qualitatively triangulated by the practitioner pilot (\S\ref{sec:practitioner}): $7$ of $8$ practitioners preferred the structured-fact-driven Set~B trajectory over the briefer direct-prompt-style Set~A. Artifact: \texttt{outputs/\allowbreak evaluation/\allowbreak part\_b\_c\_robustness/\allowbreak robustness\_report.json}.

\section{Evaluation Framework Details}
\label{sec:appendix-evalframework}

\paragraph{Utility-centered evaluation.} A synthetic note can be fluent but weak if it lacks concrete referrals, barriers, follow-up, or trajectory continuity. Conversely, a synthetic note can be distinguishable from real data while still being useful for training, prototyping, or benchmarking. The evaluation therefore asks: \emph{Do the synthetic notes preserve the functional properties needed for bounded research, training, and evaluation uses?} Table~\ref{tab:metric-schema-map} is the consolidated Part-by-Part crosswalk that pairs each Part with what it measures, the schema fields it depends on, its bounded scope, and what it does \emph{not} test; the rest of this appendix expands the per-Part protocol detail.

\begin{table*}[!htbp]
\centering
\small
\setlength{\tabcolsep}{2pt}
\begin{tabular}{p{1.9cm}p{2.9cm}p{3.5cm}p{3.3cm}p{3.4cm}}
\toprule
\textbf{Part} & \textbf{What it measures} & \textbf{Schema dependency} & \textbf{Scope (supports)} & \textbf{What it does \emph{not} test} \\
\midrule
A (Coverage) & Whether synthetic clients sit in plausible regions of real trajectory space (nearest neighbors, balanced 1-NN, matched AUC, cluster coverage) & Client-level $33$/$28$-d vector aggregated from selected schema fields and trajectory timing (Table~\ref{tab:partavector}); volatility uses 4 status fields, persistence uses \texttt{issues\_reported} & Identification of trajectory-level gaps & Population-level representativeness; deployment readiness \\
B (Discourse) & Documentation organization (move marginals, bigram transitions, stickiness, template diversity) & Sentence-level $12$-move taxonomy labels (no schema dependency at scoring time; labels condition on prose only) & Broad documentation-structure comparison & Gold-standard human discourse annotation (M2) \\
C (Embedding) & Semantic overlap and separability under selected encoders (MMD, Fr\'echet, classifier AUC, precision / recall / coverage) & Note-level prose embeddings (BGE-Large, MPNet; no schema dependency) & Semantic relatedness under selected encoders & Full distributional equivalence; trajectory-level fidelity \\
D (Overlap) & Above-threshold overlap with real text (few-shot-leak pairs, unique-$n$-gram intersection) & Few-shot \texttt{*\_examples.json} provenance + token-level overlap on prose (no schema dependency) & No above-threshold copying detected under this audit & Formal anonymization guarantee \\
E (Alignment) & Lexical fact recall plus a synthetic-only LLM-judge audit of semantic realization and contradiction (\S\ref{sec:parte}) & $12$ schema fields split concrete~/~soft (\texttt{issues\_reported}, \texttt{services\_provided}, \texttt{housing\_status}, \texttt{employment\_status}, \texttt{document\_types}, \texttt{resource\_types}, \texttt{contacted\_parties} concrete; \texttt{visit\_type}, \texttt{contact\_method}, \texttt{visit\_location}, \texttt{next\_step\_type}, \texttt{crisis\_type} soft) & Structured-content grounding in prose & Stylistic faithfulness; unsupported claims outside the schema (M8); human validation (M9) \\
F (Transfer) & Whether train-on-synth transfers to a held-out real downstream task (macro-F1 transfer ratio vs.\ direct-prompt baseline) & \texttt{presenting\_issues.\allowbreak issues\_reported[0]} in extracted real or sampled synthetic facts; direct-prompt scenario seeds & Bounded task-level utility on a held-out real classifier & Human-coded clinical primary-need accuracy (M7); transfer to arbitrary tasks \\
G (Privacy) & Detectable few-shot-membership signal in fact and prose representations (ROC-AUC vs.\ project-defined $0.55$ follow-up threshold, per channel) & Sparse $358$-d multi-hot over 33 singleton + 7 list schema fields (fact channel); MPNet $768$-d note embeddings (prose channel) & Absence of detectable membership signal under this screen & Formal differential privacy; full attribute-inference attack \\
Practitioner & Domain judgments about utility, realism, trajectory coherence (eight caseworker interviews; \S\ref{sec:practitioner}) & --- (qualitative coding of practitioner responses; no scoring-time schema dependency) & Domain-informed use cases and limitations & Population-level estimates of all caseworker views \\
\bottomrule
\end{tabular}
\caption{Evaluation crosswalk for Parts A--G and the practitioner pilot: measurement, schema dependency, supported claim, and explicit non-claim; cross-Part interpretation appears in \S\ref{sec:discussion}.}
\label{tab:metric-schema-map}
\end{table*}

\paragraph{Part A: client scenario coverage.} Track 1 contains multi-note clients with a 33-feature trajectory vector (volatility and persistence features included); Track 2 contains single-note clients with a reduced 28-feature vector because trajectory volatility is undefined for one-note cases (Table~\ref{tab:tracks}). We report nearest-neighbor distances, real coverage, synthetic precision, balanced 1-NN accuracy as a classifier two-sample statistic \citep{lopezpaz2017classifier}, and size-matched logistic AUC where sample size permits. Balanced accuracy is used instead of raw accuracy because the real and synthetic sample sizes are highly imbalanced --- at the Track 1 imbalance of $1{,}331{:}54$ a constant ``real'' classifier achieves raw accuracy $0.961$, so raw accuracy is uninformative.

\begin{table}[!htbp]
\centering
\small
\begin{tabular}{lrrl}
\toprule
\textbf{Track} & \textbf{Real $n$} & \textbf{Synth $n$} & \textbf{Features} \\
\midrule
Track 1 (multi) & 1{,}331 & 54 & 33 \\
Track 2 (single) & 350 & 33 & 28 \\
\bottomrule
\end{tabular}
\caption{Part A tracks. Track~2 ($n_\text{synth}=33$) reports size-matched AUC; Track~1 covers multi-note clients.}
\label{tab:tracks}
\end{table}

\paragraph{Part B: documentation-structure similarity.} Part B labels each sentence with one of 12 documentation moves: \texttt{client\_report}, \texttt{presenting\_need}, \texttt{background\_context}, \texttt{caseworker\_action}, \texttt{resource\_referral}, \texttt{barrier}, \texttt{client\_response}, \texttt{risk\_crisis}, \texttt{follow\_up\_plan}, \texttt{administrative\_update}, \texttt{collateral\_contact}, and \texttt{other}. Labels are assigned using a local embedding-based zero-shot classifier. For each label, we build a prototype from the label name, definition, and examples; encode prototypes using \texttt{BAAI/bge-large-en-v1.5}; encode sentences with the same model; and assign the highest-cosine label. This avoids sending real client text to an external API. Per-label confidence (top1-minus-top2 cosine margin) ranges from $0.046$ for \texttt{resource\_referral} to $0.016$ for \texttt{other} and \texttt{background\_context}; the full table is in Appendix~\ref{sec:partb-margins}. We report move-marginal Jensen-Shannon divergence, note-level permutation tests with Holm correction, row-normalized bigram transition differences, and template diversity. A template is the sequence of move labels within a note. We use top-$k$ template shares and Heaps' law to test whether synthetic notes collapse into a narrow set of repeated documentation patterns (cf.\ \citealp{namuduri2025qudsim}). Heaps' $\beta$ summarizes distinct-template growth across corpus sizes, complementing top-$k$ shares. This labeler is not a human gold standard; low-margin labels are interpreted cautiously throughout the results section (limitation~M2).

\paragraph{Part C: embedding-space semantic fidelity.} Part C embeds notes using two local sentence encoders: \texttt{BAAI/bge-large-en-v1.5} and \texttt{sentence-transformers/all-mpnet-base-v2}. Using two encoders reduces dependence on a single embedding geometry. We report (i)~Maximum Mean Discrepancy with an RBF kernel \citep{gretton2012mmd}; (ii)~Fr{\'e}chet distance in a shared PCA-to-256-dimensional subspace \citep{heusel2017gans} --- the projection reduces covariance dimension before the matrix square-root calculation; (iii)~size-matched classifier AUC, with real subsets sampled without replacement of size $n_\text{synth}$ to avoid sample-size confounds; and (iv)~Kynk{\"a}{\"a}nniemi-style precision/recall \citep{kynkaanniemi2019improved} plus a coverage variant. Embedding metrics are useful for measuring semantic overlap and distributional gaps, but they do not directly measure actionability, trajectory coherence, or practitioner usefulness.

\paragraph{Practitioner-pilot protocol.} We conducted eight caseworker interviews to elicit domain judgments about utility, realism, trajectory coherence, and possible uses. Alongside schema review, the note-review tasks comprised three components: (i)~\textbf{chronological note sets} --- short synthetic client trajectories used to assess whether participants could infer case progression; (ii)~\textbf{pairwise note comparisons} --- synthetic note pairs differing in detail, structure, or tone, used to elicit documentation-quality judgments; (iii)~\textbf{Turing-style authenticity task} --- mixed real redacted and synthetic notes, used to elicit features practitioners associate with authenticity and usefulness. We analyze interviews using a hybrid deductive/inductive approach. Deductive codes correspond to utility dimensions in the evaluation framework: actionability, continuity, training value, benchmarking value, realism, and safety. Inductive codes capture additional use cases and concerns raised by participants. Because the goal is exploratory domain feedback rather than population-level estimation, we report practitioner-supported themes rather than prevalence estimates.

\FloatBarrier
\section{Coverage and Discourse Details}
\label{sec:appendix-coverage-discourse}

This appendix collects the Part A Track-2 detail and the Part B per-label / template-diversity tables and figures that were summarized in \S\ref{sec:partacoverage} and \S\ref{sec:partbsummary}.

\begin{figure*}[t]
\centering
\includegraphics[width=\textwidth]{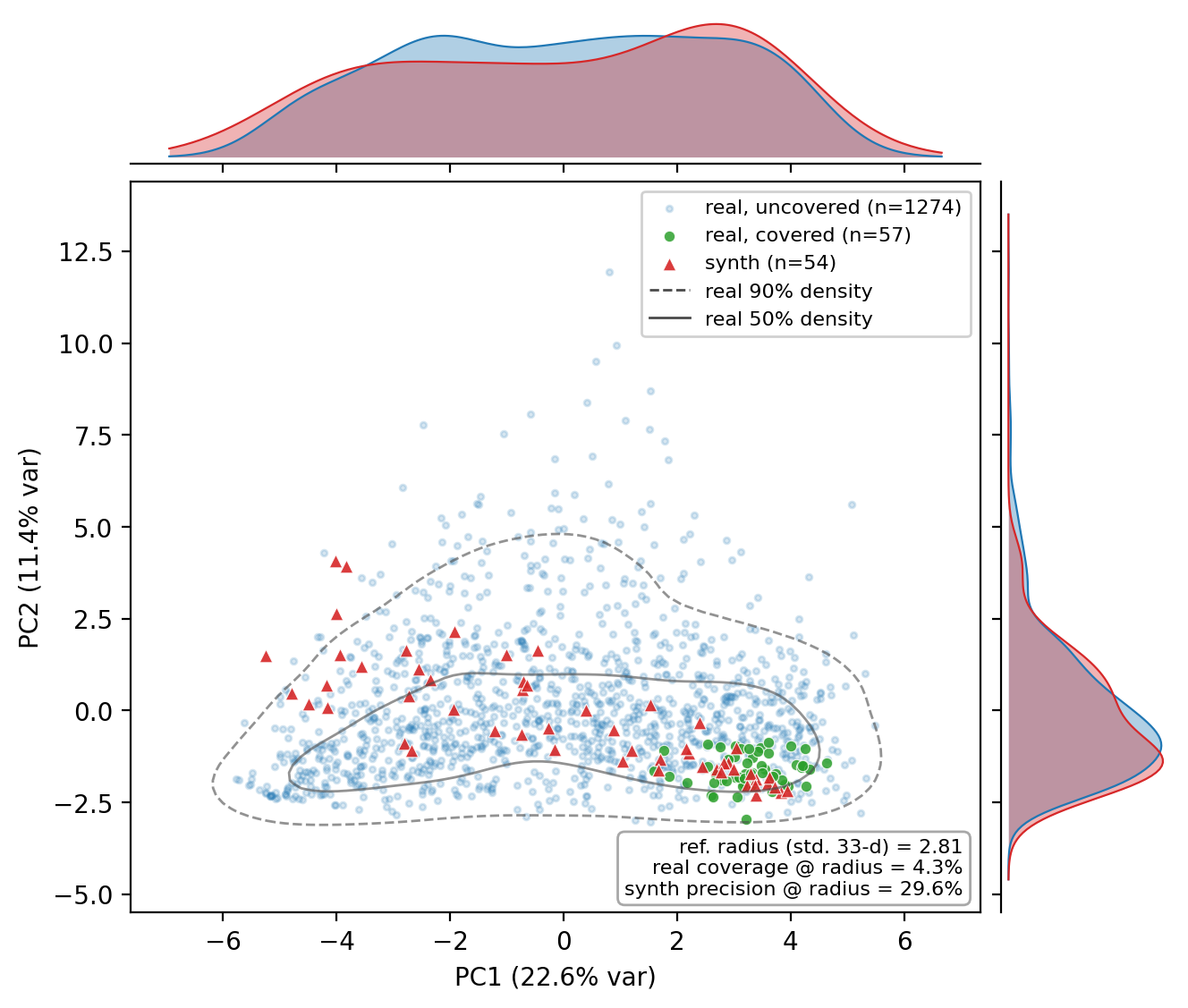}
\caption{PCA projection of multi-note clients (real $n=1{,}331$, blue; synthetic $n=54$, red) on the first two principal components of the standardized 33-feature client vector. Green circles mark the $57$ real clients within the reference radius, visualizing $4.3\%$ real-coverage; the in-axes annotation reports $29.6\%$ synth-precision. Gray contours mark the $50\%/90\%$ real-density mass.}
\label{fig:partapca}
\end{figure*}

\paragraph{Part A Track 1 PCA (multi-note clients).}
Figure~\ref{fig:partapca} projects the standardized 33-feature client vector for $1{,}331$ real and $54$ synthetic multi-note clients onto the first two principal components. Synthetic trajectories occupy plausible central regions of the real trajectory space (high precision: $29.6\%$ of synthetic clients sit within the reference radius of a real client) while leaving the tails sparser (bounded recall: $4.3\%$ of real clients are covered). Much of that bounded-recall gap is a geometric consequence of $54$ synthetic clients against $1{,}331$ real ones, but the size-matched control (Table~\ref{tab:partacontrol}) still leaves a residual diversity gap after accounting for pool size (coverage is $23\%$ of the same-size real reference; same-size real subsample precision $49.9\%$ vs.\ synth $29.6\%$), so the figure alone does not rule out under-coverage of the real trajectory tails.

\paragraph{Part A cluster ratios.} K-means on real multi-note clients ($k=4$, silhouette $0.157$) yields per-cluster synth/real share ratios $0.60, 0.76, 0.92, 1.34$ on the fixed corpus. The contingency test is non-significant ($\chi^2 = 3.90$, $p = 0.27$, Cram\'er's $V = 0.05$); all four ratios fall inside the project's $0.5\times$--$2.0\times$ flag band. Under weak clustering this is consistent with no single cluster being systematically empty, but it does not establish that no meaningful region is missing.

\paragraph{Part A radius-sensitivity sweep.} The headline numbers fix the reference radius at $\rho = 2.81$, the median real-real nearest-neighbor distance ($p_{50}$). Table~\ref{tab:partaradius} sweeps the radius across $p_{50}$, $p_{75}$, $p_{90}$, and $p_{95}$ of the real-real distance distribution and reports the corresponding real-coverage and synth-precision values, so the reader can judge how the precision/recall split moves as the radius is loosened. Both columns increase monotonically, as expected: at the looser $p_{90}$ radius the synthetic corpus covers $61\%$ of real clients at $89\%$ precision; at $p_{95}$, $82\%$ coverage at $94\%$ precision. The precision side is stable above $53\%$ from $p_{75}$ onward, while coverage rises because more real clients fall within the looser radius. The $p_{50}$ row is the primary result.

\begin{table}[t]
\centering
\small
\setlength{\tabcolsep}{3pt}
\begin{tabular}{lrrr}
\toprule
\textbf{Percentile} & \textbf{$\rho$} & \textbf{Real cov.} & \textbf{Synth prec.} \\
\midrule
$p_{50}$ (primary)   & 2.806 & 4.3\%  & 29.6\% \\
$p_{75}$             & 3.722 & 21.2\% & 53.7\% \\
$p_{90}$             & 5.108 & 61.1\% & 88.9\% \\
$p_{95}$             & 6.210 & 81.6\% & 94.4\% \\
\bottomrule
\end{tabular}
\caption{Part A Track 1 radius-sensitivity sweep: real-coverage and synth-precision at four percentiles of the real-real nearest-neighbor distance distribution. The $p_{50}$ row is the primary result; the remaining rows are reported here so the reader can read precision/recall directly off the curve.}
\label{tab:partaradius}
\end{table}

\paragraph{Part A Track 2 (single-note clients).} Track 2 contains $350$ real and $33$ synthetic clients. Balanced 1-NN accuracy is $0.584$, size-matched logistic AUC is $0.573$ $[0.438, 0.682]$, real coverage within the reference radius is $26.3\%$, and synthetic precision is $21.2\%$. Both classifier diagnostics sit close to the null, consistent with single-note clients carrying less per-unit signal than multi-note trajectories.

\paragraph{Part A size-matched real-real coverage control.} The raw coverage numbers are bounded by pool size: $54$ (Track~1) or $33$ (Track~2) points cannot cover $1{,}331$ or $350$ real clients at the tight $p_{50}$ radius even in principle. The control estimates a same-size real reference by drawing $200$ random real subsamples of exactly $n_\text{synth}$ clients, scoring the \emph{remaining} real clients against each subsample with the identical radius and NaN-handling protocol, and averaging (Table~\ref{tab:partacontrol}). Track~1 synthetic coverage ($4.3\%$) is $23\%$ of the real-control reference ($18.6\%$ [$13.6, 22.2$]); Track~2 ($26.3\%$) is $83\%$ of its reference ($31.7\%$ [$24.0, 36.6$]). The same-size real subsample also out-covers the synthetic pool on precision (Track~1: $49.9\%$ vs.\ $29.6\%$), so a genuine diversity gap remains on the multi-note track after accounting for size---consistent with the under-populated collateral/temporal fields (Table~\ref{tab:fieldfreq}) and the Part~C recall reading. Artifact: \texttt{outputs/\allowbreak evaluation/\allowbreak size\_matched\_\allowbreak controls.json}.

\begin{table}[t]
\centering
\small
\setlength{\tabcolsep}{3pt}
\begin{tabular}{lrrr}
\toprule
\textbf{Track} & \textbf{Synth} & \textbf{Real-control ref.} & \textbf{Frac.} \\
\midrule
1 (multi, $k{=}54$)  & 4.3\%  & 18.6\% [13.6, 22.2] & 0.23 \\
2 (single, $k{=}33$) & 26.3\% & 31.7\% [24.0, 36.6] & 0.83 \\
\bottomrule
\end{tabular}
\caption{Part A size-matched coverage control: v1.0 corpus coverage versus the mean from $200$ random real subsamples of the same size ($95\%$ interval in brackets). ``Frac.''\ is the ratio to that real-control reference.}
\label{tab:partacontrol}
\end{table}

\begin{figure*}[t]
\centering
\includegraphics[width=\textwidth]{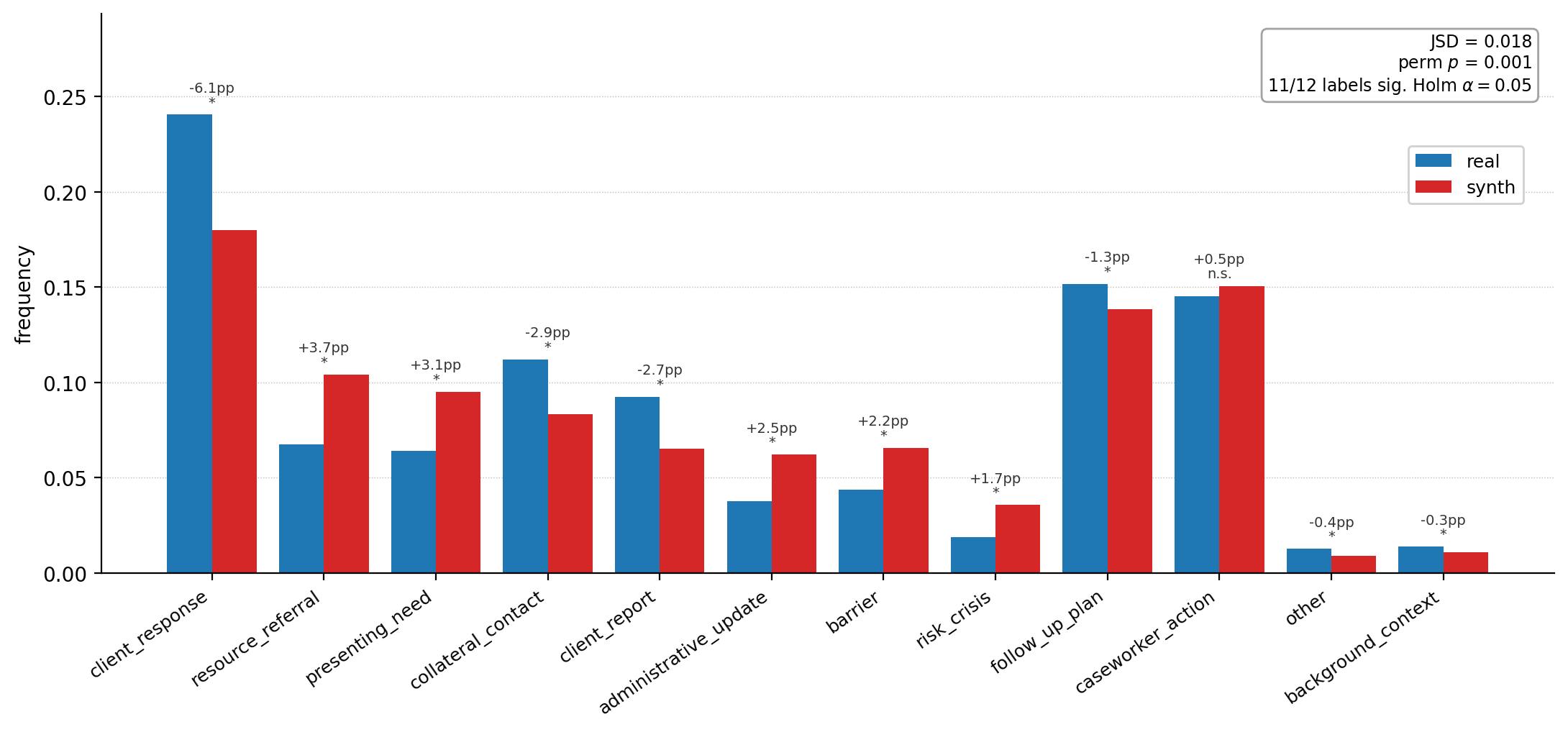}
\caption{Marginal move-label frequency in real ($n=62{,}901$ notes) vs.\ synthetic ($n=3{,}137$ notes), sorted by $|\Delta|$. JSD $=0.018$ (perm $p=0.001$); $11/12$ labels significant at Holm $\alpha{=}0.05$. Largest gaps: \texttt{client\_response} $-6.1$~pp, \texttt{resource\_referral} $+3.7$~pp, \texttt{presenting\_need} $+3.1$~pp.}
\label{fig:partbmove}
\end{figure*}

\paragraph{Part B per-label move-marginal gaps.} Figure~\ref{fig:partbmove} plots the per-label marginal distributions for all 12 moves, with signed deltas and Holm-significance markers annotated directly on each bar pair. Synthetic notes overrepresent explicit referrals, presenting needs, and administrative updates, while underrepresenting client responses, collateral contacts, and client reports.

\begin{figure*}[t]
\centering
\includegraphics[width=\textwidth]{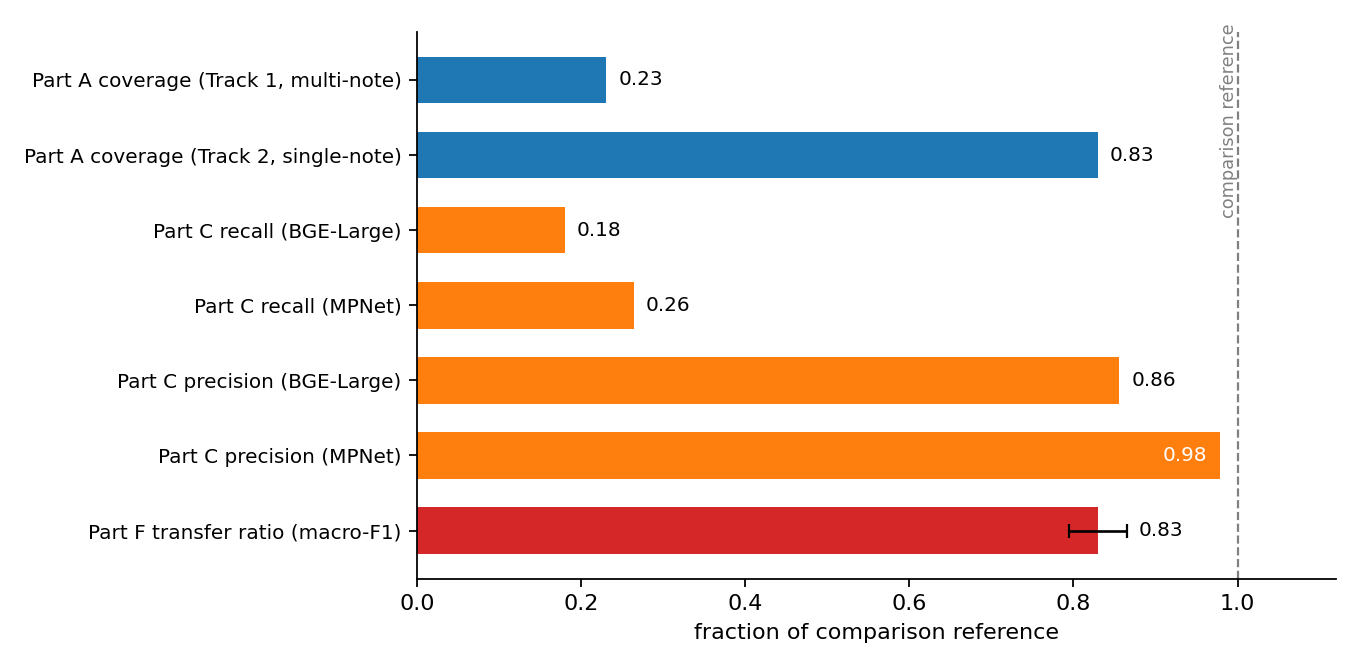}
\caption{Headline metrics normalized by their comparison references: same-size real controls for Part~A, disjoint real-vs-real controls for Part~C, and the real-on-real upper bound for Part~F.}
\label{fig:ceilingsummary}
\end{figure*}

\paragraph{Reference-normalized summary.} Figure~\ref{fig:ceilingsummary} collects the headline metrics on a single axis, each normalized by its comparison reference rather than an absolute $1.0$: Part~A coverage by the same-size real-vs-real control, Part~C precision/recall by the disjoint real-vs-real manifold control, and Part~F by the real-on-real macro-F1 upper bound. The pattern across Parts is consistent: metrics measuring \emph{local plausibility or utility} reach $83$--$98\%$ of their references (Track~2 coverage, Part~C precision under both encoders, and Part~F transfer), while metrics measuring \emph{diversity} reach $18$--$26\%$ (Track~1 coverage and Part~C recall). Artifact: \texttt{outputs/\allowbreak evaluation/\allowbreak ceiling\_summary.json}.

\paragraph{Part B bigram transitions.} Frobenius distance between row-normalized $12 \times 12$ transition matrices is $0.559$. Per-row JSDs are largest for \texttt{other} ($0.176$, $n_\text{synth\,out}=20$ --- partly small-sample), \texttt{background\_context} ($0.080$, $n_\text{synth\,out}=164$), and \texttt{client\_report} ($0.074$, $n_\text{synth\,out}=967$); the latter two align with known fuzzy label boundaries (limitation~M2).

\paragraph{Part B move stickiness.} Stickiness $P(\text{move}_t = \text{move}_{t+1} \mid \text{move}_t = L)$ is the row-diagonal of the row-normalized transition matrix --- the probability that a sentence's move label repeats in the next sentence. \citet{zhan2026discoursediversity} introduce stickiness as a discourse-level diversity diagnostic and show that empathic-dialogue LLMs reuse tactics in the next turn at nearly double the human rate ($0.50$--$0.56$ vs.\ $0.27$); on case-note documentation we observe the \emph{opposite} pattern (Table~\ref{tab:stickiness}). Corpus-mean stickiness drops from \textbf{$0.209$} (real) to \textbf{$0.136$} (synth), a $-0.073$ shift, and $9/12$ labels are individually significant under a note-level cluster permutation with Holm correction ($n_\text{perm}=1000$, same harness as the marginal permutation). Eight labels lose stickiness in synth (largest: \texttt{client\_report} $-0.144$, \texttt{client\_response} $-0.111$, \texttt{follow\_up\_plan} $-0.105$); four gain ($\texttt{resource\_referral}$ $+0.042$, $\texttt{administrative\_update}$ $+0.036$, $\texttt{presenting\_need}$ $+0.033$, $\texttt{barrier}$ $+0.009$), and all four gainers also have positive marginal-frequency gaps in Figure~\ref{fig:partbmove}, suggesting that the structured-first pipeline shifts \emph{some} sentence-level mass into explicit single-purpose moves rather than long client-narrative or client-response runs. The directional reading (synth less repetitive at the move level than real) is consistent with the documentation-style synthesizer prompt, which favors short single-purpose sentences over the long colloquial spans real caseworkers use to recount client speech and reactions.

\begin{table}[t]
\centering
\small
\setlength{\tabcolsep}{3pt}
\begin{tabular}{lrrr}
\toprule
\textbf{Label} & \textbf{Real} & \textbf{Synth} & \textbf{$\Delta$} \\
\midrule
\texttt{follow\_up\_plan}      & 0.288 & 0.183 & $-0.105^{***}$ \\
\texttt{client\_response}      & 0.275 & 0.165 & $-0.111^{***}$ \\
\texttt{client\_report}        & 0.225 & 0.081 & $-0.144^{***}$ \\
\texttt{caseworker\_action}    & 0.207 & 0.133 & $-0.074^{***}$ \\
\texttt{collateral\_contact}   & 0.185 & 0.106 & $-0.079^{***}$ \\
\texttt{resource\_referral}    & 0.149 & 0.191 & $+0.042^{***}$ \\
\texttt{risk\_crisis}          & 0.125 & 0.115 & $-0.009$ \\
\texttt{other}                 & 0.114 & 0.050 & $-0.064$ \\
\texttt{presenting\_need}      & 0.093 & 0.126 & $+0.033^{***}$ \\
\texttt{barrier}               & 0.090 & 0.099 & $+0.009$ \\
\texttt{administrative\_update}& 0.088 & 0.125 & $+0.036^{***}$ \\
\texttt{background\_context}   & 0.070 & 0.012 & $-0.058^{***}$ \\
\midrule
\textbf{Corpus mean}           & \textbf{0.209} & \textbf{0.136} & \textbf{$-0.073$} \\
\bottomrule
\end{tabular}
\caption{Part B move stickiness by label, sorted by the real rate. $\Delta=$ synth $-$ real; $^{***}$ marks Holm-adjusted $p<0.05$ under note-level cluster permutation ($9/12$ labels).}
\label{tab:stickiness}
\end{table}

\begin{figure*}[t]
\centering
\includegraphics[width=\textwidth]{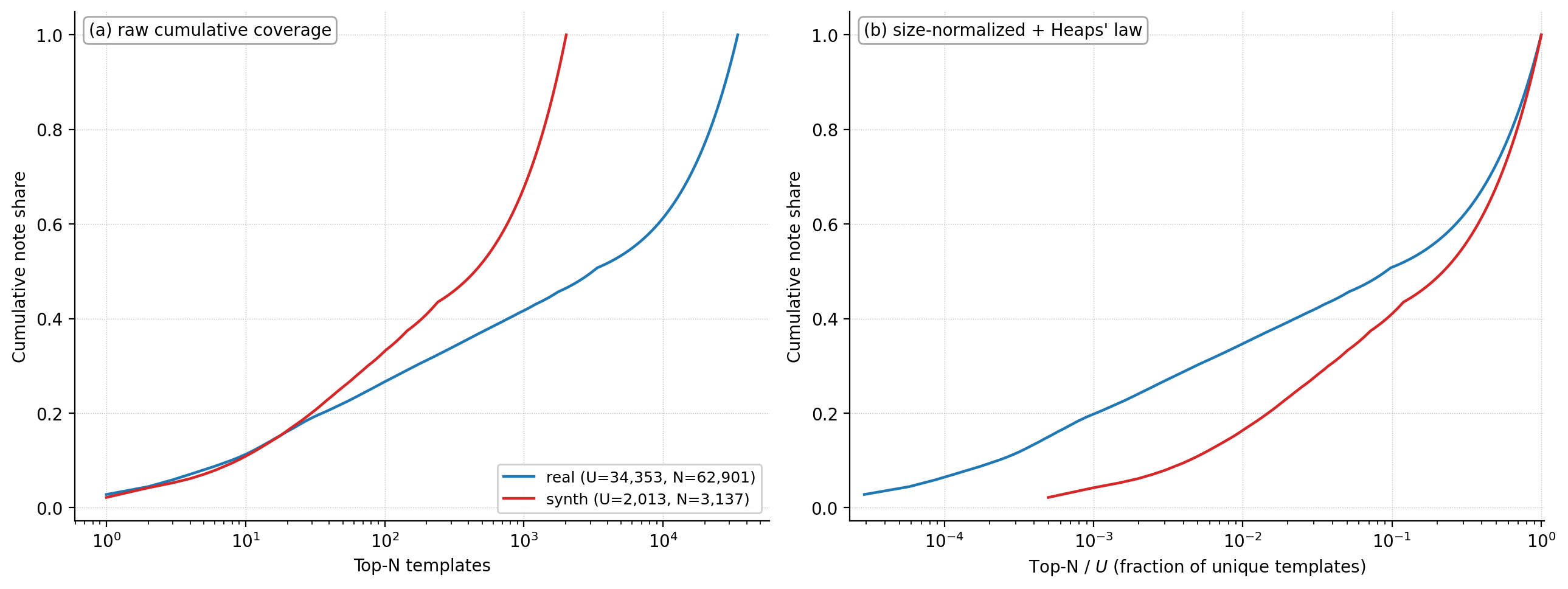}
\caption{Template-diversity coverage curves (template $=$ move-label sequence per note). \textbf{(a)}~Raw cumulative coverage (log-$N$): synthetic saturates faster due to vocabulary size ($U_\text{synth}{=}2{,}013$ vs.\ $U_\text{real}{=}34{,}353$), not a per-note diversity deficit. \textbf{(b)}~Size-normalized ($N/U$): both corpora follow a near-identical growth regime; Heaps' $\beta_\text{real}{=}0.930$ vs.\ $\beta_\text{synth}{=}0.920$.}
\label{fig:partbtemplate}
\end{figure*}

\paragraph{Part B template-diversity.} Real notes have $34{,}353$ unique move-sequence templates across $62{,}901$ notes; synthetic notes have $2{,}013$ unique templates across $3{,}137$ notes. Top-$1$ template share is $2.2\%$ (synth) vs.\ $2.8\%$ (real); top-$50$ is $25.5\%$ vs.\ $22.0\%$; Heaps' $\beta$ is $0.920$ (synth) vs.\ $0.930$ (real) (Table~\ref{tab:templates}; Figure~\ref{fig:partbtemplate}). Mild concentration appears in the synthetic top-50 tail; QUDsim-style template collapse is not observed at this corpus size.

\begin{table}[t]
\centering
\small
\begin{tabular}{lrr}
\toprule
\textbf{Metric} & \textbf{Real} & \textbf{Synthetic} \\
\midrule
Notes & 62{,}901 & 3{,}137 \\
Unique templates & 34{,}353 & 2{,}013 \\
Top-1 template share & 2.8\% & 2.2\% \\
Top-10 template share & 11.3\% & 10.9\% \\
Top-50 template share & 22.0\% & 25.5\% \\
Heaps' $\beta$ & 0.930 & 0.920 \\
\bottomrule
\end{tabular}
\caption{Template-diversity metrics. Template growth regimes are comparable; mild concentration appears in the synthetic top-50 tail.}
\label{tab:templates}
\end{table}

\FloatBarrier
\section{Embedding-Space Details}
\label{sec:appendix-embeddings}

\begin{figure*}[t]
\centering
\includegraphics[width=\textwidth]{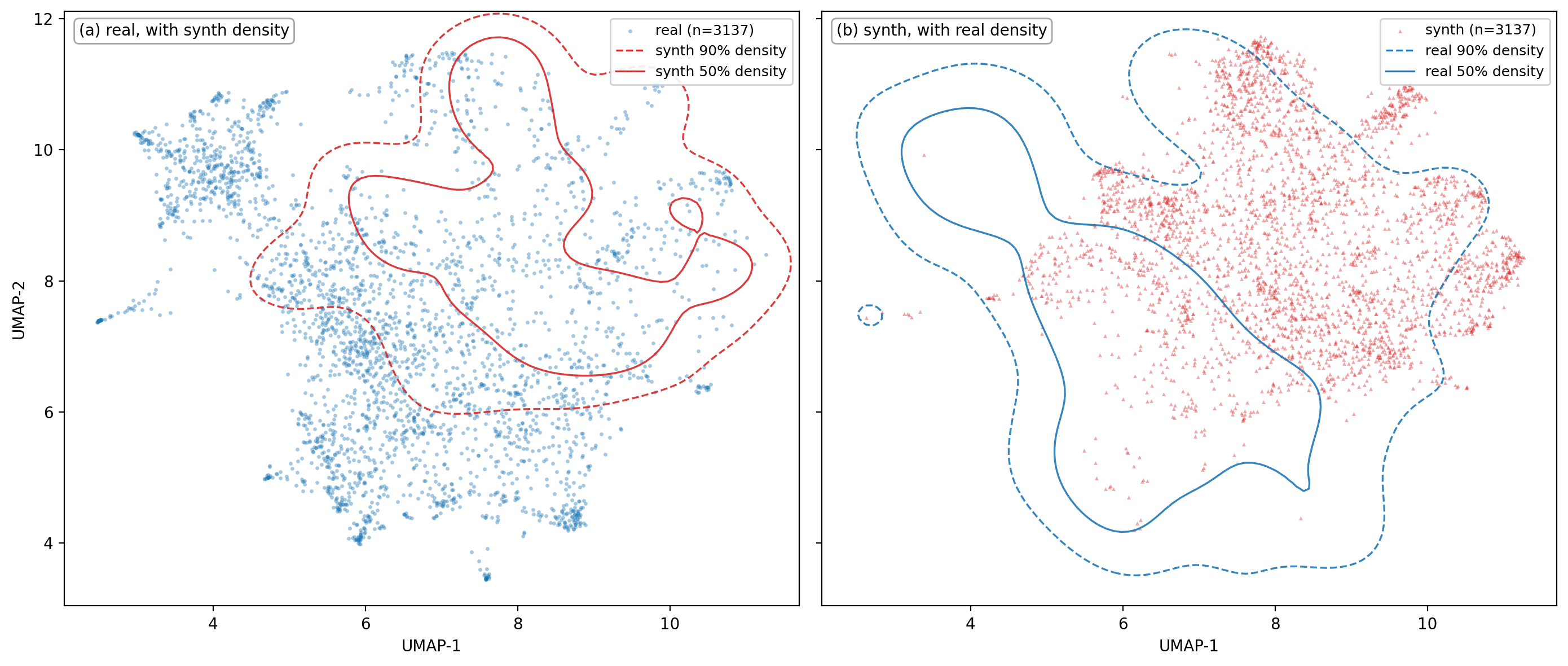}
\caption{Two-panel UMAP under \texttt{sentence-transformers/all-mpnet-base-v2} (768-d), shared UMAP fit on $n{=}3{,}137$ each. \textbf{(a)}~Real points (blue) with synthetic density contours; \textbf{(b)}~synthetic points (red) with real density contours. Both encoders show a consistent distributional gap; metric evidence in Table~\ref{tab:embedmpnet}.}
\label{fig:partcmpnet}
\end{figure*}

This appendix expands the compressed Part C summary in main with per-encoder tables (BGE-Large: Table~\ref{tab:embedbge}; MPNet: Table~\ref{tab:embedmpnet}) and a two-panel UMAP (Figure~\ref{fig:partcmpnet}). Running two encoders ensures that conclusions are not artifacts of either geometry alone; the encoder-consistent distributional gap (MMD$^2$, classifier AUC, precision/recall/coverage) across both tables is the intended evidence pattern.

\paragraph{Real-vs-real precision/recall/coverage control.} The Kynk\"a\"anniemi numbers are not read against a $1.0$ ceiling: finite samples of \emph{real} notes do not achieve perfect mutual support either. The control draws two disjoint $2{,}000$-note real samples ($10$ repetitions) and computes the identical $k{=}5$ manifold precision/recall/coverage between them: BGE-Large $0.806{\pm}.011\,/\,0.800{\pm}.006\,/\,0.971{\pm}.004$ and MPNet $0.847{\pm}.011\,/\,0.843{\pm}.007\,/\,0.972{\pm}.003$. Relative to these reference levels, synthetic precision reaches $86\%$ (BGE) and $98\%$ (MPNet), while recall reaches $18\%$ and $26\%$: local-support retention is substantially stronger than manifold coverage, without implying note-level indistinguishability. Artifact: \texttt{outputs/\allowbreak evaluation/\allowbreak size\_matched\_\allowbreak controls.json}.

\paragraph{What drives the classifier AUC.} Three diagnostics localize the separability signal under the identical size-matched protocol. First, note length does not explain it: within each combined-corpus word-count quartile the AUC stays between $0.95$ and $0.99$ (BGE-Large), and re-running the full classifier against a real pool resampled to match the synthetic word-count histogram leaves the AUC unchanged ($0.965$ BGE-Large, $0.939$ MPNet, vs.\ $0.966$/$0.938$ unmatched). Second, shallow surface style carries much of the signal: a logistic classifier on $16$ hand-crafted surface features (length, punctuation rates, all-caps token rate, sentence-length variance --- no content representation) reaches AUC $0.822$, while word count alone is near chance ($0.554$). The remaining gap to the encoder AUC is residual signal captured by the sentence representation; this diagnostic cannot separate content from unmeasured style. The practical reading for calibration work is that register-level surface cues are one recoverable component of separability, consistent with the practitioner pilot's ``too polished'' authenticity cues. Artifact: \texttt{outputs/\allowbreak evaluation/\allowbreak part\_c/\allowbreak auc\_probe.json}.

\begin{table}[t]
\centering
\small
\begin{tabular}{lr}
\toprule
\textbf{Metric (BGE-Large)} & \textbf{Value} \\
\midrule
MMD$^2$ & 0.02430 \\
Fr{\'e}chet distance, PCA$\rightarrow$256 & 0.0778 \\
LogReg AUC & 0.966 [0.963, 0.968] \\
GBT AUC & 0.969 [0.966, 0.971] \\
Precision & 0.690 \\
Recall & 0.144 \\
Coverage & 0.566 \\
\bottomrule
\end{tabular}
\caption{Part C embedding metrics under \texttt{BAAI/bge-large-en-v1.5} on the v1.0 corpus ($n_\text{synth}{=}3{,}137$).}
\label{tab:embedbge}
\end{table}

\begin{table}[t]
\centering
\small
\begin{tabular}{lr}
\toprule
\textbf{Metric (MPNet)} & \textbf{Value} \\
\midrule
MMD$^2$ & 0.02523 \\
Fr{\'e}chet distance, PCA$\rightarrow$256 & 0.1198 \\
LogReg AUC & 0.938 [0.935, 0.943] \\
GBT AUC & 0.949 [0.947, 0.953] \\
Precision & 0.829 \\
Recall & 0.223 \\
Coverage & 0.604 \\
\bottomrule
\end{tabular}
\caption{Part C embedding metrics under \texttt{sentence-transformers/all-mpnet-base-v2} on the v1.0 corpus ($n_\text{synth}{=}3{,}137$).}
\label{tab:embedmpnet}
\end{table}

\FloatBarrier
\section{Practitioner Pilot --- Full Report}
\label{sec:appendix-practitioner}

This appendix expands the compressed \S\ref{sec:practitioner} into the full descriptive writeup.

\paragraph{Setting and protocol.} We ran the pre-specified $N=8$ caseworker pilot to its planned completion: $8$ of $8$ remote single-facilitator sessions, each scheduled for 60 minutes, across school counseling (P1), hospital social work (P2), pediatric medical (P3), county mental-health-court diversion (P4), outpatient digestive-health clinical supervision (P5), a case-management team lead at the source organization (P6), community-college disability services (P7), and higher-education student basic-needs (P8). The fixed review packet contained a schema review, two chronological note sets, two pairwise comparisons, and a three-note Turing task. It included synthetic notes and a redacted real note for the authenticity task.

\paragraph{Schema endorsement.} All $8/8$ participants endorsed the fact schema as comprehensive for their setting; P7 and P8 independently rated it $9/10$. Critiques targeted over-specificity for individual settings (\emph{e.g.}, hospital, pediatric) rather than missing fields.

\paragraph{Trajectory plausibility.} Two synthetic note sets were compared. The structured-fact-driven Set~B was preferred over the briefer Set~A by $7/8$ practitioners; the one dissent (P7) preferred Set~A's brevity on protective-minimalism grounds in disability/LGBTQI/immigration services rather than on fidelity grounds. Set~A drew trajectory-coherence flags from $4$ of $8$ practitioners (lack of inter-note referencing, presenting reason landing in the third note rather than the first).

\paragraph{Turing task.} Aggregate accuracy was $15/24 = 62.5\%$, sitting inside the pre-specified $33$--$67\%$ indistinguishability window. The per-note breakdown reveals a strong note-length asymmetry: the long synthetic note (152 words) was correctly identified as synthetic by $7/8$; the long real note (134 words) by $7/8$; but the short synthetic note (73 words) was misclassified as real by $7/8$ --- only P1 detected it. Four practitioners independently articulated the same convergent heuristic (``AI is polished / over-generates''), which correctly handles long synthetic notes but systematically fails on short clipped synthetic ones, exactly the failure mode the Note-3 result exposes. This is an authenticity confound, not a fidelity claim: short synthetic notes happen to live in a region the practitioner heuristic cannot distinguish.

\paragraph{Use-case endorsements.} Eight use cases emerged with attribution-tracked counts (Table~\ref{tab:usecases}). Training / new-employee education was endorsed by all $8/8$ participants. The strongest secondary uses were trend-and-pattern analysis ($4/8$) and fundraising/grant reporting ($3/8$, with one explicit \emph{dis}-endorsement on validity grounds). Court/legal documentation drew $3/8$ and documentation aids drew $3/8$. P8 surfaced three new categories not anticipated in the methodology: substitute-caseworker handoff, open-records / records-request transparency exemplars, and embedding into employee-manual reference exemplars. \textbf{Zero of eight} participants endorsed operational use with real clients.

\begin{table}[!htbp]
\centering
\small
\setlength{\tabcolsep}{3pt}
\begin{tabular}{lr}
\toprule
\textbf{Use case} & \textbf{Endorsed} \\
\midrule
Training / new-employee education       & 8/8 \\
Trend / pattern analysis (policy)       & 4/8 \\
Fundraising / grant reporting           & 3/8$^{*}$ \\
Court / legal documentation             & 3/8 \\
Documentation aids                      & 3/8 \\
Substitute-caseworker handoff           & 1/8 \\
Open-records / records-request exemplars & 1/8 \\
Employee-manual exemplars               & 1/8 \\
Clinical use against real clients       & \textbf{0/8} \\
\bottomrule
\end{tabular}
\caption{Practitioner-endorsed use cases ($N=8$, attribution-tracked). $^{*}$1 dis-endorsement.}
\label{tab:usecases}
\end{table}

\paragraph{Concrete defects flagged.} Practitioners named five surface-language artifacts (\emph{e.g.}, ``client arrived irritable,'' ``remained appropriate and focused,'' ``in plain terms / on scrap paper,'' ``still waiting on movement,'' ``no barriers today, forms go through''), two process-validity errors (a rapid-rehousing referral without coordinated entry; an acute episode co-occurring with ID-paperwork tasks), one trajectory-chronology error (Set~A's presenting reason landing in note 3), and one pronoun-asymmetry observation (Set~B uses he/his while Set~A does not) worth full-corpus auditing. We discuss the corpus-specific all-caps resource convention as an authenticity confound rather than a generation defect (limitation H1).

\paragraph{Tier-ranked use cases.} Re-coding the eight endorsed use cases and adjacent practitioner-supported applications against the strength of their cross-participant support yields three tiers. \emph{Tier 1 (strongest fit)} comprises (i)~training, onboarding, and classroom instruction (P3, P4, P5, P6, P7, P8); (ii)~documentation-quality benchmarking and standardization (P3, P4, P5, P6, P7, P8); (iii)~handoff and substitute-caseworker continuity simulation (cross-participant support comes from the longitudinal-continuity practitioner finding endorsed by P1, P3, P4, P5, P8 = $5/8$; the explicit ``substitute-caseworker handoff'' use phrase in Table~\ref{tab:usecases} was named only by P8 = $1/8$, so we anchor Tier-1 placement in the conceptual substrate rather than in the surfaced use-case label); (iv)~next-step recoverability and actionability evaluation (P1, P2, P3, P4, P5, P8); and (v)~longitudinal summarization / ``living-document'' generation (P3, P5, P8 explicitly; P1, P4 implicitly). \emph{Tier 2 (promising but qualified)} covers aggregate reporting / grant-metric prototyping (P1, P2, P5, P6, P7), system-level needs assessment (P1, P5, P6, P7, P8), synthetic vignettes for research and policy discussion (P4, P5, P7, P8), and NLP benchmarking (a strong research use, but practitioners named it only indirectly). \emph{Tier 3 (adjacent / future work)} includes special-education and mediation documentation training (P8) and synthetic illustrative narratives for advocacy (P4 with caveats). The strongest invariant across tiers is that practitioner endorsements are framed as \emph{evaluation and training} uses, not as substitutes for real records --- consistent with $0/8$ participants endorsing operational use with real clients.

\paragraph{Useful-note structure.} Across interviews, useful notes shared a consistent seven-element structure: (1) reason for contact / why the worker reached out (P2, P4, P8); (2) client-reported information, preferably attributed (P2, P4, P8); (3) caseworker action --- what was done, offered, explained, or coordinated (P1, P2, P3, P5, P6, P8); (4) client response --- accepted, declined, agreed, missed, followed through (P1, P2, P4, P5, P8); (5) barriers (missing documents, transportation, waitlists, eligibility, legal issues) (P1, P2, P4, P5, P8); (6) next steps for client and worker (P2, P3, P4, P5, P8); and (7) longitudinal continuity --- how the current note relates to prior notes (P1, P3, P4, P5, P8). This list maps onto schema fields used by the structured generation pipeline (\S\ref{sec:source-constraints}) and aligns with several Part E concrete fields that were well-grounded in prose (\S\ref{sec:parte}: \texttt{resource\_types} $81\%$, \texttt{document\_types} $82\%$, \texttt{housing\_status} $80\%$, \texttt{contacted\_parties} $77\%$, \texttt{issues\_reported} $69\%$, \texttt{services\_provided} $57\%$, \texttt{employment\_status} $55\%$; pooled $68.7\%$). Low lexical recall for \texttt{visit\_type} ($21\%$) may reflect latent phrasing, but \texttt{next\_step\_type} ($44\%$) intersects a practitioner-identified load-bearing element and remains a fidelity concern.

\paragraph{Controlled imperfection.} Practitioners did not expect real notes to be polished. Realistic notes routinely include abbreviations, concise phrasing, inconsistent grammar, missing punctuation, short fragments, agency-specific wording, and variable detail across workers (P2, P4, P5, P8). P8 explicitly reasoned that grammar mistakes and dropped apostrophes \emph{signal} human-authored notes because caseworkers are busy and focused on capturing the interaction; P1 and P5 made related observations from child-welfare documentation experience. This grounds deliberate post-generation grammar degradation and prohibited-phrase pass-through in practitioner reasoning rather than only in distributional metrics: ``too clean'' is not just a stylistic mismatch, it is a detectability cue that practitioners use to classify a note as synthetic.

\section{No-Copy / Overlap Audit (Extended)}
\label{sec:appendix-nocopy}

Two programmatic audits test the structural privacy story. The \emph{few-shot leak} audit scores all $18{,}822$ synthetic-note $\times$ used-few-shot-example pairs for both contiguous shared-token runs and 10-gram Jaccard overlap. The \emph{real-corpus phrase reuse} audit compares every unique 10-gram in the synthetic corpus against all 10-grams in the $63{,}142$-note real corpus.

For few-shot leakage, $0$ pairs cross either the $\geq 10$-token run threshold or the Jaccard $\geq 0.10$ threshold across all $18{,}822$ scored synth/example pairs; the longest contiguous shared run anywhere in the corpus is $8$ tokens. Per-note maximum shared run quantiles are q50 $=2$, q90 $=4$, q95 $=5$, q99 $=5$ tokens. For real-corpus phrase reuse, the $3{,}137$ synthetic notes contain $210{,}039$ unique 10-grams; only $13$ of these intersect any 10-gram in the real corpus ($6.2\times 10^{-5}$), touching $14$ real notes ($2.2\times 10^{-4}$ of the real corpus). The full set is enumerated in Table~\ref{tab:overlap-boilerplate}; it spans voicemail, meeting, bus-pass, check-in, and disposition templates, and a single bus-pass 10-gram contributes $5$ of the $17$ raw hits across $5$ real notes.

\paragraph{Direct-prompt baseline comparison.} The same audit on the existing direct-prompt baseline corpus ($162$ notes, $10{,}419$ unique 10-grams) finds $0$ real-corpus 10-gram intersections. The baseline has no few-shot rotation and therefore no \texttt{*\_examples.json} provenance, so the few-shot-leak sub-audit is skipped for the baseline (no provenance pool exists to leak from). The baseline result is reported as a directional check, not as a fraction comparable to the v1.0 value of $13 / 210{,}039$: the v1.0 corpus unique-10-gram pool is $\sim\!20\times$ larger than the baseline pool, so the v1.0 fraction is a larger-denominator estimate, not a worse outcome.

\begin{table*}[t]
\centering
\small
\setlength{\tabcolsep}{4pt}
\begin{tabular}{lp{0.62\textwidth}r}
\toprule
\textbf{Category} & \textbf{10-gram (verbatim, post-redaction)} & \textbf{$n_{\text{real}}$} \\
\midrule
Voicemail   & case manager called and left a message requesting a return & 1 \\
Voicemail   & manager called and left a message requesting a return call & 1 \\
Meeting     & case manager met with in the office to check in            & 1 \\
Meeting     & manager met with in the office to check in on              & 1 \\
Meeting     & case manager met with client at for follow up client       & 1 \\
Bus pass    & a bus pass to get back to his camp case                    & 1 \\
Bus pass    & bus pass to get back to his camp case manager              & 1 \\
Bus pass    & client asked for a bus pass to get to a                    & 1 \\
Bus pass    & client was given a 1-day bus pass to get to                & 1 \\
Bus pass    & week client was given one 7-day bus pass and one           & 5 \\
Check-in    & he is still looking for work and has been using            & 1 \\
Check-in    & client called to check in and said he is still             & 1 \\
Disposition & not ready to move forward at this time case manager        & 1 \\
\bottomrule
\end{tabular}
\caption{The $13$ unique synthetic 10-grams from the v1.0 corpus that intersect the redacted real corpus ($14$ real notes / $17$ raw hits; rate $6.2\times 10^{-5}$). All are domain boilerplate. Visible gaps (\emph{e.g.}, ``met with in the office'') are post-redaction artifacts where the NER+regex pass removed a name or place inside the span. $n_{\text{real}}$ = count of real notes containing each 10-gram.}
\label{tab:overlap-boilerplate}
\end{table*}

We interpret these results as evidence against memorization or paraphrase-level copying, not as a formal privacy guarantee. The structural boundary (unredacted source notes, source identifier records, and real fact trajectories stay local; the PII-redacted exemplar pool is never released), the Part D verbatim-overlap audit, and the Part G fact and prose MIA sketches (both below the project-defined $0.55$ follow-up threshold) together form a defense-in-depth posture. A fuller PromptEHR-style suite---shadow-training MIA plus attribute inference through a synthetic-trained imputation adversary \citep{wang2022promptehr}---and scaled MIA \citep{puerto2025scaling} remain future work.

\section{Fact-to-Text Alignment (Extended)}
\label{sec:appendix-alignment}

A fourth audit checks whether the structured synthetic facts that condition each generation actually surface in the resulting prose. We frame the audit as \emph{schema-conditional fact recall}: every populated schema field becomes a presence test against the prose, with the rubric auto-derived from the structured facts that conditioned generation rather than hand-curated, so the test is trivially extensible whenever the schema gains a field (related to checklist-style summarization evaluation more broadly \citep{dou2026gavel,zhou2025feedback}). For every populated enum-valued field in a synthetic fact JSON, we test whether any surface form (canonical token, schema-synonym map inversion, or a hand-curated extras table) appears as a substring of the case-folded note text. Implicit expression that does not surface a literal token (\emph{e.g.}, communicating ``triage'' through phrasing rather than the word) is counted as missing.

\paragraph{Recall vs.\ precision (what this audit does and does not measure).} The lexical audit is recall-only: for each populated field, did the prose surface it? The symmetric \emph{precision} question --- did the prose claim a value the structured facts do not contain, or claim a value at a different intensity than the facts indicate? --- is partially addressed by the automatic LLM-judge diagnostic below, which labels populated pairs as explicit, paraphrased, omitted, or contradicted. Prose claims outside the populated schema fields remain unmeasured, and the judge itself lacks human-gold validation; both gaps require a human-coded subset on the order of $N{\approx}100$ notes. The other side of the same trade-off: practitioners (\S\ref{sec:practitioner}, P3, P5, P8) endorsed the schema as comprehensive, so a populated-field checklist is a reasonable proxy for ``what should the prose say,'' but they did not endorse a closed-world assumption on the prose.

Across all $3{,}137$ synthetic notes in the v1.0 corpus and $17{,}722$ populated fields, the headline numbers are: \textbf{concrete-field recall $68.7\%$} ($6{,}762 / 9{,}848$), \textbf{soft-field recall $41.3\%$} ($3{,}250 / 7{,}874$), and overall pooled recall $56.5\%$ ($10{,}012 / 17{,}722$). Per-note overall recall has q50 $=57\%$, q90 $=83\%$, and q95 $=100\%$; $162$ notes ($5.2\%$) have perfect recall on every populated field. Table~\ref{tab:partefield} gives the per-field breakdown.

\begin{table}[t]
\centering
\small
\setlength{\tabcolsep}{3pt}
\begin{tabular}{llrr}
\toprule
\textbf{Kind} & \textbf{Field} & \textbf{$n_\text{pop}$} & \textbf{Recall} \\
\midrule
Concrete & \texttt{document\_types}      & 1{,}241 & 81.9\% \\
Concrete & \texttt{resource\_types}      & 1{,}291 & 81.4\% \\
Concrete & \texttt{housing\_status}      &   852 & 79.9\% \\
Concrete & \texttt{contacted\_parties}   &   237 & 76.8\% \\
Concrete & \texttt{issues\_reported}     & 2{,}380 & 68.6\% \\
Concrete & \texttt{services\_provided}   & 3{,}544 & 57.3\% \\
Concrete & \texttt{employment\_status}   &   303 & 55.4\% \\
\midrule
\textbf{Concrete pooled} & --- & \textbf{9{,}848} & \textbf{68.7\%} \\
\midrule
Soft & \texttt{visit\_location}          & 1{,}738 & 60.1\% \\
Soft & \texttt{contact\_method}          & 2{,}414 & 53.4\% \\
Soft & \texttt{next\_step\_type}         &   514 & 43.8\% \\
Soft & \texttt{crisis\_type}             &   143 & 39.2\% \\
Soft & \texttt{visit\_type}              & 3{,}065 & 20.7\% \\
\midrule
\textbf{Soft pooled} & --- & \textbf{7{,}874} & \textbf{41.3\%} \\
\midrule
\textbf{Overall pooled} & --- & \textbf{17{,}722} & \textbf{56.5\%} \\
\bottomrule
\end{tabular}
\caption{Per-field lexical recall over $3{,}137$ synthetic notes in the v1.0 corpus, split into concrete (verbatim-expected) and soft / latent-shaper fields. Concrete-field pooled recall is the primary grounding measure; soft-field recall is reported separately because these fields legitimately shape prose without appearing verbatim.}
\label{tab:partefield}
\end{table}

\paragraph{LLM-judge grounding audit --- design.} A second, semantic reading of the same audit re-scores every populated (field, value) pair with an LLM judge that reads the full note and returns one of four verdicts per fact: \emph{explicit} (stated near-verbatim or via a direct synonym), \emph{paraphrased} (conveyed by meaning but never named), \emph{omitted}, or \emph{contradicted} (the note states something incompatible with the value). The judge is \texttt{gpt-5.6-terra} --- a different OpenAI model from the \texttt{gpt-5.2} generator, softening self-preference bias --- with low reasoning effort, JSON-schema output, and one call per note; verdict-schema compliance is $100\%$ over all $3{,}137$ notes. The judge sees only synthetic fact JSON and synthetic note text, both of which crossed the API boundary at generation time, so no real-derived text is involved. Judge-grounded means explicit or paraphrased.

\paragraph{LLM-judge grounding audit --- results.} Over the identical $17{,}722$ pairs, the judge labels \textbf{$91.7\%$} of concrete-field pairs grounded (vs.\ $68.7\%$ lexical), $68.2\%$ of soft-field pairs grounded (vs.\ $41.3\%$), and $81.3\%$ grounded overall (vs.\ $56.5\%$). It labels $680$ pairs ($3.8\%$) \emph{contradicted}, concentrated in the soft fields \texttt{visit\_type} ($502$) and \texttt{visit\_location} ($142$); every concrete field has a judge-labeled contradiction rate ${\leq}0.5\%$. These are automatic diagnostic labels under the judge's rubric, not human-gold estimates. The audit does not close limitation~M8 because it checks only populated fact-JSON fields and cannot detect prose claims outside the schema, or limitation~M9 because the judge has not been human-validated on this corpus. Artifact: \texttt{outputs/\allowbreak evaluation/\allowbreak part\_e/\allowbreak llm\_judge\_report.json}; driver: \texttt{python -m pipeline.tools.llm\_judge \allowbreak --seed 42} (the module lives outside \texttt{pipeline/\allowbreak evaluation/\allowbreak part\_e/} because evaluation Parts do not import external-API clients).

\paragraph{Judge-vs-lexical failure taxonomy.} Crossing the two audits on identical pairs buckets every disagreement (Table~\ref{tab:judgetaxonomy}). The two dominant buckets are pairs the judge labels grounded but the lexical matcher misses: \emph{explicit beyond the synonym map} ($2{,}994$ --- the value is named in the prose through a surface form the synonym map lacks) and \emph{paraphrase beyond lexical} ($1{,}592$ --- conveyed by meaning only). Under the judge's rubric, both are consistent with lexical under-recall (M4). The reverse direction is small: $129$ \emph{spurious lexical matches} (a substring hit the judge labels as not conveying the value) and $64$ lexical hits the judge labels contradicted, together $1.1\%$ of all pairs.

\begin{table}[!htbp]
\centering
\small
\setlength{\tabcolsep}{3pt}
\begin{tabular}{lrr}
\toprule
\textbf{Bucket (lexical $\times$ judge)} & \textbf{$n$} & \textbf{\%} \\
\midrule
\multicolumn{3}{l}{\textit{Agreement ($12{,}943$ pairs)}} \\
\quad hit $\times$ grounded          & 9{,}819 & 55.4 \\
\quad miss $\times$ omitted          & 2{,}508 & 14.2 \\
\quad miss $\times$ contradicted     &   616 &  3.5 \\
\midrule
\multicolumn{3}{l}{\textit{Disagreement ($4{,}779$ pairs)}} \\
\quad miss $\times$ explicit (beyond synonym map) & 2{,}994 & 16.9 \\
\quad miss $\times$ paraphrased (beyond lexical)  & 1{,}592 &  9.0 \\
\quad hit $\times$ omitted (spurious match)       &   129 &  0.7 \\
\quad hit $\times$ contradicted                   &    64 &  0.4 \\
\bottomrule
\end{tabular}
\caption{Failure taxonomy crossing lexical hits and misses with automatic LLM-judge labels on the same $17{,}722$ populated pairs. Most disagreements are pairs the judge labels grounded but the lexical matcher misses.}
\label{tab:judgetaxonomy}
\end{table}

\section{Extended Discussion}
\label{sec:appendix-discussion}

\paragraph{Utility rather than indistinguishability.} The central result is not that synthetic notes are indistinguishable from real notes. They are not. Multi-note client trajectories are distinguishable under balanced 1-NN and matched logistic AUC. Note embeddings are separable under both BGE-Large and MPNet. These results rule out claims of distributional equivalence. However, indistinguishability is not the only relevant standard. The intended use is not to replace real data, but to provide a documented, shareable corpus for method development, training, and benchmark construction under privacy constraints. A high classifier AUC undermines claims of equivalence; it does not by itself undermine bounded utility. The appropriate interpretation is that the synthetic corpus is a high-precision, lower-recall resource. Many synthetic notes lie in regions associated with real notes, and the corpus preserves broad documentation structure and template diversity. At the same time, it covers only part of the real distribution and retains systematic artifacts.

\paragraph{A practitioner-derived utility lens.} The eight practitioner interviews supply a complementary lens on what ``utility'' means for synthetic case notes (\S\ref{sec:practitioner}). The strongest endorsed uses --- training and onboarding without exposing real clients, documentation-quality benchmarking, handoff, and substitute-caseworker simulation, next-step recovery, longitudinal summarization, and NLP benchmarking for extraction / retrieval / next-step prediction --- share a common shape: they are evaluation and training uses, not substitutes for real records. Practitioners did not evaluate notes only by surface realism. They evaluated them by whether the notes preserved the longitudinal, actionable, and ethically constrained structure that another reader (a substitute caseworker, a supervisor, an auditor, a court, an open-records requester) would need. Three observations follow. First, useful documentation is selective, not maximal: more detail is not always better, and over-documentation of legal status, immigration, mental health, or substance use creates downstream disclosure risk (P2, P3, P4, P7, P8). Second, controlled imperfection is part of realism: notes that are too polished read as synthetic to practitioners (P4, P8). Third, and most centrally for this paper's claim, synthetic longitudinal case notes are most appropriate where statistical realism \emph{supports} practical reasoning, and least appropriate for tasks requiring factual correspondence to a real person, real event, or real service outcome.

\paragraph{Why longitudinal structure matters.} Many synthetic text datasets generate independent examples. Case notes are different because their value often appears across sequences. A single note may document a need, but a trajectory shows whether that need persisted, whether follow-up occurred, whether barriers recurred, and whether the case stabilized or escalated. This has methodological consequences. Matching note-level style is not enough. Synthetic case-note generation should model first-note priors, time gaps, transitions, stability, persistence, and recurrence. The pipeline's structured trajectory sampling directly addresses this need, and the evaluation separates multi-note and single-note cases because they support different claims.

\paragraph{What the corpus enables.} The corpus enables early-stage research and evaluation in a domain where real data is difficult to share. Because notes are paired with structured synthetic facts, users can construct supervised or weakly supervised tasks such as primary-need classification, referral extraction, barrier identification, service/action extraction, status prediction, and trajectory summarization. Because notes are grouped into synthetic client trajectories, the corpus also supports handoff summarization, retrieval over case histories, and longitudinal reasoning experiments. The corpus is also useful for education and training. Synthetic notes can expose students, interns, annotators, or new practitioners to realistic documentation patterns without exposing real clients. This use does not require perfect indistinguishability; it requires realistic enough variation to support practice, comparison, and feedback.

\paragraph{What the corpus does not enable.} The corpus should not be used to make decisions about real people, infer real service outcomes, evaluate actual caseworker performance, or estimate population-level statistics about the source agency. It should not be treated as a formal privacy guarantee. It should not be used as the only validation set for deployment-ready systems. The corpus is a synthetic research artifact. Its value lies in providing shareable case-note-like data for bounded research and evaluation uses, not in replacing real operational records.

\paragraph{Why structured-first beats direct prompting.} The Part B JSD ($4.1\times$) and Part C MMD$^2$ ($7.1\times$) gaps show that the complete structured-first stack is closer to the real corpus than this sparse direct-prompt baseline under the same OpenAI model and documentation-style instruction; the structured pipeline additionally receives redacted style exemplars (the ablation-scope caveat in \S\ref{sec:baseline-summary}), so the experiment does not isolate fact conditioning from exemplar conditioning. A reading consistent with the design is that the structured-first pipeline gives the generator a fixed, corpus-aligned fact bundle to verbalize, whereas the direct-prompt baseline must infer both the latent fact bundle and prose surface; the observed Part B/C metrics show greater homogenization in the baseline but do not identify a single causal component. The size-matched paired difference of $+0.050$ in Part F transfer ratio (\S\ref{sec:partf}) is consistent with the same stack-level gap.

\paragraph{Sparse-class failures and targeted expansion.} The two long-tail labels that drop to ratio $0.0$ (\texttt{childcare\_family}) and $0.58$ (\texttt{safety\_domestic\_violence}) are sparsely represented in the v1.0 training pool. A targeted-expansion ablation ($30$ category-conditioned clients biased $0.4 / 0.4$ toward the two flagged labels, with a persistence-floor boost so the niche thread propagates across each trajectory --- documented in Appendix~\ref{sec:appendix-transfer} but not part of the fixed corpus) improves both labels within the same eval split. The structured-first pipeline's advantage as a corpus-construction tool is exactly this composability: coverage of a sparse label can be improved by reallocating synthesis budget, without retraining the generator. For these two labels, the observed gaps respond to targeted coverage within this evaluation split.

\paragraph{Why soft / latent-shaper fields go under-surfaced by design.} Part E (\S\ref{sec:parte}; per-field detail in Appendix~\ref{sec:appendix-alignment}) reports lexical fact recall of $21\%$ for \texttt{visit\_type} and $44\%$ for \texttt{next\_step\_type}; these fields are intended to shape how the prose reads. The schema separates two field classes: \emph{concrete} fields (\texttt{issues\_reported}, \texttt{services\_provided}, \texttt{document\_types}, etc.) whose values should appear verbatim because they record what actually happened, and \emph{soft / latent-shaper} fields (\texttt{visit\_type}, \texttt{next\_step\_type}, \texttt{crisis\_type}, etc.) whose values shape \emph{how} the prose reads (a \texttt{triage} visit and an \texttt{intake} visit produce different opener patterns and tense distributions even when no literal token ``triage'' appears). Practitioners corroborated this split: P4 and P8 named what we call concrete fields as load-bearing for handoff while noting that \texttt{visit\_type} labels are rarely written verbatim in real notes either (\S\ref{sec:practitioner}). The right test for soft-field fidelity is therefore high upstream Part B / Part C similarity, not per-field substring match.

\section{Full Limitations Catalog}
\label{sec:appendix-limitations}

This appendix expands the categorized summary in the limitations section into the full per-item enumeration. The keys (M, C, P, Pr, G, H) match the limitations catalog for cross-reference.

\paragraph{Methodological.}
\textbf{M1 (encoder dependence).} Part C MMD, Fr\'echet, and classifier-AUC numbers are encoder-specific to BGE-Large and MPNet. Absolute Fr\'echet ($0.0778$ vs.\ $0.1198$) is not comparable to prior work using different encoders. \emph{Mitigation:} dual encoders for both Part C distribution-similarity metrics and the Part F downstream-task transfer head (the latter recovers transfer ratio $0.82\text{--}0.83$ across BGE-Large and MPNet); only conclusions consistent across both encoders are encoder-robust.
\textbf{M2 (zero-shot labeler granularity).} Part B uses an embedding-prototype cosine-argmax classifier rather than human-coded labels. Mean overall margin is $0.027$; per-label margins range $0.046$ down to $0.016$ (Appendix~\ref{sec:partb-margins}). Borderline labels (\texttt{client\_report}/\texttt{presenting\_need}; \texttt{other}/\texttt{background\_context}) carry most of the residual classification noise.
\textbf{M3 (single extractor LLM).} Extraction uses only \texttt{qwen3:32b-q8\_0}. Systematic extraction biases (under-extraction of crisis content, over-confidence on ambiguous notes) are not quantified and propagate into every downstream stage. \emph{Future work:} ensemble extraction with a second local LLM and inter-rater agreement on a sampled subset.
\textbf{M4 (Part E lexical recall misses implicit expression).} Lexical-substring matching under-recalls implicit expression: a \texttt{visit\_type=triage} note may communicate ``triage'' through phrasing without the literal word and is counted as missing. The pooled lexical recall is $56.5\%$ overall (Appendix~\ref{sec:appendix-alignment}); the primary grounding measure in the main text is the concrete-field pooled recall of $68.7\%$, where verbatim mention is expected. Under its rubric, the automatic LLM judge labels concrete grounding at $91.7\%$ and overall grounding at $81.3\%$, consistent with lexical under-recall; a human-coded gold subset remains the outstanding validation (M9). Lexical matches can also be spurious or contradicted.
\textbf{M5 (weak K-means structure).} Best-$k$ silhouette is $0.157$ at $k=4$; cluster results are descriptive, not confirmatory. The headline coverage metrics (Lopez-Paz, matched AUC, NN distances) do not depend on cluster structure.
\textbf{M6 (severity and engagement are interpretive).} Practitioners (P1, P2, P7) noted that fields like \texttt{issue\_severity} and engagement-level encode whose lens is being used --- agency, clinician, funder, or client --- rather than an observer-independent ground truth. Distributional fidelity on these fields therefore measures consensus on a documented judgment, not factual correctness. P1 specifically suggested separating \emph{amount} of engagement from \emph{quality} of engagement; P2 noted that severity is hard to define for housing or legal issues; P7 emphasized that the same situation can be ``moderate'' to a worker and ``severe'' to the client. Future work could split these fields into separate documented-judgment vs.\ behavioral-fact axes.
\textbf{M7 (Part F silver labels share an extractor with the synthetic corpus).} Part F real train/eval labels are derived from \texttt{presenting\_issues.issues\_reported[0]} in locally extracted real facts. Structured-synthetic labels use that field in the sampled synthetic conditioning facts, whose distributions were learned from the extracted real corpus. Both the synthetic-train and real-eval label distributions therefore share the extractor's biases on the same axis the eval scores against, and the $0.83$ transfer ratio measures transfer \emph{to the extractor's notion of primary need} rather than to a human-coded gold standard. Using a shared label schema makes the comparison internally consistent, but it does not fully de-risk the cross-condition ordering: because the extractor's outputs also shaped the learned sampling distributions, the synthetic condition may align with the eval labels better than a human-gold standard would show, so both the absolute ratio and, potentially, part of the synth-vs-baseline margin are bounded by this design. An author-coded $n{=}250$ gold subset on the held-out real-eval slice checks the eval-side half of this exposure: silver--gold agreement is $0.688$ [$0.631, 0.745$], and re-scoring the real-trained and synth-trained logistic-regression heads against the author gold gives a gold-referenced transfer ratio of $0.84$ [$0.74, 0.94$], against the silver-referenced $0.83$ (Appendix~\ref{sec:appendix-transfer}, Table~\ref{tab:transfer-gold}). The residual exposure is the train side (training labels remain unchanged) and the single-coder design; a relabeling pass by a second extractor model remains open (Appendix~\ref{sec:appendix-futurework}).
\textbf{M8 (Part E has no full precision-side hallucination audit).} Part~E is framed as schema-conditional fact recall (\S\ref{sec:parte}, Appendix~\ref{sec:appendix-alignment}) and reports \emph{recall} of populated structured-fact fields in the generated prose. On the precision side, the automatic LLM judge labels $3.8\%$ of populated pairs \emph{contradicted} under its rubric (Appendix~\ref{sec:appendix-alignment}), but it checks only fields the fact JSON populates and cannot detect unsupported prose claims outside the schema. A full unsupported-claim audit and validation of the judge require a human-coded gold subset on the order of $N{\approx}100$ notes and are left as future work. Practitioners (\S\ref{sec:practitioner}, P3 / P4 / P8) also flagged process-validity errors qualitatively, motivating a future precision-side checklist.
\textbf{M9 (Part B and Part E lack human validation).} The Part~B sentence-level move labeler, the Part~E lexical fact-recall measure, and the Part~E LLM-judge re-audit are all fully automatic; none has been validated against a human-coded gold subset on this corpus. The practitioner pilot (\S\ref{sec:practitioner}, $N=8$) evaluates \emph{utility} (which uses the synthetic notes support, what schema fields look load-bearing, whether trajectories cohere) but not the validity of any automatic metric: practitioners did not score sentence-level move labels and did not check fact-to-text alignment per note. A second-wave validation would require a $200$-sentence Part~B subset with two annotators and a $100$-note Part~E subset with concrete-field annotation; both are deferred to future work (Appendix~\ref{sec:appendix-futurework}). Until then, Part~B / Part~E numbers should be read as automatic-metric distributional comparisons, not as human-aligned scores.

\paragraph{Corpus.}
\textbf{C1 (single organization).} The real corpus comes from one US-based agency. Generalization claims must be scoped to that setting.
\textbf{C2 (date semantics).} Source timestamps record documentation creation, which need not coincide with the encounter. Real \texttt{CREATIONDATETIME} spans 2016--2022. Synthetic trajectories start $6$--$24$ months before the sampling anchor and advance by sampled gaps. Case-duration shapes are comparable; absolute calendar timing is not. No claim in the paper relies on absolute calendar timing.
\textbf{C3 (class imbalance).} The synthetic corpus is $\sim 5\%$ the size of the real corpus on the prose side ($3{,}137 / 62{,}901$ notes; $87 / 1{,}681$ clients). Primary corpus comparisons use this release; auxiliary Part~F analyses are identified separately. \emph{Mitigation:} balanced 1-NN headline metric and size-matched subsampling for AUC where sample sizes differ.
\textbf{C4 (per-issue persistence on observed transitions).} Issue persistence is computed over consecutive-note pairs where the issue appears in the earlier note. An issue missing from the next note counts as non-persistence. Issues that are systematically under-extracted will look like ``issues that resolve quickly.''

\paragraph{Pipeline.}
\textbf{P1 (caseworker identity not modeled).} The sampler picks one writer style per client and holds it across the trajectory. It does not model multi-CM clients (handoffs) or CM-specific scenario tendencies (some CMs document more crisis events because they are on the crisis team). \emph{Future work:} explicit CM identity as a sampler axis.
\textbf{P2 (no email-thread sub-genre).} The real corpus contains $\sim 9\%$ email-thread style notes (quoted forwarded messages, signatures, subject-line fragments); the synthetic pipeline produces $0\%$ of this genre.
\textbf{P3/P4 (lumpiness, typos, numeric detail, writer switching).} Real notes have uneven topic emphasis within a single note, real typos beyond apostrophe drops, specific dollar amounts and phone-number-like fragments beyond \texttt{[PHONE]} placeholders, and within-client style drift across long trajectories. None are addressable by a single calibration knob.
\textbf{P5 (prose-quality calibration not scaled to full corpus).} The prose-quality calibration covers $25$ clients / $339$ notes rather than the full fixed corpus of $87$ clients / $3{,}137$ notes. A full-corpus prose assessment remains future work.

\paragraph{Privacy.}
\textbf{Pr1 (NER+regex residual leakage).} The redactor is high-precision-low-recall in this domain --- atypical proper nouns and unusual organization names can slip through. A residual identifier in the $146$-exemplar style pool \emph{can} reach the external synthesis API (API-exposure risk); Parts~D and~G audit release-side copying and membership, not residual identifiers in the prompts themselves. A manual residual-identifier review of all $146$ exemplars used for the fixed corpus found $0$ residual PII. Defense-in-depth is the small rotated pool, private exemplars excluded from release, and the Part~D no-copy audit over every v1.0 note against exactly the exemplars that conditioned it.
\textbf{Pr1b (privacy is broader than PII).} Practitioner interviews (P3, P7, P8) identified legal status, immigration, disability, mental health, substance use, gender identity, sexuality, and court involvement as sensitive even after PII removal. Notes can also be accessed via complaints, Title~IX review, or open-records processes (P8). The current redaction layer targets named-entity surface forms (names, organizations, phone numbers, addresses, emails) and does not implement per-attribute sensitivity controls. \emph{Future work:} attribute-level sensitivity tagging during extraction so downstream releases can selectively suppress or coarsen sensitive non-PII content.
\textbf{Pr2 (few-shot examples are real redacted text).} An earlier manual verbatim audit found $0/338$ notes with verbatim few-shot strings, and the Part D programmatic audit (\S\ref{sec:partd}) finds $0$ pairs above the $\geq 10$-token leak threshold across all $18{,}822$ note-example pairs in the current corpus.

\paragraph{Generalization.}
\textbf{G1 (English, US, community case management).} The corpus, schema, and prompts are entirely English and US-centric. The fact schema was developed against this corpus and may not transfer to clinical case notes, protective-services notes, non-US case management, or non-English documentation.
\textbf{G1b (setting-specific schema customization).} Although the eight practitioners spanned CPS, K--12 special education, hospital social work, outpatient pediatrics, mental-health-court diversion, higher-education disability services, basic-needs support, and case management --- and unanimously endorsed the fact schema as comprehensive --- they each named different setting-specific extensions: medical respite / skilled nursing / temporary housing nuance (P5), motivational-interviewing stages (P6), school counselors / teachers / coaches as collateral contacts (P7), and basic-needs / student-support categories (P8). The schema works as a broad superset; practical deployment in a different setting would require setting-specific customization rather than direct reuse.
\textbf{G2 (not tested cross-corpus).} The pipeline has not been run end-to-end on a different real corpus. \emph{Future work:} apply the pipeline (without re-tuning) to a publicly-available redacted case-note corpus and report transfer performance.

\paragraph{Practitioner evaluation (H1).} Eight interviews are descriptive only; the protocol pre-specifies narrative coding rather than inferential tests at this scale. \emph{The pilot evaluates utility, not the validity of any automatic metric}: practitioners reviewed schema adequacy, trajectory coherence, use cases, and a Turing-style authenticity task, but did not score Part~B sentence-level move labels and did not check Part~E fact-to-text alignment per note. The pilot therefore supports the bounded-utility headline (\S\ref{sec:practitioner}) and the schema-adequacy claim ($8/8$, \S\ref{sec:practitioner}) but does not human-validate the automatic Part~B / Part~E numbers --- that gap is captured separately as M9. Pair-comparison coverage was reduced under time pressure (Pair 1 reviewed by $2/8$, Pair 2 by $6/8$; schema review and the Turing task have full coverage). Recruitment skews toward source-organization adjacency (P4 and P6 inside the source-organization context; P7/P8 a snowball pair; only P1 fully outside both adjacency rings). The convergent ``AI is polished / over-generates'' heuristic articulated by $4/8$ is exploitable: it correctly identifies long synthetic and conversational real notes but systematically fails on short clipped synthetic notes (the failure mode that drove the near-universal Note-3 miss, $1/8$ correct). The all-caps resource convention is a corpus-specific authenticity confound that three participants treated as evidence of synthetic origin --- including one who therefore misclassified a real note. Future work: a second pilot wave with revised stimuli to test whether the Note-3 miss replicates under different short-note selection, plus a $200$-sentence Part~B and $100$-note Part~E human-coded subset to discharge M9.

\paragraph{Bounded-use coverage (Cov1).} Part~A registers high-precision / low-recall trajectory geometry ($29.6\%$ synth-precision / $4.3\%$ real-recall at the median real-real radius, \S\ref{sec:partacoverage}) and the field-frequency table (Appendix~\ref{sec:appendix-schema}, Table~\ref{tab:fieldfreq}) shows that the synthetic corpus systematically under-populates collateral ($-30$ to $-37$~pp), next-step ($-26$~pp), and temporal ($-59$~pp) fields at the structured-fact level. Together, these mean the corpus's bounded-utility claim is validated for \emph{training / onboarding / method-development / benchmarking} uses where high-precision case-note coverage of common scenarios is more important than full coverage of the real-trajectory tail and full prose-side coverage of every documentation pattern. It is \emph{not} validated for population-representative analysis, operational deployment, or any task that depends on the under-populated longitudinal fields above. In the practitioner pilot, $0/8$ endorsed operational use with real clients (\S\ref{sec:practitioner}). This is the same boundary surfaced from a different evidence stream; we read it as a positive honest-bounds finding rather than a negative authenticity result.

\paragraph{What is not in this paper.}
\textbf{No multi-annotator primary-need gold set.} We report a downstream-task transfer reading (Part F, \S\ref{sec:partf}) using extractor-derived real labels and sampled synthetic conditioning labels within the same label schema, with the absolute transfer ratio bounded by that extractor's notion of primary need (limitation~M7). A single-author $n{=}250$ gold check on the eval slice reports silver--gold agreement of $0.688$ and a gold-referenced ratio of $0.84$ (Appendix~\ref{sec:appendix-transfer}), but training labels remain unchanged and one coder is not a consensus standard: a multi-annotator gold set with inter-annotator agreement is left as future work; the closest external precedents for that extension are Synth-SBDH \citep{mitra2025synthsbdh} and label-to-data EHR \citep{li2023twodirections}.
\textbf{No full privacy-attack suite.} We run a lightweight nearest-neighbor membership-inference sketch on fact and prose channels (Part G, \S\ref{sec:appendix-mia}; fact-feature AUC $0.48$ and prose AUC $0.53$, both below the project-defined $0.55$ follow-up threshold) and a trajectory-uniqueness singling-out screen on the fact channel (Appendix~\ref{sec:appendix-mia}; the synthetic $k{=}1$ rate is $1.1\%$, below the $4$--$6\%$ real-vs-real control), but not shadow-training membership inference or imputation-based attribute inference against the synthesis model. Our privacy posture is structural (unredacted source notes, source identifier records, and real fact trajectories stay local; the PII-redacted exemplar pool remains subject to residual-redaction risk), empirical (the Part D no-copy audit, \S\ref{sec:partd}), and the Part G screens; we do not claim formal differential privacy. PromptEHR provides the two attack precedents \citep{wang2022promptehr}; \citet{puerto2025scaling} provides a scaled membership-inference precedent.
\textbf{No demographic-fairness subgroup disaggregation.} The source \texttt{client\_info.csv} does not record race, gender, or age (Data Statement, \S\ref{sec:datastatement}), so we cannot disaggregate fidelity / utility along demographic axes. The Part F appendix does report subgroup robustness along project-observable axes (trajectory length, severity, primary-issue category; Appendix~\ref{sec:appendix-transfer}). Demographic-fairness disaggregation in the spirit of \citet{ramesh2024dpeval,ramesh2025synthtexteval} remains future work.
\textbf{No higher-order trajectory dynamics.} We model trajectories at the structured-fact level (transitions, persistence, days-between gaps) but do not visualize or quantify time-axis higher-order dynamics in the prose itself (\emph{e.g.}, evolving risk language, tone trajectories), in the spirit of \citet{cbtarcs2025} for emotional arcs and \citet{petherapy2025} for protocol-adherence trajectories.

\section{Future Work}
\label{sec:appendix-futurework}

Several extensions would strengthen the corpus and evaluation beyond the bounded use summarized in \S\ref{sec:conclusion}.

First, Part~A Track-2 reports size-matched logistic AUC ($0.573$ [$0.438, 0.682$]; \S\ref{sec:partacoverage}; Table~\ref{tab:tracks}) for the single-note track ($n_\text{synth}=33$). What remains open is the symmetric prose-embedding analysis: a Track-2-scoped size-matched classifier AUC on note embeddings (analogous to Part~C but restricted to single-note clients) has not been added. Closing this gap requires embedding the $33$ single-note synthetic notes under both project encoders and running the size-matched protocol against the $350$ single-note real clients; it is the natural counterpart to the fact-side Track-2 AUC already reported.

Second, future releases should include standard benchmark splits for primary-need classification, referral extraction, caseworker-action extraction, barrier identification, next-step prediction, handoff summarization, and trajectory summarization, together with a downstream-task validation experiment in the spirit of Synth-SBDH \citep{mitra2025synthsbdh} and label-to-data EHR \citep{li2023twodirections}.

Third, longitudinal trend modeling should be improved. Practitioner feedback (\S\ref{sec:practitioner}; Appendix~\ref{sec:appendix-practitioner}) suggests that unresolved needs, delayed follow-up, recurring barriers, and long-term escalation or resolution are especially important for continuity-of-care utility.

Fourth, the structural privacy posture, overlap-based no-copy audit (\S\ref{sec:partd}), and Part G screens (nearest-neighbor MIA on fact and prose channels plus the trajectory-uniqueness singling-out screen; Appendix~\ref{sec:appendix-mia}) should be upgraded with shadow-training membership inference and synthetic-trained imputation for attribute inference \citep{wang2022promptehr}, scaled MIA \citep{puerto2025scaling}, and attribute-level sensitivity tagging during extraction (Pr1b in Appendix~\ref{sec:appendix-limitations}).

Fifth, practitioner evaluation should be broadened with more participants, more organizations, more structured tasks, and explicit coding reliability checks. A second pilot wave should also test whether the Note-3 short-synthetic-note misclassification replicates under different short-note selection (H1 in Appendix~\ref{sec:appendix-limitations}).

Sixth, fairness and subgroup-disaggregated fidelity / utility metrics across the practitioner-flagged sensitive attributes (legal status, immigration, disability, mental health, substance use, gender identity, sexuality, court involvement) are an obvious next step for high-stakes-domain synthetic data \citep{ramesh2024dpeval,ramesh2025synthtexteval}.

Seventh, higher-order time-axis prose dynamics (evolving risk language, tone trajectories) should be quantified in the spirit of \citet{cbtarcs2025} for emotional arcs and \citet{petherapy2025} for protocol-adherence trajectories.

Eighth, content-level discourse similarity could be assessed with \emph{Questions-Under-Discussion (QUD)} machinery in the spirit of \citet{namuduri2025qudsim}: a templated-QUD analysis over the 12-label documentation-move taxonomy would test whether real and synthetic notes \emph{answer} the same questions about each encounter, complementing the move-marginal / transition-matrix view in Part~B. Part~B as currently constructed measures intention-level discourse similarity (which moves appear, in what order, with how much stickiness); a case-note QUD analysis could instead assess content-level similarity (which questions a reader can recover answers to from each note). The two views are complementary: a synthetic note can preserve the documentation-move marginal while changing which under-the-surface questions get answered (or vice versa). We treat this as a natural extension rather than a current contribution because adapting QUD generation and answerability alignment to case notes would require validation in this setting.

Ninth, the Part~E precision side should be completed: the automatic LLM judge labels $3.8\%$ of populated pairs contradicted under its rubric (Appendix~\ref{sec:appendix-alignment}), but whether the prose claims values \emph{outside} the populated schema fields needs a $\sim\!100$-note human-coded gold subset to score. Closing that gap, together with human validation of the judge itself, would retire limitations~M8 and M9.

Tenth, the Part~F gold audit should be extended: the single-author $n{=}250$ gold subset (Appendix~\ref{sec:appendix-transfer}) reports silver--gold agreement and a gold-referenced transfer ratio, but a multi-annotator gold set with inter-annotator agreement, plus a relabeling pass by a second local extractor model, would directly address limitation~M7.

Eleventh, broader content and prose diversity remain open beyond the joint observation patterns and documentation gaps tested in Appendix~\ref{sec:appendix-v2}. On the content side, the v1 sampler draws fact trajectories from learned marginals, same-note conditionals, and pairwise transitions; a learned generative model over complete fact vectors (autoregressive or latent-variable, fit to the extracted-fact corpus) could capture higher-order field dependencies the factored sampler cannot, widening manifold recall (Part~C) and Track-1 coverage (Part~A) beyond what re-weighting the existing sampler reaches. On the discourse side, conditioning generation on a target documentation-move profile would steer the per-note move mix toward the real marginal, addressing the systematic offsets in Figure~\ref{fig:partbmove} (client responses and collateral contacts underrepresented; referrals and administrative updates overrepresented). On the prose side, register calibration of the synthesis output --- matching the real corpus's surface statistics such as sentence-length rhythm, punctuation and capitalization habits, and fragment rates --- targets the shallow-surface component that carries most of the Part~C classifier signal (Appendix~\ref{sec:appendix-embeddings}) and aligns with the practitioner observation that over-polished prose reads as synthetic (\S\ref{sec:practitioner}); possible levers include widening the per-note decoding-parameter range and relaxing homogenizing style constraints in the synthesis prompt. A complementary decoding-side mechanism is generation-time candidate selection --- sampling several realizations per note and retaining the one closest to corpus-level discourse and surface statistics --- which pursues the same targets without modifying the generator.

\section{Reproducibility}
\label{sec:reproducibility}

The appendix records experimental settings and results; Appendix~\ref{sec:companion} distinguishes the companion package from private reproduction inputs. The settings below describe the paper corpus; Appendix~\ref{sec:appendix-v2} separately records the software-v2 sampling experiments and their common-prose diagnostic.

\paragraph{Hardware.} Pipeline development and small-scale debugging used an Apple Silicon laptop (macOS), with spaCy \texttt{en\_core\_web\_trf} on the MPS PyTorch backend (CPU-only is supported via \texttt{NER\_DEVICE}) and local Ollama on \texttt{127.0.0.1:11434}. \emph{Production-scale stage-3 extraction} of the $1{,}682$-client / ${\sim}63$k-note real corpus was run on a public HPC cluster using NVIDIA Grace Hopper (GH200) GPU nodes, with an Apptainer-contained Ollama serving \texttt{qwen3:32b-q8\_0} and a Slurm-array workflow ($\sim 25$ clients per task and up to $20$ concurrent tasks per the user-policy concurrency cap, split into multiple waves). The model name and decoding parameters (\texttt{seed=42}, \texttt{temperature=0}, \texttt{num\_ctx=16384}, greedy + JSON mode) are held fixed across these environments. Disk: $\sim 76$\,MB for the source CSV, $\sim 100$\,MB across $\sim 63$k extracted-fact JSONs, a few hundred MB per Part C embedding cache.

\paragraph{Software.} The numerical evaluation dependencies are pinned in \texttt{requirements.txt}. The spaCy redaction stack has version ranges. Its listed transformer bridge conflicts with the evaluation pin for \texttt{transformers}, so the file does not resolve as a single environment. The exact historical redaction environment is unverified. The recorded evaluation interpreter is Python \texttt{3.14.4}. Bumping any hard-pinned version requires re-running the evaluation pipeline because numeric outputs of the sentence encoders differ across \texttt{torch} / \texttt{transformers} minor versions. Headline pins (last pinned 2026-05-08): \texttt{numpy 2.4.4}, \texttt{scipy 1.17.1}, \texttt{scikit-learn 1.8.0}, \texttt{statsmodels 0.14.6}, \texttt{spaCy 3.7+}, \texttt{openai 2.15.0} (synthesis and the Part~E judge), \texttt{sentence-transformers 5.4.1}, \texttt{transformers 5.8.0}, \texttt{torch 2.11.0}.

\begin{samepage}
\paragraph{Models.}
\begin{itemize}\sloppy
\item \emph{Redaction:} spaCy \texttt{en\_core\_web\_trf}.
\item \emph{Extraction:} \texttt{qwen3:32b-q8\_0} via local Ollama; \texttt{temperature=0}, \texttt{top\_k=1}, \texttt{repeat\_penalty=1.0}, \texttt{format="json"}, \texttt{seed=42}, \texttt{num\_ctx=16384}.
\item \emph{Synthesis + direct-prompt baseline:} OpenAI \texttt{gpt-5.2}.
\item \emph{Part E LLM judge:} OpenAI \texttt{gpt-5.6-terra}; low reasoning effort, \texttt{seed=42}, JSON output; receives synthetic fact--note pairs only.
\item \emph{Part B prototypes + Part C primary:} \texttt{BAAI/\allowbreak bge-large-en-v1.5} (1024-d).
\item \emph{Part C secondary:} \texttt{sentence-transformers/\allowbreak all-mpnet-base-v2} (768-d).
\end{itemize}
\end{samepage}

\paragraph{Seeds.} The project default is seed $42$; the original Part~A command instead defaulted to \texttt{random\_state=0}. Per-client fact-sampling seeds derive as \texttt{base\_seed + client\_id} so incremental rebuilds are stable client-by-client. Repeated matched-subset and bootstrap draws advance seeded random-number streams, with classifier cross-validation seeds fixed within each call.

\paragraph{Selected hyperparameters.}
\emph{Sampling:} Laplace $\alpha=1.0$; writer-style weights \texttt{terse\_cm 0.10 / formal\_cm 0.42 / this\_writer 0.15 / this\_cm 0.08 / first\_person 0.25}; structure-mode weights \texttt{standard 0.55 / fragments 0.15 / sentence\_per\_action 0.20 / narrative 0.10}; short/long-gap split at $7$ days; log-normal target word counts with type-conditioned log-scale $\sigma\in[0.3,1.0]$ (fallback $[0.3,0.9]$).
\emph{Synthesis:} note-type-aware base temperatures (Crisis $0.5$, Documentation-only $0.5$, Intake $0.65$, others $0.75$), jittered $\mathrm{gauss}(\text{base}, 0.12)$ clamped to $[0.45, 1.05]$; writer-specific frequency penalties $0.03$--$0.12$ and presence penalties $0.03$--$0.10$, with fallback $0.05$ for both.
\emph{Part A:} K-means $k\in\{4,\ldots,12\}$; 5-fold Lopez-Paz CV; 30 matched subsets for AUC; publish guards $\mathrm{MIN\_REAL}=100$, $\mathrm{MIN\_SYNTH}=30$.
\emph{Part B:} taxonomy version \texttt{1.0.0} (sha \texttt{d1d6de67\ldots}); 1{,}000 permutation reps; Holm correction over $n_\text{labels}=12$.
\emph{Part C:} median-heuristic RBF bandwidth on combined sample; 1{,}000 permutation reps for BGE-Large; 2{,}000 subsample per side; Fr\'echet PCA-256; classifier PCA-64; AUC 30 matched subsets with 5-fold CV; Kynk\"a\"anniemi $k=5$.

Matched AUC in Parts~A and C uses $30$ real subsets sampled without replacement, each matching the fixed synthetic pool in size, with five-fold cross-validation. Bracketed AUC intervals are the $2.5$th and $97.5$th percentiles across these repetitions.

\paragraph{Stochasticity.} We record seeds and pinned package versions for local stages: redaction, extraction (Ollama \texttt{seed} + \texttt{temp=0} + greedy + JSON mode), distribution analysis, fact sampling, Part A, Part B labeling, Part B metrics, Part C embedding, Part C metrics. The OpenAI stages are not guaranteed bit-identical: synthesis uses temperature $>0$, while the Part~E judge uses \texttt{seed=42} with low reasoning effort. Synthesis example rotation, style blocks, and prompt construction are deterministic for fixed inputs.

\paragraph{End-to-end recipe.} Run \texttt{python -m\allowbreak\ pipeline.stages.redaction.\allowbreak redact\_pii\_ner} $\rightarrow$ \texttt{python -m\allowbreak\ pipeline.stages.extraction.\allowbreak extract\_facts} $\rightarrow$ \texttt{python -m\allowbreak\ pipeline.stages.analysis.\allowbreak analyze\_distributions} $\rightarrow$ \texttt{python -m\allowbreak\ pipeline.tools.\allowbreak build\_synth\_corpus} $\rightarrow$ \texttt{python -m\allowbreak\ pipeline.tools.\allowbreak parallel\_synthesis} $\rightarrow$ Parts A/B/C/D/E/F $\rightarrow$ consolidator. The Part~E semantic re-audit runs as \texttt{python -m pipeline.tools.llm\_judge --seed 42}. Part G is invoked separately as \texttt{python -m\allowbreak\ pipeline.evaluation.\allowbreak part\_g\_privacy.nn\_mia\allowbreak\ --seed 42\allowbreak\ --export-scores\allowbreak\ outputs/\allowbreak evaluation/part\_g/\allowbreak nn\_mia\_scores.npz} (fact channel) and \texttt{python -m\allowbreak\ pipeline.evaluation.\allowbreak part\_g\_privacy.nn\_mia\_prose\allowbreak\ --seed 42\allowbreak\ --export-scores\allowbreak\ outputs/\allowbreak evaluation/part\_g/\allowbreak nn\_mia\_prose\_scores.npz} (prose channel); ROC figure rendering via \texttt{python -m\allowbreak\ pipeline.evaluation.\allowbreak part\_g\_privacy.\allowbreak render\_roc\_figure}. The companion package documents the available module commands and their required inputs; source-dependent stages require private inputs.

\paragraph{What cannot be reproduced externally.} The real corpus is held under a data-use agreement and is not redistributed; without it, stages 1--3 cannot be re-run from scratch and Parts A/B/C/D/F/G that read private real-derived inputs cannot be re-run. The package supplies the fixed synthetic corpus, schema, method implementations, and offline tests. Aggregate results are reported in this paper. Part~E checks can operate on the synthetic fact--note pairs.

\section{Data Statement}
\label{sec:datastatement}

A data statement in the style of \citet{bender2018data}, covering both the real (private) and synthetic (v1.0) corpora; a Datasheet-style companion for the fixed corpus is in Appendix~\ref{sec:datasheet}.

\paragraph{A. Curation rationale.} The real corpus was assembled to study how case managers document client encounters in a community case-management setting. The motivating question is whether a synthetic corpus generated from extracted distributions of the real corpus can preserve scenario coverage, discourse-move structure, and embedding-space semantic geometry while limiting release-side content carryover under explicit no-copy audits. Notes were filtered to those with non-empty \texttt{NOTE} text; no sub-sampling by demographic, time period, or case type was applied at curation.

\paragraph{B. Language variety.} English (en-US). Register: professional case-note documentation. Dialect / style: US-English social-services prose with conventions including \texttt{f/u}, \texttt{[NAME]}-style placeholders post-redaction, occasional ALL-CAPS for resource-given blocks, organizational acronyms. Time span (real): \texttt{CREATIONDATETIME} field spans 2016--2022. Time span (synthetic): trajectories start $6$--$24$ months before the sampling anchor and advance by sampled gaps; absolute calendar timing is not preserved (limitation~C2).

\paragraph{C. Speaker demographics.} Note authors are case managers at a single US-based agency; the corpus does not record CM-level demographics. Subjects of notes are clients of the agency; the structured fields describe socioeconomic circumstances such as housing and employment, but the source metadata does not record race, gender, or age for demographic analysis. Caseworker identity is not modeled in the synthetic pipeline (limitation~P1).

\paragraph{D. Annotator demographics.} Redaction is fully automated (spaCy \texttt{en\_core\_web\_trf} NER plus a regex layer); no human annotators redact. Sentence-move labeling (Part B) is fully automated via the zero-shot embedding labeler; no human annotators label sentences. The schema and synonym map were authored by the project team and iterated against the corpus. A single-author $n{=}250$ primary-need audit of held-out real-eval notes is reported in Appendix~\ref{sec:appendix-transfer}; no crowdworkers or paid annotators participated.

\paragraph{E. Speech situation.} Setting: documentation written by case managers after (or during) client encounters. Encounters include phone calls, in-person office visits, home visits, and contacts with collateral parties (housing providers, medical providers, courts, family). Purpose: internal program record-keeping for service delivery, reimbursement, and case continuity. Audience: other CMs, supervisors, and audit reviewers within the agency. Modality: asynchronous text typed into a case-management database.

\paragraph{F. Text characteristics.} Real-corpus statistics (over an earlier 63{,}115-note snapshot of the redacted corpus; corpus has since grown by $\sim 30$ notes to 63{,}142 raw redacted, 62{,}901 usable for evaluation; aggregates are stable to within rounding). Mean word count $91.3$, median $66$, mean sentence length $14.4$ words; tense distribution past $33.6\%$ / mixed $56.5\%$ / present $9.9\%$; visit type direct\_contact $25.3\%$ / follow\_up $15.7\%$ / triage $11.4\%$; contact method in\_person $45.6\%$ / phone $23.8\%$.

\paragraph{G. Recording quality.} Text-only; no audio. Notes are typed by humans into a software system; typos, abbreviations, and stylistic drift are inherent to the source.

\paragraph{H. Other characteristics.} PII redaction replaces names, organizations, phone numbers, addresses, and emails with placeholders (\texttt{[NAME]}, \texttt{[ORG]}, \texttt{[PHONE]}, \texttt{[ADDRESS]}, \texttt{[EMAIL]}); residual leakage risk is documented in limitation~Pr1. Notes can describe substance use, mental health crises, intimate-partner violence, eviction, and other sensitive circumstances. The real corpus is treated as confidential under the source-agency data-use agreement and is not redistributed; only the synthetic corpus and statistical summaries are shareable.

\paragraph{I. Provenance.} The real corpus is gitignored. All evaluation over real data runs locally; the boundary is enforced by the absence of \texttt{openai} imports under \texttt{pipeline/evaluation/part\_*/}. Extraction uses a local Ollama daemon (\texttt{qwen3:32b-q8\_0}, fixed seed, $\text{temperature}=0$). Synthesis uses OpenAI \texttt{gpt-5.2}; its prompts contain synthetic structured facts plus a rotated pool of PII-redacted real exemplar notes for style conditioning. The Part~E LLM-judge audit sends synthetic fact--note pairs only to OpenAI \texttt{gpt-5.6-terra}. Unredacted source notes, source identifier records, and real extracted fact trajectories stay local; the exemplar pool remains subject to residual-redaction risk (Pr1).

\paragraph{J. Ethics and IRB.} Redacted case-note text used for methodological development falls under the standard exempt-research category at the host institution; specific IRB documentation is retained by the project team. The load-bearing technical mitigations are the locality boundary (unredacted source notes, source identifier records, and extracted real facts stay local; only PII-redacted exemplars, subject to residual-redaction risk, accompany synthetic facts in synthesis prompts) and the Part D no-copy audit over the fixed corpus.

\paragraph{Synthetic corpus.} Generated per-client by sampling a structured fact trajectory from analysis-derived distributions and synthesizing prose from each fact JSON via OpenAI \texttt{gpt-5.2}, conditioned on synthetic facts plus PII-redacted few-shot exemplars and aggregate stylistic patterns. The design avoids unredacted real text and real fact trajectories in generation prompts but retains residual redaction risk (Pr1/Pr2). Audits found no residual PII in the $146$ exemplars, $0/3{,}137$ v1.0 notes with a verbatim 10-token few-shot leak, and only $13$ of $210{,}039$ unique synthetic 10-grams intersecting the $63{,}142$-note real corpus, all domain boilerplate (\S\ref{sec:partd}). \emph{Intended use:} research into synthetic-data generation, evaluation, and discourse analysis for case-management documentation; not intended for training systems that will produce real client documentation.

\section{Datasheet for the Synthetic Corpus}
\label{sec:datasheet}

A Datasheet in the style of \citet{gebru2021datasheets}, complementing the Bender--Friedman Data Statement in \S\ref{sec:datastatement}. The Datasheet covers the \emph{synthetic corpus}; the private real corpus is not redistributed.

\paragraph{1. Motivation.} \emph{For what purpose was the dataset created?} To enable NLP research, training, and benchmarking on longitudinal human-service case notes without exposing real client histories. The release responds to a structural gap: human-service case-note NLP is comparatively un-resourced \citep{gondimalla2024aligning,nelson2024homelessness}, the closest precedent \citep{pickens2023synthhumanservices} releases structured records only, and clinical adjacent corpora differ in purpose, structure, and language from case-management documentation. \emph{Who funded the dataset?} This dataset was developed as part of graduate research at The University of Texas at Austin. See Acknowledgments for funding and computing support. \emph{Comments.} The corpus, pipeline, evaluation, and Datasheet are part of the same release bundle.

\paragraph{2. Composition.} \emph{Instances.} $3{,}137$ free-text synthetic case notes across $87$ synthetic clients ($54$ multi-note, $33$ single-note), each paired with the structured synthetic fact JSON that conditioned its generation. Language variety is described in Data Statement \S B; real-source text statistics are reported in \S F. \emph{Total count.} The full fixed corpus; no sub-sampling. \emph{Sample of a larger set?} No; the release is the full generation run. \emph{Raw data?} Yes --- the note text and the fact JSONs are the raw artifacts (no derived features, no labels beyond what is structurally encoded in the fact JSON). \emph{Sensitive content?} The notes cover the sensitive-topic surface enumerated in Data Statement \S H (substance use, mental-health crisis, intimate-partner violence, eviction, legal involvement, benefits); topics are sensitive in genre but the records describe synthetic clients. Privacy verification: Part~D no-copy audit (\S\ref{sec:partd}) and Part~G fact and prose MIA sketches (\S\ref{sec:appendix-mia}).

\paragraph{3. Collection process.} \emph{How was each instance acquired?} Indirectly, via the four-stage structured-first pipeline whose privacy boundaries, local-extraction model, and external-synthesis model are recorded in Data Statement \S I (Provenance); the sampler is \texttt{sample\_facts\_proto.py} and synthesis is \texttt{generate\_notes\_proto.py}. \emph{Mechanisms.} The pipeline uses Python. Appendix~\ref{sec:reproducibility} lists the numerical package versions and redaction-environment constraints. Non-API stages use recorded seeds (Appendix~\ref{sec:reproducibility}), with per-client fact-sampling seeds \texttt{base\_seed + client\_id}, while OpenAI generation and judge outputs are not guaranteed bit-identical. \emph{Time frame.} Generation in late 2025; absolute calendar timing is not preserved on the synthetic side (Data Statement \S B, limitation~C2). \emph{Crowdworkers / paid?} No crowdworkers or paid annotators; generation and the automatic (silver) evaluation labels involve no human labeling, with a single-author $n{=}250$ primary-need audit subset on the held-out real-eval slice (Appendix~\ref{sec:appendix-transfer}). \emph{Ethics review.} See Data Statement \S J; the synthetic corpus contains no records of real individuals, though it is derived from distributions learned over human-subjects records.

\paragraph{4. Preprocessing, cleaning, labeling.} \emph{Preprocessing.} Per-note structured-fact JSON is validated against \texttt{config/valid\_schema\_values.json}; invalid values are surfaced in \texttt{data/derived/\allowbreak analysis/\allowbreak invalid\_values.json} and not silently dropped. Synonym maps are applied for known surface variants. \emph{Was raw data saved?} Yes. The dataset archive includes the synthetic facts and prose. \emph{Software available?} The companion archive includes implementations of redaction, extraction, analysis, sampling, and synthesis under \texttt{pipeline/stages/}. \emph{Labeling.} No human labels are distributed with the corpus. Part~F uses extracted real labels and sampled synthetic conditioning labels, and reports a separate single-author $n{=}250$ gold audit on the held-out real-evaluation slice (limitation~M7).

\paragraph{5. Uses.} \emph{Tasks the dataset has been used for.} The six-Part utility evaluation (Parts~A--F), the Part~G privacy-side MIA sketch, the $N{=}8$ practitioner pilot (\S\ref{sec:practitioner}), a same-model direct-prompt baseline ablation (\S\ref{sec:baseline-summary}), and a targeted category-conditioned re-sampling ablation (Appendix~\ref{sec:appendix-transfer}, not part of the release). \emph{Repository.} The corpus is available through Zenodo with individual access approval. \emph{Other tasks the dataset could be used for.} Method development for longitudinal trajectory reasoning, primary-need / barrier / referral extraction, documentation-quality assessment, discourse-move taxonomy benchmarking, train-on-synth / evaluate-on-real ablations, prompt, and pipeline robustness studies. \emph{Unsupported uses.} Producing or substituting documentation for real clients; making operational case-management decisions about real clients; redistributing under license terms that imply real-world clinical validity. The intended-use guidance is stated jointly with the Data Statement (\S\ref{sec:datastatement}, final paragraph) and reiterated in the dataset README.

\paragraph{6. Distribution.} \emph{To whom?} We provide the synthetic corpus and software to individually approved researchers, including company researchers, for strictly noncommercial research. \emph{When?} Researchers can request access at any time. \emph{How?} As versioned dataset and software archives on Zenodo; companion \texttt{CITATION.cff}. \emph{Subject to any IP, copyright, ToS?} Code and data are subject to the accompanying research-use terms; redistribution requires separate written approval. \emph{Export controls / regulatory restrictions?} None known. \emph{Maintenance.} The project team maintains the release; Hender Lin is the contact for access requests, issues, and corrections.

\paragraph{7. Maintenance.} \emph{Hosting.} The host is Zenodo, with the project team responsible for maintenance. \emph{Erratum process.} Issues reported to Hender Lin; corrections applied as a versioned point release. \emph{Will the dataset be updated?} The fixed corpus is v1.0 ($87$ synthetic clients, $3{,}137$ notes); the main release results use this corpus. A documented v1.1 targeted-expansion ablation (Appendix~\ref{sec:appendix-transfer}) explores closing the residual Part F per-class gap on the two flagged sparse-niche labels (\texttt{safety\_domestic\_violence} and \texttt{childcare\_family}) by adding $30$ category-conditioned synthetic clients ($+198$ labeled notes, $+296$ prose notes) biased $0.4/0.4$ toward those labels, but those clients are explicitly excluded from the release tarball. Future versions will be tagged separately with their own dataset version, schema version, and changelog. \emph{Older versions.} Retained under their version tags; no silent supersession. \emph{Contributions?} Approved researchers may propose additions, which the release maintainer reviews to verify the release-side privacy guard (no real-corpus text ever ships) and the schema-validation invariant (no invalid-enum values).

\section{Residual Risk Catalog}
\label{sec:riskcatalog}

Table~\ref{tab:risks} catalogs the residual risks of the release.

\begin{table*}[t]
\centering
\small
\begin{tabular}{p{3.4cm}p{3.4cm}p{3.4cm}p{3.4cm}}
\toprule
\textbf{Risk} & \textbf{Source} & \textbf{Mitigation} & \textbf{Residual limitation} \\
\midrule
Residual PII in private intermediates & Imperfect redaction & Real notes not released; local processing & No formal privacy guarantee \\
Residual PII reaching the external API & Redacted style exemplars in synthesis prompts & Manual residual-identifier review of all $146$ exemplars found $0$ residual PII; exemplars never released & Manual / point-in-time audit; no formal privacy guarantee \\
Sensitive non-PII after redaction & Legal / immigration / health / identity content survives name removal (Pr1b) & Intended-use guidance & No per-attribute sensitivity controls \\
Memorization or copying & Real redacted few-shot examples & Deterministic rotation; programmatic overlap audit (\S\ref{sec:partd}: 0/3{,}137 leaks at $\geq 10$ tokens) & Conservative threshold; not a formal attack-based audit \\
Overgeneralization & Single source organization & Data statement and scope limits & Not cross-corpus validated (G2) \\
Operational misuse & Realistic-looking synthetic notes & Intended-use guidance & Cannot technically prevent all misuse \\
Bias reproduction & Learned source patterns & Documentation and bounded use & External users cannot fully audit source bias \\
Distributional artifacts & LLM generation and sampling & Quantitative distinguishability reporting & Synthetic data remains separable \\
\bottomrule
\end{tabular}
\caption{Residual-risk catalog. Each row pairs a known risk with the project's mitigation and the limitation that mitigation does \emph{not} address.}
\label{tab:risks}
\end{table*}

\section{Companion Materials}
\label{sec:companion}

The companion package contains the fixed synthetic corpus, schema, dataset documentation, license, checksum manifest, and the paper-version method implementations with offline tests. The code includes the 12-label discourse taxonomy and evaluation procedures. Software v2 additionally implements the selected sampler in Appendix~\ref{sec:appendix-v2}. Its integration tests establish implementation parity, while the appendix reports historical experimental measurements. The packaged prompts and prototypes are sanitized adaptations; identical historical generation and scores are not claimed.

The paper itself supplies the schema and metric crosswalk (Tables~\ref{tab:schemafields} and~\ref{tab:metric-schema-map}), client-vector construction (Table~\ref{tab:partavector}), per-encoder results (Tables~\ref{tab:embedbge} and~\ref{tab:embedmpnet}), and practitioner protocol (Appendix~\ref{sec:appendix-practitioner}).

\section{Part F: Downstream-task Transfer Detail}
\label{sec:appendix-transfer}

\paragraph{Setup.} Single-label primary-need classification: predict \texttt{presenting\_issues.issues\_reported[0]} from note text. Embeddings are the cached BGE-Large-EN-v1.5 vectors from Part C (no extra inference cost). Real-corpus split: $80\%/20\%$ by \texttt{client\_id} (seed $42$); $1{,}232$ real clients enter the train pool, $308$ go to the held-out eval. Notes whose extracted-fact \texttt{issues\_reported} is empty are excluded from the labeled set ($36{,}470/62{,}901$ real notes have a silver label; $1{,}897/3{,}137$ synthetic notes on the v1.0 corpus, with $2$ singleton-class rows dropped for the MLP early-stopping head leaving an effective $1{,}895$-note train pool; $162/162$ direct-prompt-baseline notes via the \texttt{scenario} field of \texttt{\_baseline\_run\_metadata.json}). Eval-side label space is the $18$ canonical \texttt{issues\_reported} enum tokens. Heads: multinomial logistic regression (L2, balanced class weights, max\_iter $2{,}000$) and a single-hidden-layer MLP ($256$ ReLU, $200$ epochs, early stopping on a $10\%$ validation slice).

\begin{figure*}[t]
\centering
\includegraphics[width=\textwidth]{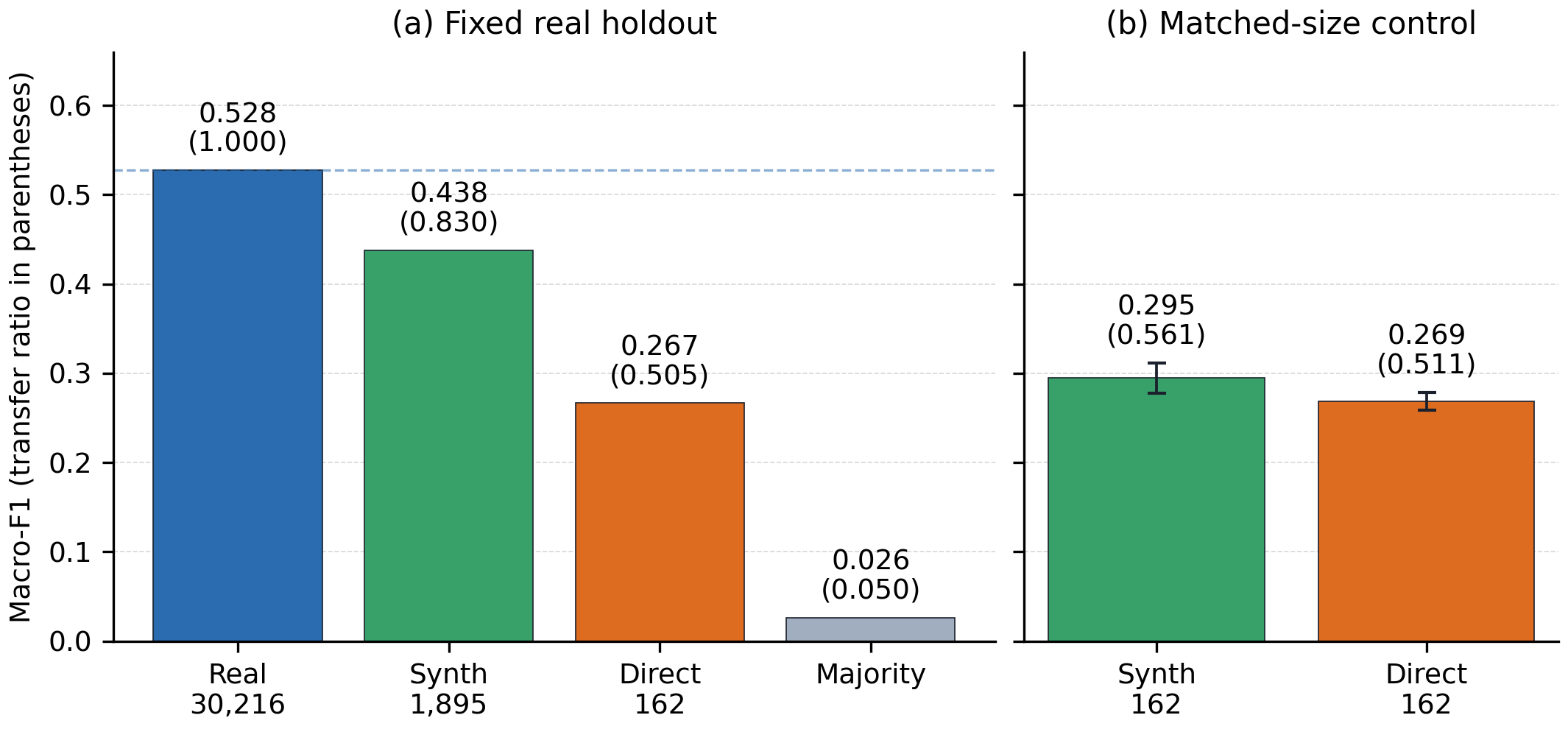}
\caption{Part F macro-F1 with BGE-Large logistic regression. Panel (a) uses the fixed real holdout; its dashed line is the real-trained reference. Panel (b) compares $162$-note training sets across ten real splits. Labels show macro-F1 and transfer ratio. Error bars in (b) show $\pm1$ SD across $200$ synthetic fits or ten direct-prompt fits.}
\label{fig:transfer}
\end{figure*}

\begin{figure*}[t]
\centering
\includegraphics[width=\textwidth]{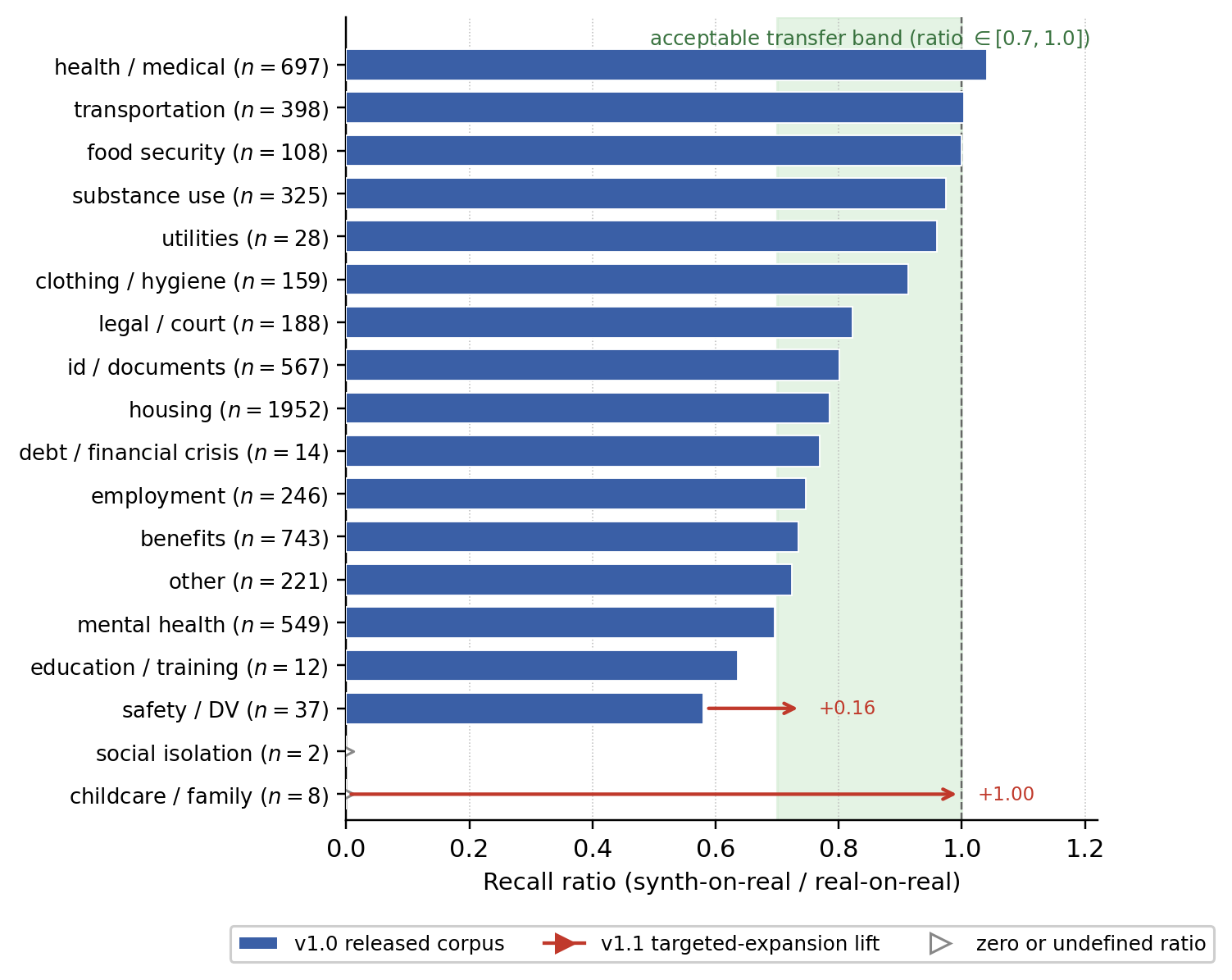}
\caption{Part F per-class recall ratios for v1.0 (BGE-Large logistic regression). Green marks the $[0.7,1.0]$ band; open triangles mark zero or undefined ratios (\texttt{social\_isolation}: $0/0$); red arrows show the non-release v1.1 lift for the two targeted sparse classes.}
\label{fig:transfer-per-class}
\end{figure*}

\paragraph{Headline numbers (v1.0 corpus).} Both heads (Table~\ref{tab:transfer-detail}) order conditions identically: train-on-real $>$ train-on-synth $\gg$ train-on-direct-prompt-baseline $\gg$ majority. The synthetic train pool is $\sim 16\times$ smaller than the real train pool ($1{,}895$ vs.\ $30{,}216$), yet recovers \textbf{$0.83$} of the upper-bound macro-F1 under logistic regression and $0.62$ under the MLP (Figure~\ref{fig:transfer} plots the logistic-regression conditions). The direct-prompt baseline ($162$ notes) recovers only $0.51$ and $0.06$ respectively under the two heads, when evaluated on the same held-out real notes; this mirrors the qualitative gap reported under Parts B and C in \S\ref{sec:baseline-summary}. The $1{,}895$ vs.\ $162$ training-pool asymmetry (${\sim}12\times$) makes Part F a size-confounded contrast for the baseline comparison; the size-insensitive Part B JSD and Part C MMD$^2$ ratios (\S\ref{sec:baseline-summary}) isolate the qualitative-fidelity effect of the structured-first pipeline. The table also shows the v1.1 ablation pool ($2{,}093$ labeled notes from $117$ clients, transfer ratio $0.83$, MLP $0.62$) so the per-class lift on the two sparse-niche labels (Figure~\ref{fig:transfer-per-class}; Table~\ref{tab:transfer-subgroup}) is visible directly --- the headline ratio remains $0.83$ while the ablation improves per-class transfer.

\paragraph{Matched-size control.} We draw $162$ training notes without replacement from all $1{,}897$ labeled release notes, retaining singleton-label candidates. We use proportional, capped stratified sampling and cached BGE-Large embeddings. Ten $80/20$ by-client real splits use outer seeds $42$--$51$; each has $20$ synthetic draws, with inner seed $1000S+r$ for outer seed $S$ and $r=0,\ldots,19$. Macro-F1 uses the labels present in each holdout ($17$ for seed $44$, otherwise $18$). The synth transfer ratio is $0.561\pm0.032$ across $200$ fits, versus $0.511\pm0.017$ across ten direct-prompt fits. We average the synthetic draws within each outer split before comparing conditions. The mean paired difference is $+0.050$ (bootstrap $95\%$ CI $[+0.038,+0.061]$; $10{,}000$ resamples of the ten paired differences, seed $42$). All ten differences are positive; the two-sided normal-approximation Wilcoxon calculation gives $p{=}0.005$.

\paragraph{Uncertainty on the headline ratio.} A by-client bootstrap attaches an interval to the $0.83$: holding both trained classifiers fixed, the $308$ eval clients are resampled with replacement ($1{,}000$ replicates, seed $42$) and the macro-F1 pair re-scored on each replicate, so the resampling unit matches the by-client split unit. The transfer ratio is $0.830$ with a $95\%$ interval of $[0.794, 0.865]$ (bootstrap SD $0.018$; macro-F1 intervals $[0.494, 0.553]$ real-on-real and $[0.409, 0.459]$ synth-on-real). The interval stays well clear of both the direct-prompt baseline ratio ($0.51$) and the size-matched synth ratio ($0.56$), so the headline ordering is robust to eval-holdout composition. Artifact: \texttt{outputs/\allowbreak evaluation/\allowbreak part\_f/\allowbreak transfer\_ratio\_ci.json}.

\paragraph{Encoder-robustness check.} Re-running the same train/eval split end-to-end under the second project encoder, \texttt{sentence-transformers/all-mpnet-base-v2} ($768$-d; cached during the Part C dual-encoder run), recovers a transfer ratio of $0.82$ under logistic regression (macro-F1 $0.407/0.499$; weighted-F1 $0.532$; accuracy $0.508$) and $0.63$ under the MLP (macro-F1 $0.324/0.517$). Both heads land within $\pm 0.02$ of the BGE-Large numbers ($0.83$ logreg, $0.62$ MLP) and preserve the same condition ordering, so the headline transfer-ratio claim is not a property of the BGE-Large embedding geometry. Direct-prompt baseline transfer ratios under MPNet are $0.51$ (logreg) and $0.35$ (MLP). We therefore present Part F as a dual-encoder result, mirroring the Part C protocol.

\paragraph{Author-coded gold audit of the silver labels.} A single-author gold subset bounds how much the silver labels themselves distort the eval-side reading (limitation~M7). One author labeled $250$ notes from the held-out real-eval slice with the primary-need label (same $18$-value enum), sampled proportionally by silver label with at most $2$ notes per client (seed $42$); the annotation sheet showed only the note text (no silver label, no classifier prediction) and interleaved $25$ repeated notes among the $250$ as a self-consistency check ($275$ rows total; $25/25$ repeats relabeled identically). Silver--gold agreement is $0.688$ [$0.631, 0.745$] unweighted ($0.696$ weighted back to the eval-slice label distribution; Cohen's $\kappa{=}0.646$, chance agreement $0.119$), so roughly $3$ in $10$ eval labels differ from a careful human read --- agreement is high where the primary need is unambiguous (\texttt{housing} $0.85$ at $n{=}65$; \texttt{transportation} $0.94$ at $n{=}16$) and lowest on the catch-all \texttt{other} ($2/10$ confirmed) and on multi-need notes where extractor and coder pick different threads. The consequence for the headline is small: holding the real-trained, synth-trained, and direct-prompt-baseline logistic-regression heads of Table~\ref{tab:transfer-detail} fixed and re-scoring them against the author gold on these $250$ notes gives macro-F1 $0.598$ (real), $0.502$ (synth), and $0.388$ (baseline), so the gold-referenced transfer ratio is \textbf{$0.84$} [$0.74, 0.94$, note-level bootstrap] for synth and $0.65$ for the baseline --- the synth-over-baseline ordering survives under gold eval labels, and the gold-referenced synth ratio interval covers the silver-referenced $0.83$ (Table~\ref{tab:transfer-gold}). Silver-label noise therefore shifts absolute macro-F1 while leaving the cross-condition ordering intact. Caveats: a single author coder (no inter-annotator agreement), the gold check covers the eval side only (training labels remain unchanged), and the second-extractor relabel remains open (Appendix~\ref{sec:appendix-futurework}). Artifacts: \texttt{outputs/\allowbreak evaluation/\allowbreak part\_f/\allowbreak gold\_report.json}, \texttt{gold\_extra\_conditions.json}; scoring via \texttt{python -m\allowbreak\ pipeline.\allowbreak evaluation.\allowbreak part\_f\_downstream.\allowbreak score\_gold} and \texttt{score\_gold\_extra\_conditions}.

\begin{table}[t]
\centering
\small
\setlength{\tabcolsep}{3pt}
\begin{tabular}{lr}
\toprule
\textbf{Quantity} & \textbf{Value} \\
\midrule
Gold subset size                & $250$ \\
Intra-rater consistency         & $25/25$ \\
Silver--gold agreement          & $0.688$ [$0.631, 0.745$] \\
\quad weighted                  & $0.696$ \\
Gold macro-F1 (real head)       & $0.598$ \\
Gold macro-F1 (synth head)      & $0.502$ \\
Gold macro-F1 (baseline head)   & $0.388$ \\
Gold-referenced ratio (synth)   & \textbf{$0.84$} [$0.74, 0.94$] \\
Gold-referenced ratio (baseline)& $0.65$ \\
Silver-referenced ratio         & $0.83$ [$0.79, 0.87$] \\
\bottomrule
\end{tabular}
\caption{Author-coded audit of Part F silver labels on $250$ held-out real notes (single blinded coder; seed $42$). Intervals are Wald for agreement, note-level bootstrap for the gold-referenced synth ratio, and by-client bootstrap for the silver-referenced ratio; ``weighted'' follows the evaluation label mix.}
\label{tab:transfer-gold}
\end{table}

\begin{table}[t]
\centering
\small
\setlength{\tabcolsep}{3pt}
\begin{tabular}{llrrr}
\toprule
\textbf{Head} & \textbf{Cond.} & \textbf{Macro-F1} & \textbf{Acc.} & \textbf{Ratio} \\
\midrule
\multirow{5}{*}{logreg}
  & real                       & 0.528 & 0.674 & 1.00 \\
  & synth (v1.0 release)       & \textbf{0.438} & 0.566 & \textbf{0.83} \\
  & \quad + v1.1 ablation$^{\dagger}$ & 0.438 & 0.564 & 0.83 \\
  & baseline                   & 0.267 & 0.346 & 0.51 \\
  & majority                   & 0.026 & 0.312 & 0.05 \\
\midrule
\multirow{5}{*}{mlp}
  & real                       & 0.565 & 0.737 & 1.00 \\
  & synth (v1.0 release)       & \textbf{0.348} & 0.615 & \textbf{0.62} \\
  & \quad + v1.1 ablation$^{\dagger}$ & 0.349 & 0.597 & 0.62 \\
  & baseline                   & 0.033 & 0.106 & 0.06 \\
  & majority                   & 0.026 & 0.312 & 0.05 \\
\bottomrule
\end{tabular}
\caption{Part F transfer across logistic-regression and MLP heads on $6{,}254$ held-out real notes ($308$ clients; $18$ labels). Bold rows use v1.0; $^{\dagger}$ rows add the non-release v1.1 targeted-expansion ablation.}
\label{tab:transfer-detail}
\end{table}

\paragraph{Per-class behavior (synth-on-real, logreg, v1.0).} Figure~\ref{fig:transfer-per-class} plots the v1.0 per-class recall-ratio breakdown together with the per-class lift arrows obtained under the v1.1 ablation. Top-support classes transfer cleanly on v1.0: \texttt{id\_or\_documents} F1 $0.70$ ($n{=}567$), \texttt{health\_medical} $0.66$ ($n{=}697$), \texttt{housing} $0.64$ ($n{=}1{,}952$). Mid-support classes are stable in the $0.5$--$0.6$ range. Transfer failures cluster at the long tail: \texttt{childcare\_family} ($n{=}8$) and \texttt{social\_isolation} ($n{=}2$) collapse to F1 $0.0$ on v1.0, and the catch-all \texttt{other} drops to $0.18$. The v1.1 ablation (next paragraph) improves two targeted sparse-niche labels on the same eval split but is not part of the release. The residual long-tail behavior motivates expanded synthetic coverage (limitation~C3).

\paragraph{Targeted-expansion ablation (v1.1, category-conditioned re-sampling; not in release).} We tested whether transfer on the two flagged sparse-niche labels (\texttt{safety\_domestic\_violence} ratio $0.58$, \texttt{childcare\_family} ratio $0.00$ at $n_{\text{eval}}{=}8$) could be improved without re-architecting the pipeline by generating $30$ targeted synthetic clients ($+198$ labeled notes, $+296$ prose notes), biased $0.4/0.4$ toward the two flagged labels with a persistence-floor boost of $0.6$ so the niche thread propagates across each trajectory. The v1.0~$\rightarrow$~v1.1 lift on the same eval split is \textbf{$0.74$} for DV (+$0.16$) and \textbf{$1.00$} for childcare-family (+$1.00$), without changing the headline ratio ($0.830 \rightarrow 0.831$). Per-label recalls lift on the two targeted classes (\texttt{childcare\_family} $0.00 \rightarrow 0.375$; \texttt{safety\_domestic\_violence} $0.486 \rightarrow 0.622$) and on selected mid-support classes (\emph{e.g.}, \texttt{health\_medical} $0.84 \rightarrow 0.86$). The $30$ v1.1 clients are documented here as an ablation that demonstrates targeted sparse-niche transfer can be improved with modest synthesis budget without re-architecting the pipeline; they are explicitly excluded from the dataset tarball.

\paragraph{Subgroup-robustness audit (v1.0 vs.\ v1.1 ablation).} We refit logistic regression and evaluate on the same $20\%$ by-client real holdout, stratified along three project-observable axes (Table~\ref{tab:transfer-subgroup}). Unlike the headline experiment, this audit retains singleton-label training examples, giving synthetic training pools of $1{,}897$ notes for v1.0 and $2{,}094$ for the v1.1 ablation. We deliberately omit demographic-fairness disaggregation: the source \texttt{client\_info.csv} does not record race, gender, or age (data statement, \S\ref{sec:datastatement}). \emph{Trajectory length.} Transfer remains strongest on long trajectories (6+ notes, ratio $0.83$ on v1.0); the $30$ targeted-expansion clients contribute $296$ notes, and the ablation lifts the $2$--$5$-note bucket from $0.72$ to $0.89$ as a side effect. \emph{Extractor-derived documented severity.} Ratios are $0.82$--$0.89$ for minor / moderate / major / crisis on v1.0; this audit detects no performance drop with documented severity under the shared extractor. \emph{Primary-issue category.} The two flagged sparse-niche labels improve under the v1.1 ablation: \texttt{safety\_domestic\_violence} $0.58 \rightarrow 0.74$ and \texttt{childcare\_family} $0.00 \rightarrow 1.00$ (Table~\ref{tab:transfer-subgroup}), at the cost of a small drop on the residual catch-all \texttt{other} bucket ($0.67 \rightarrow 0.51$) and a marginal dip in \texttt{housing} ($0.78 \rightarrow 0.77$). Common categories transfer at or above parity in both versions (\texttt{health\_medical} $1.04$, \texttt{transportation} $1.00$, \texttt{food\_security} $1.00$, \texttt{substance\_use} $0.97$). This single targeted ablation shows that two sparse-category gaps can improve with additional category-conditioned synthesis budget; it does not establish a general scaling law.

\begin{table*}[t]
\centering
\small
\setlength{\tabcolsep}{3pt}
\begin{tabular}{lrrr}
\toprule
\textbf{Stratum} & \textbf{$n_{\text{eval}}$} & \textbf{v1.0 ratio} & \textbf{+v1.1 ablation} \\
\midrule
\multicolumn{4}{l}{\textit{By trajectory length}} \\
\quad 1 note               &    51 & 0.70 & 0.72 \\
\quad 2--5 notes           &   181 & 0.72 & 0.89 \\
\quad 6+ notes             & $6{,}022$ & 0.83 & 0.83 \\
\midrule
\multicolumn{4}{l}{\textit{By issue severity}} \\
\quad minor                &   297 & 0.85 & 0.87 \\
\quad moderate             & $2{,}900$ & 0.83 & 0.84 \\
\quad major                &   811 & 0.82 & 0.86 \\
\quad crisis               &   130 & 0.89 & 0.96 \\
\quad unknown              & $2{,}116$ & 0.82 & 0.82 \\
\midrule
\multicolumn{4}{l}{\textit{By primary issue (recall ratio)}} \\
\quad housing                       & 1{,}952 & 0.78 & 0.77 \\
\quad benefits\_financial           &   743 & 0.73 & 0.76 \\
\quad health\_medical               &   697 & 1.04 & 1.07 \\
\quad id\_or\_documents             &   567 & 0.82 & 0.82 \\
\quad mental\_health                &   549 & 0.70 & 0.73 \\
\quad transportation                &   398 & 1.00 & 0.98 \\
\quad substance\_use                &   325 & 0.97 & 0.97 \\
\quad employment                    &   246 & 0.74 & 0.76 \\
\quad legal\_court                  &   188 & 0.82 & 0.79 \\
\quad clothing\_hygiene             &   159 & 0.92 & 0.91 \\
\quad food\_security                &   108 & 1.00 & 1.00 \\
\quad safety\_domestic\_violence    &    37 & \textbf{0.58} & \textbf{0.74} \\
\quad utilities                     &    28 & 0.96 & 0.96 \\
\quad debt\_financial\_crisis       &    14 & 0.77 & 0.77 \\
\quad education\_training           &    12 & 0.64 & 0.64 \\
\quad childcare\_family             &     8 & \textbf{0.00} & \textbf{1.00} \\
\bottomrule
\end{tabular}
\caption{Part F subgroup robustness for v1.0 and the non-release v1.1 ablation. Trajectory-length and severity rows report macro-F1 ratio; issue rows report recall ratio; bold marks the two targeted sparse classes.}
\label{tab:transfer-subgroup}
\end{table*}

\paragraph{Reproduction.} Invoke as \texttt{python -m pipeline.\allowbreak evaluation.\allowbreak part\_f\_downstream.\allowbreak run\_transfer \allowbreak --seed 42}. Inputs are the cached BGE-Large embeddings under \texttt{outputs/\allowbreak evaluation/\allowbreak part\_c/} plus the local fact JSONs under \texttt{data/\allowbreak private/\allowbreak processed/\allowbreak extracted\_facts/} and \texttt{data/\allowbreak synthetic/\allowbreak facts/}; outputs are \texttt{outputs/\allowbreak evaluation/\allowbreak part\_f/\allowbreak transfer\_report.json} and \texttt{per\_class\_report.json}. The subgroup-robustness audit is invoked as \texttt{python -m pipeline.\allowbreak evaluation.\allowbreak part\_f\_downstream.\allowbreak subgroup\_robustness \allowbreak --seed 42}; output is \texttt{outputs/\allowbreak evaluation/\allowbreak part\_f/\allowbreak subgroup\_robustness.json}.

\section{Part G: MIA and Singling-Out Detail}
\label{sec:appendix-mia}

\paragraph{Threat model.} Honest-but-curious adversary with white-box access to the synthetic corpus and to a candidate real note. The adversary scores each candidate by $s = -\min_{s_j \in \text{synth}} d_{\cos}(v(\text{real}), v(s_j))$; under the membership-inference hypothesis a real note that conditioned synthesis (\emph{i.e.}, served as a few-shot exemplar) should land closer to the synthetic neighborhood, yielding a higher score. We treat ROC-AUC as the headline and run the sketch on the two natural representations of a case note: its extracted structured-fact JSON and its prose text.

\paragraph{Structured-fact channel --- featurization.} Each fact JSON is encoded as a sparse $358$-dimensional multi-hot vector over every populated value of $33$ singleton categorical fields (\emph{e.g.}, \texttt{visit\_type}, \texttt{contact\_method}, \texttt{client\_affect}) and $7$ list-valued categorical fields (\emph{e.g.}, \texttt{issues\_reported}, \texttt{services\_provided}, \texttt{document\_types}, \texttt{contacted\_parties}). Distances are L2-normalized cosine; min-distance is computed by block-matrix multiplication so the $61{,}771 \times 3{,}137$ pairwise comparison fits in memory.

\paragraph{Membership ground truth.} A real note is a \emph{member} iff its redacted text matches a file in \texttt{data/private/\allowbreak processed/\allowbreak few\_shot\_examples/} either as a normalized exact match (case-folded, whitespace-collapsed) or as a contiguous substring (with an 80-character minimum to avoid stock-phrase false positives). $127$ of $146$ few-shot files map onto $127$ unique real notes; $126$ remain after dropping notes whose fact vector is all-zero for the structured-fact channel ($127$ retained for the prose channel, whose featurization is dense). Non-members: $61{,}645$ real notes scored on the fact channel; $62{,}774$ on the prose channel.

\paragraph{Structured-fact channel --- result.} AUC $= 0.475$ (Table~\ref{tab:mia-detail}); members and non-members largely overlap on the min-cosine-distance distribution (members mean $0.512$, non-members mean $0.503$; medians $0.508$ vs.\ $0.499$). The project-defined $0.55$ follow-up threshold is not exceeded for the structured-fact channel.

\paragraph{Prose channel --- featurization and encoder rationale.} Each note's prose is encoded with the cached Part C MPNet embeddings (\texttt{sentence-transformers/all-mpnet-base-v2}, $768$-d). We use MPNet because its symmetric sentence-pair objectives (NLI, paraphrase, STS) match the note-to-note nearest-neighbor structure of this attack without assigning query and document roles. The synthetic pool is the full v1.0 prose corpus ($3{,}137$ notes).

\paragraph{Prose channel --- result.} AUC $= 0.535$ (Table~\ref{tab:mia-detail}); members and non-members again overlap on the min-cosine-distance distribution (members mean $0.241$, non-members mean $0.254$; medians $0.241$ vs.\ $0.243$). The project-defined $0.55$ follow-up threshold is not exceeded for the prose channel either. The prose result lies closer to the threshold than the fact result and therefore warrants greater caution.

\paragraph{Threshold rationale.} AUC $0.5$ corresponds to a random adversary. We pre-specify a project-defined AUC $\leq 0.55$ no-trigger band (a $+0.05$ margin over chance), AUC $\in (0.55, 0.65]$ as a weak-signal follow-up band, and AUC $> 0.65$ as a stronger-signal follow-up band. These are operational triage rules, not literature-standard privacy guarantees. The cutoff is encoded as the default \texttt{--leak-threshold 0.55} in the attack scripts (\texttt{pipeline/\allowbreak evaluation/\allowbreak part\_g\_privacy/\allowbreak nn\_mia.py}, \texttt{nn\_mia\_prose.py}) prior to running the audit, so the pass/fail call cannot be tuned post hoc.

\begin{table}[!htbp]
\centering
\small
\setlength{\tabcolsep}{3pt}
\begin{tabular}{lrr}
\toprule
\textbf{Metric} & \textbf{Fact} & \textbf{Prose} \\
\midrule
Encoder & sparse 358-d & MPNet 768-d \\
ROC-AUC & 0.475 & 0.535 \\
Pre-specified threshold & 0.55 & 0.55 \\
Verdict & not detected & not detected \\
\midrule
Members scored ($n$) & 126 & 127 \\
Non-members scored ($n$) & 61{,}645 & 62{,}774 \\
Synth pool size & 3{,}137 & 3{,}137 \\
\midrule
\multicolumn{3}{l}{\textit{Mean min-cos-distance}} \\
\quad members & 0.512 & 0.241 \\
\quad non-members & 0.503 & 0.254 \\
\bottomrule
\end{tabular}
\caption{Part G nearest-neighbor MIA on structured-fact and prose channels for the v1.0 corpus. Both AUCs are below the project-defined $0.55$ follow-up threshold; this screen is not a formal privacy guarantee.}
\label{tab:mia-detail}
\end{table}

\begin{figure}[!htbp]
\centering
\includegraphics[width=\columnwidth]{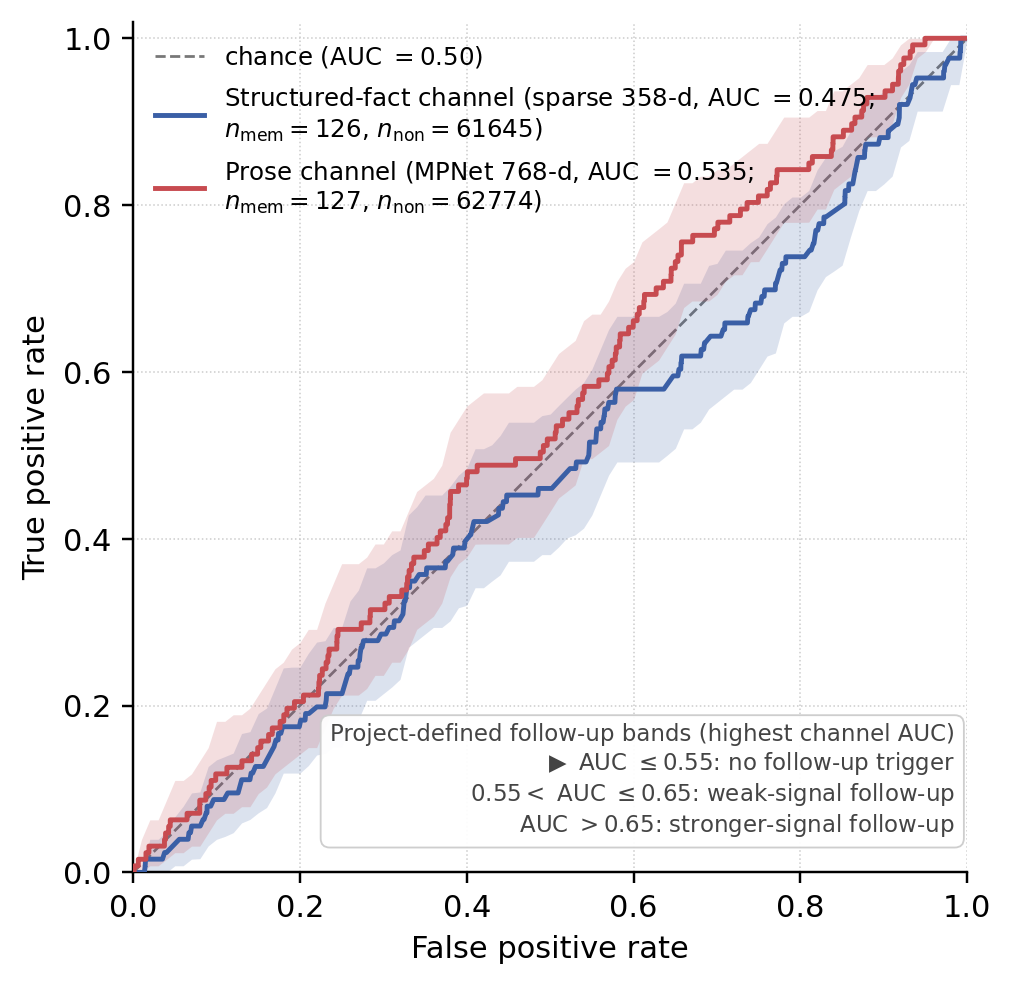}
\caption{Part G ROC curves for the structured-fact and prose channels, with $95\%$ bootstrap intervals and project-defined follow-up bands.}
\label{fig:partg-roc}
\end{figure}

\paragraph{Trajectory-uniqueness / singling-out screen.} The MIA sketch tests few-shot membership; a separate question is whether a synthetic trajectory's \emph{combination} of events could single out a specific real client. We reduce every client (real and synthetic) to a trajectory signature---the set of categorical field$=$value tokens populated anywhere in the client's fact trajectory, over the same enum fields as the MIA featurization---and count, for each of the $87$ synthetic clients, the number $k$ of real clients whose signature contains the synthetic signature. $k{=}1$ is the singling-out case: exactly one real person exhibits every attribute combination the synthetic trajectory encodes. Two adversary granularities are scored: a \emph{stable-profile} view restricted to eight persistent quasi-identifier fields (housing, employment, benefits, substance-use, mental- and physical-health statuses, reported issue categories, crisis types), and a \emph{full-trajectory} view over all $40$ enum fields. A leave-one-out real-vs-real control calibrates both readings.

\paragraph{Singling-out result.} At the stable-profile granularity the median synthetic signature matches $157$ real clients; $16$ of $87$ synthetic clients match none (fictional composites) and $1$ of $87$ ($1.1\%$) matches exactly one. At the full-trajectory granularity the median is $31$, with $26$ matching none and again $1$ matching exactly one. The real-vs-real control shows genuinely observed people are \emph{more} individually identifying than the synthetic clients at both granularities: $4.4\%$ of real clients are $k{=}1$ unique against the rest of the corpus on stable profiles and $5.7\%$ on full trajectories. Both $k{=}1$ synthetic clients resolve to the same mechanism: in each case the single ``matching'' real client is an extreme-coverage trajectory ($448$ and $362$ notes; signature supersets of $69$ and $259$ tokens, respectively), so the containment reflects that client's breadth rather than a re-identifying resemblance. The screen is one adversary model (set containment on categorical trajectory signatures), not a formal $k$-anonymity guarantee. Artifact: \texttt{outputs/\allowbreak evaluation/\allowbreak part\_g/\allowbreak trajectory\_\allowbreak uniqueness.\allowbreak json}.

\paragraph{Reproduction.} Invoke \texttt{python -m pipeline.evaluation.\allowbreak part\_g\_privacy.\allowbreak nn\_mia \allowbreak --seed 42 \allowbreak --export-scores outputs/\allowbreak evaluation/\allowbreak part\_g/\allowbreak nn\_mia\_scores.\allowbreak npz} (fact channel; writes \texttt{outputs/\allowbreak evaluation/\allowbreak part\_g/\allowbreak mia\_report.\allowbreak json} and \texttt{outputs/\allowbreak evaluation/\allowbreak part\_g/\allowbreak nn\_mia\_scores.\allowbreak npz}) and \texttt{python -m pipeline.evaluation.\allowbreak part\_g\_privacy.\allowbreak nn\_mia\_prose \allowbreak --seed 42 \allowbreak --export-scores outputs/\allowbreak evaluation/\allowbreak part\_g/\allowbreak nn\_mia\_prose\_scores.\allowbreak npz} (prose channel; writes \texttt{nn\_mia\_prose.\allowbreak json} and \texttt{nn\_mia\_prose\_scores.\allowbreak npz}); both read local fact JSONs / cached Part C MPNet embeddings and few-shot exemplar texts. The singling-out screen runs via \texttt{python -m pipeline.evaluation.\allowbreak part\_g\_privacy.\allowbreak trajectory\_\allowbreak uniqueness}. Figure~\ref{fig:partg-roc} rendered via \texttt{python -m pipeline.evaluation.\allowbreak part\_g\_privacy.\allowbreak render\_roc\_figure}.

\clearpage
\section{Structured-Sampling Extension (Software v2)}
\label{sec:appendix-v2}

\subsection{Relationship to the paper method}

Can changes to fact sampling address the trajectory undercoverage in \S\ref{sec:partacoverage}? We report a controlled follow-up and a downstream diagnostic for the structural sampler incorporated into software v2. The fixed $3{,}137$-note corpus and main results remain v1.0. This follow-up used separate implementations and synthetic histories; it is also distinct from the v1.1 targeted-expansion ablation in Appendix~\ref{sec:appendix-transfer}.

The extension builds on a profile-based sampler. The profile parent assigns each synthetic client an observation profile governing which fields are documented and a topic profile governing which needs recur. Conditional collateral-contact and next-step events are inherited from this parent. The selected extension replaces fieldwise observation masks with joint patterns conditioned on history length, first-note status, and the preceding pattern, while retaining topic profiles. It also uses full length-conditioned distributions of gaps between notes. These changes target documentation regularities; documentation order does not establish encounter chronology or clinical coherence. We call this combination the \emph{selected sampler}; its implementation identifier is \texttt{both\_profiles\_\_OG}, where O denotes joint observation and G denotes documentation gaps.

Software v2 incorporates this combination with training-only fitting, explicit documentation ordering, deterministic random streams, and checks on synthetic inputs and output locations. It excludes the experimental joint contact/status mechanism, whose interpretation remained unresolved, and later prose-generation policies. The comparator below is the profile parent, not the original paper pipeline. The software retains the paper's generation model and policy; Appendix~\ref{sec:companion} describes the release adaptations and limits of implementation-parity tests.

\subsection{Controlled evaluation protocol}

The structured comparison retains a grouped split of $799$ training, $266$ development, and $266$ test multi-note clients, fixed with seed $42$. The clients assigned to the test split had already been studied during development and do not form a fresh confirmation set. We use the $33$ trajectory features described in Appendix~\ref{sec:appendix-schema}, standardized using training clients only. Distances are Euclidean; the radius is the median training-client nearest-neighbor distance.

\emph{Coverage} is the fraction of the $266$ query clients with a synthetic neighbor inside this radius. Short-history coverage restricts the denominator to $165$ clients with $2$--$20$ notes; long-history coverage uses $66$ clients with $51+$ notes. The remaining $35$ clients have $21$--$50$ notes. \emph{Proximity} is the fraction of synthetic clients within the radius of a real training client. It measures structured distributional fit, not prose quality or privacy.

The comparison family was fixed after development screening. Each of $50$ paired generation repetitions produces $200$ synthetic clients per arm, with matched trajectory lengths and note budgets. A nested first-$54$-client subset tests sensitivity to corpus size. Paired differences use $10{,}000$ bootstrap draws over generation repetitions. Simultaneous $95\%$ intervals use the maximum standardized bootstrap deviation across the original family of $70$ endpoints: five parent comparisons, two corpus sizes, and seven metrics. They condition on the fixed real cohort and training fit; they do not include training-fit, extraction, or population uncertainty.

\subsection{Structured coverage results}

The selected sampler improves overall structured coverage relative to its profile parent by $3.95$ percentage points [$2.81,5.10$] across $50$ paired repetitions at $200$ synthetic clients per arm. Table~\ref{tab:v2-coverage} reports overall coverage, the primary short-history outcome, the long-history tradeoff, and proximity. The original simultaneous adjustment is retained for this subset of outcomes.

\begin{table*}[t]
\centering
\small
\begin{tabular}{lrrr}
\toprule
\textbf{Measure} & \textbf{Profile parent (\%)} & \textbf{Selected sampler (\%)} & \textbf{Change in pp [95\% interval]} \\
\midrule
Overall coverage & 23.74 & 27.69 & $+3.95\ [2.81,5.10]$ \\
Coverage, 2--20 notes & 3.72 & 9.13 & $+5.41\ [3.80,7.01]$ \\
Coverage, 51+ notes & 71.30 & 69.73 & $-1.58\ [-3.34,0.19]$ \\
Proximity & 46.21 & 57.06 & $+10.85\ [9.37,12.33]$ \\
\bottomrule
\end{tabular}
\caption{Structured sampling comparison: means over $50$ paired repetitions with $200$ synthetic clients per arm. Coverage denominators are $266$ real clients overall, $165$ for short histories, and $66$ for long histories. Proximity uses the $200$ synthetic clients and the real training reference. Changes are selected sampler minus profile parent, with simultaneous intervals over the original $70$-endpoint family. These settings differ from Part~A's historical corpus evaluation.}
\label{tab:v2-coverage}
\end{table*}

At $200$ synthetic clients per arm, the mean short-history gain exceeded the predefined three-point practical target and its simultaneous interval excluded zero. At the $54$-client sensitivity budget, short-history coverage increased from $0.92\%$ to $2.96\%$: $+2.04$ points [$1.05,3.02$], below the predefined three-point practical target. Long-history coverage at the larger budget had a negative point estimate, with an interval spanning zero. Thus the gains depend on history length and corpus size. The historical Part~A score uses a different reference cohort, geometry, and generation budget; subtracting it from these controlled results would not estimate the extension's effect.

\subsection{Downstream evidence and remaining limits}
\label{sec:v2-downstream}

Do these structured gains improve note-classification transfer? A later diagnostic compared the selected sampler with its parent under a common experimental prose policy. It used \texttt{gpt-6.1-sol} with low reasoning effort and a detailed free-prose prompt, retaining the previous accepted synthetic note and its facts as context within each history. Both arms used the same policy and $75$ audited training-only PII-redacted style exemplars selected from the existing $146$-exemplar pool. Real extracted trajectories were not sent to the external model, and exemplars are not distributed. Drafts underwent automated factual review with \texttt{gpt-5.6-terra}, with up to five generation attempts per note; these judgments are not practitioner validation.

Each of three repetitions generated $15$ complete new histories per arm. Generated note budgets matched within each repetition, but the number of notes eligible for classification differed because some sampled facts lacked a first-issue label (Table~\ref{tab:v2-transfer}). All planned notes were generated for both arms before analysis. Classification used normalized local BGE-Large embeddings and class-balanced logistic regression ($C=1$, \texttt{lbfgs}, maximum $2{,}000$ iterations). The fixed first-issue task had $18$ labels and $7{,}410$ evaluation notes from $313$ real clients. This reused evaluation cohort and experimental prose policy differ from the paper's Part~F protocol.

\begin{table}[t]
\centering
\small
\setlength{\tabcolsep}{3pt}
\begin{tabular}{rrrr}
\toprule
\textbf{Rep.} & \shortstack{\textbf{Notes}\\\textbf{/arm}} & \shortstack{\textbf{Eligible}\\\textbf{parent/selected}} & \shortstack{\textbf{F1 change (pp)}\\\textbf{[95\% interval]}} \\
\midrule
1 & 1{,}119 & 703/707 & $+2.32\ [1.37,3.47]$ \\
2 & 861 & 518/510 & $+0.50\ [-1.06,1.84]$ \\
3 & 871 & 541/535 & $-0.09\ [-1.10,0.81]$ \\
\bottomrule
\end{tabular}
\caption{Common-prose downstream diagnostic with $15$ generated histories per arm per repetition and a reused $313$-client evaluation cohort ($7{,}410$ notes). Notes/arm counts all generated notes; eligible counts are label-eligible training notes. Macro-F1 differences are selected sampler minus profile parent. Intervals are descriptive paired real-client bootstrap intervals conditional on each generated training set; they do not use the structural comparison's simultaneous adjustment.}
\label{tab:v2-transfer}
\end{table}

The macro-F1 changes were inconsistent across the three repetitions and failed the predefined screen requiring at least a two-point gain in every repetition. This does not establish absence of benefit, but it does not support reliable downstream improvement. The diagnostic evaluates a different prose policy from stable software v2; it is not an end-to-end evaluation of that software.

Neither experiment establishes improved clinical realism, privacy, or generalization to another population. Coverage after re-extracting facts from generated prose was not measured, and a replacement generator was not validated for the v1.0 pipeline.

\end{document}